\documentclass[twoside,11pt]{article}

\usepackage[T1]{fontenc}
\usepackage[utf8]{inputenc}
\usepackage{amsmath,amssymb,amsthm}
\usepackage[preprint]{jmlr2e}
\usepackage{booktabs}
\usepackage{array}
\hypersetup{hidelinks}

\theoremstyle{definition}
\newtheorem{counterexample}[theorem]{Counterexample}

\jmlrheading{}{2026}{}{}{}{}{Takahiro Kato}
\ShortHeadings{Label-Free Calibration Certificates}{Kato}
\firstpageno{1}

\begin{document}

\title{Sharp and Stable Label-Free Calibration Certificates from Logical Coherence}

\author{\name Takahiro Kato \email taka@np-hard.co.jp \\
       \addr NP-hard, Inc.}

\maketitle

\begin{abstract}%
Query-conditioned probabilistic predictors, such as typed decision models, zero-shot classifiers and
the answer probabilities of language models, answer logically related questions one at a time. Their
answers need not be coherent, and the labels needed to check their calibration are often missing. We
ask what the law of the answers and their logical structure, without any label, can refute about
calibration. For every restriction on the information that a bettor's stakes may use, an
information-restricted minimax theorem identifies, for finitely supported laws of the answers, the
sharp label-free lower bound on the corresponding calibration residual: the largest bound that the
law of the answers and the logical structure allow. The theorem contains measures of incoherence,
feasible joint posteriors and the mutual calibration of forecasters as special cases. On bundles of
different, logically related events its mean, marginal and joint stages can separate strictly, and a
bound on a bundle cannot be attributed to a given one of its questions without assumptions on the
others. The sharp bound on marginal calibration error is discontinuous in the answers, the jump
needs logical constraints among the questions, and no label-free lower bound that is continuous in
the Wasserstein distance between laws of answers stays within a constant factor of it. Restricting
the stakes to Lipschitz functions gives a stable bound that is sharp for a smooth calibration
residual and, for every positive Lipschitz constant, is positive exactly when the sharp marginal
bound is. On an i.i.d.\ stream of bundles, stakes represented at fixed nodes and learnt by projected
online gradient ascent make it computable, with fixed-horizon certificates whose shortfall is
bounded by grid, learning and statistical terms, and predictable betting gives anytime-valid ones.
In a preregistered study of four predictors on two data sets, all seven confirmatory certificates
were positive (the eighth pair did not pass the pilot). A certificate shows only that calibration is
at least this bad; a certificate of zero means ``not refuted''.
\end{abstract}


\section{Introduction}
\label{sec:intro}

A query-conditioned probabilistic predictor takes an input together with a question posed at
inference time and returns a probability. Typed decision models, which return a probability or a
distribution over named options instead of text, work this way, and so do zero-shot classifiers
built on a natural language inference (NLI) model and readouts of the answer probabilities of a
language model. Such a predictor is used by putting many small questions to one input and combining
the answers in code: a ticket goes to billing when the probability that it asks for a refund exceeds
a threshold, and to first-line support when the probability that it asks for something else does.
Each question is evaluated separately, so nothing forces the answers to logically related questions
to be the probabilities of a single probability measure, and audits that use no labels find that
they are not. On Jev, a commercial typed decision model that we also audit in
Section~\ref{sec:experiments}, the probabilities of ``the label is $X$'' and ``the label is not
$X$'' sum to a value that differs from one by $0.064$ on average, about five times the noise of
repeated calls, and the probabilities of three mutually exclusive labels, asked one at a time, sum
to $1.14$ on average \citep{li2026beyond}; when the same decision is broken into broad categories
and finer labels, the two disagree by a mean total variation of $0.219$ to $0.349$ on three data
sets \citep{joy2026system}. The model's documentation calls it calibrated. For the version we audit,
it also showed a question and its negation, asked separately, whose probabilities summed to $1.19$,
and a question that received $0.22$ when asked for one probability and $0.01$ for the option ``yes''
when asked as a choice between yes and no, and it advised not to hold the model to arithmetic
identities between separate questions; a later revision removed these examples
\citep{typesafe2026jaggedness}.

Routing rules such as the one above mean what they say only if the answers are calibrated, that is,
if the stated probabilities match frequencies, and checking calibration needs labels. In deployment,
labels are costly, late or missing: outcomes arrive weeks later or never, some questions have no
single right answer, and new questions are written faster than labelled evaluation sets can be
built. This paper asks what the law of the answers and the logical structure of the questions,
without any label, can refute about calibration, and whether the refutation can be made
statistically usable: valid at a finite sample, computable, and not broken by rounding or noise.

The direction of the question matters. Coherence does not certify calibration: a predictor whose
answers are always coherent passes every coherence check, however unrelated its answers are to the
label (Counterexample~\ref{cx:coherent}). Nor does incoherence measure marginal miscalibration:
marginal calibration implies coherence only on average (Lemma~\ref{lem:mean-coherence}), answers can
be marginally calibrated and still violate identities on every input
(Counterexample~\ref{cx:pointwise}), and so the documentation's claim of calibration need not
conflict with its advice against arithmetic identities. What we construct are therefore lower
bounds, \emph{certificates} that calibration is at least this bad, and a certificate of zero refutes
nothing.

Which calibration notion a certificate refutes depends on what the comparison of probabilities with
frequencies conditions on, and the notions differ in value, not only in definition. Ask about an
event $A$ and its complement with equal weights, and let the answers $(F_A, F_{\neg A})$ take the
values $(0, \tfrac12)$, $(1, \tfrac12)$, $(0, \tfrac34)$ and $(1, \tfrac14)$ with probability
$\tfrac14$ each. The average answers are coherent, and nothing can be refuted about calibration on
average. Whatever the labels, the answers are miscalibrated by at least $\tfrac1{16}$ when each
answer is conditioned on its own value (marginal calibration) and by at least $\tfrac3{16}$ when it
is conditioned on the whole pair (joint calibration), and no larger bounds follow from the law of
the answers (Theorem~\ref{thm:separation}). The average distance of the answers to the coherent set
is $\tfrac3{16}$, so an audit that reports it as the degree of miscalibration claims three times
what can be proved about marginal calibration. We call a label-free lower bound \emph{sharp} when no
larger one follows from what is observable: it is the least value of the residual over all joint
laws of answers and labels that are compatible with the law of the answers and the logical
structure. This is the sense of a sharp bound in partial identification \citep{manski2003partial};
it has nothing to do with the sharpness of a predictive distribution (its concentration), which
forecasters maximise subject to calibration \citep{gneiting2007probabilistic}.

Section~\ref{sec:identification} identifies these bounds in general. For every restriction on the
information that a bettor's stakes may use, an information-restricted minimax theorem
(Theorem~\ref{thm:minimax}) equates, for a finitely supported law of the answers, the least
calibration residual over the compatible label laws with the largest expected profit of a bettor who
settles every input in its worst world (a Dutch book without outcomes). Constant stakes, stakes on a
question's own answer and stakes on the whole bundle give the sharp bounds on the mean, marginal and
joint residuals, which the example above separates, and a bound on a bundle cannot be attributed to
a fixed question without assumptions on the others (Proposition~\ref{prop:attribution}).

Section~\ref{sec:stability} asks whether a certificate is stable. A certificate used to monitor a
predictor should not jump when the answers move a little, but the sharp marginal bound does: on a
two-point example it jumps from $0$ to above $\tfrac14$ under an arbitrarily small perturbation, and
rounding the answers to two decimals sends it back (Proposition~\ref{prop:discontinuity}). The jump
needs a logical constraint among the questions (Lemma~\ref{lem:no-constraint}), and no label-free
lower bound that is continuous in the Wasserstein distance between laws of answers stays within a
constant factor of the sharp one (Corollary~\ref{cor:closed-form}). Restricting the stakes to
$H$-Lipschitz functions of their own answer gives a bound $L_H$ that is $(1+H)$-Lipschitz in the
Wasserstein distance between laws of answers, that is sharp for a smooth calibration residual
$\Gamma_H$ rather than for the marginal calibration error, and that, for every $H > 0$, is positive
exactly when the sharp marginal bound is (Theorem~\ref{thm:stable}, Corollary~\ref{cor:zero-set}).
The two adjectives of our title therefore meet in $L_H$, which is sharp for $\Gamma_H$ and stable.
For the marginal calibration error $\Theta_w$ they cannot meet (Corollary~\ref{cor:closed-form}).

Section~\ref{sec:computation} makes $L_H$ computable from an i.i.d.\ stream of bundles. Stakes
represented by their values at fixed nodes, learnt by projected online gradient ascent and fixed
before each bundle is observed, give fixed-horizon lower confidence bounds, whose shortfall from
$L_H$ is, with probability at least $1 - \alpha$, bounded by a grid, a learning and a statistical
term (Theorem~\ref{thm:fixed-t}), and the same stakes in a betting process give anytime-valid ones
(Proposition~\ref{prop:anytime}). Section~\ref{sec:experiments} checks the mechanism on synthetic
data and reports a preregistered study of four predictors on two public classification data sets (a
commercial decision model through its API, an open-weight decision model and its base model, and a
zero-shot NLI classifier). All seven confirmatory certificates were positive, from $0.029$ to
$0.194$; the eighth pair of predictor and data set did not meet the preregistered pilot criterion
for collecting confirmation data.

\emph{Contributions.} The techniques (linear programming duality, a minimax theorem, projected
online gradient ascent, the Azuma--Hoeffding inequality and betting) are standard. Our contribution
is to state the calibration notions, what is achievable without labels, the price of stability and
the guarantees of the implementation for one object.
\begin{enumerate}
  \item \emph{What is refutable.} Theorem~\ref{thm:minimax} brings four lines of work under one
  label-free minimax. For finitely supported laws, Corollary~\ref{cor:special} recovers, as special
  cases, results on measures of incoherence \citep{debona2015measuring}, on feasible joint
  posteriors \citep{arieli2021feasible} and on the mutual calibration of forecasters
  \citep{kong2026calibration}, and the stages of Corollary~\ref{cor:stages} are the refutation side
  of labelled hierarchies of calibration \citep{resin2026hierarchies,dwork2021outcome}. The equality
  is finite duality, and label-free partial identification of calibration error is not new
  \citep{kong2026calibration}. The closest work, \citet{andrews2026dutch}, measures the incoherence
  of language models by Dutch books, without a theoretical link to calibration error. The
  joint-stage measure, the distance of a bundle to the coherent set, is a known measure of
  incoherence \citep{debona2015measuring}; we read its expectation, for finitely supported laws, as
  the sharp label-free lower bound on the joint calibration error, below which lie the marginal and
  mean stages. What is new is a single minimax for bundles of different, logically related events
  and, on such bundles, the strict separation of the stages and the impossibility of attributing a
  bound to a given question.
  \item \emph{Stability and its price.} The discontinuity of the sharp marginal bound survives the
  minimisation over the compatible label laws, is produced by the logical constraints and admits no
  stable approximation within a uniform constant factor, while $L_H$ is stable, sharp for $\Gamma_H$
  and, for every $H > 0$, positive for exactly the laws under which marginal calibration is
  refutable without labels in the population. These results have labelled analogues
  \citep{blasiok2023unifying,gopalan2026importance}; ours concern the label-free, partially
  identified quantity.
  \item \emph{Certificates and evidence.} We give fixed-horizon and anytime-valid certificates for
  the same target; for the fixed-horizon certificate with stakes learnt by projected online gradient
  ascent, the shortfall is split into three terms, while the anytime certificate claims no rate. We
  also report a preregistered study on real predictors in which every confirmatory certificate was
  positive.
\end{enumerate}

\emph{What a certificate is about.} Every certificate in this paper concerns unconditional events
about one input and one label with finitely many values; on real data, reading the probabilities of
a choice query this way rests on its options exhausting the labels (Section~\ref{sec:limitations}).
It bounds a total over the bundle with weights fixed in advance that sum to one, so it bounds a
weighted average of per-question errors, not the error of any one question
(Proposition~\ref{prop:attribution}) and not that of all answers pooled into one sequence
(Section~\ref{sec:setup}). It is computed from the answers as the predictor returns them, without
rounding. A predictor whose answers are rounded, by its vendor or by an auditor, is a different
predictor with a different residual, and rounding can move the exact marginal bound discontinuously
(Proposition~\ref{prop:discontinuity}). Finally, each certificate is a lower bound on a named
residual. $L_0$, $L_{\mathrm{marg}}$ and $L_{\mathrm{joint}}$ are sharp for the mean, marginal and
joint residuals ($L_{\mathrm{joint}}$ for finitely supported laws), and $L_{\mathrm{joint}}$ does
not bound the marginal calibration error $\Theta_w$ (Counterexample~\ref{cx:pointwise}). $L_H$ and
its computable version $L_{H,G}$ are lower bounds on $\Theta_w$, and so, at their stated levels, are
the confidence bounds on $L_{H,G}$; $L_H$ is sharp for the smooth residual $\Gamma_H$, not for
$\Theta_w$. Values of these bounds computed on the empirical law of a sample are not certificates
(Appendix~\ref{app:descriptive}).

Section~\ref{sec:setup} sets up bundles, calibration residuals and Dutch books without outcomes.
Sections~\ref{sec:identification}--\ref{sec:computation} give the results, with the proofs in
Appendices~\ref{app:proofs-identification}--\ref{app:proofs-computation}, and
Section~\ref{sec:experiments} the experiments. Section~\ref{sec:related} discusses related work,
Section~\ref{sec:limitations} the limitations, and Section~\ref{sec:conclusion} concludes.

\section{Setup}
\label{sec:setup}

\emph{Bundles and possible worlds.} A query-conditioned probabilistic predictor receives an input
together with a query and returns a probability. For one input we ask $m$ event queries and collect
the answers in a \emph{bundle} $F = (F_1, \dots, F_m) \in [0,1]^m$. Each queried event $B_i$ is a
function of a label $Y$ that takes finitely many values and is never observed. A \emph{possible
world} $\ell \in \mathcal{W}$ is a value of the label, with indicator vector $z^\ell =
(\mathbf{1}\{\ell \in B_i\})_{i \le m} \in \{0,1\}^m$; the realised indicator vector is $Z = z^Y$. A
\emph{choice query}, which returns a distribution over a set of options, enters the bundle as one
component per option. The \emph{coherent polytope}
\[
  \mathcal{C} = \operatorname{conv}\{z^\ell : \ell \in \mathcal{W}\}
\]
is the set of vectors $(\mu(B_1), \dots, \mu(B_m))$ obtained from probability measures $\mu$ on
$\mathcal{W}$, so a bundle is coherent exactly when it lies in $\mathcal{C}$.

Without labels one observes the law $P_F$ of the bundle and the logical structure $\{z^\ell\}$,
nothing else. We write $\mathcal{P}(P_F)$ for the set of joint laws $P$ of $(F, Z)$ on $[0,1]^m
\times \{z^\ell : \ell \in \mathcal{W}\}$ whose first marginal is $P_F$; the conditional law of $Z$
given $F$ is unrestricted. Each component carries a weight $w_i > 0$, with $\sum_i w_i = 1$, fixed
in advance with the bundle according to the role of its query: a choice query, an \emph{atom} $\{Y =
k\}$, a \emph{complement} $\{Y \ne k\}$ or a \emph{union} of labels. When certificates that use
fewer queries are compared with one another, each is computed on the same bundle and weights, with
zero stakes on the components it does not use, so that all of them bound the same residual. We use
the weighted $\ell_1$ distance $\operatorname{dist}^w_1(v, \mathcal{C}) = \min_{c \in \mathcal{C}}
\sum_i w_i \lvert v_i - c_i \rvert$ and the weighted Wasserstein distance $W_{1,w}(P, Q) =
\inf_{\gamma \in \Pi(P, Q)} \mathbb{E}_\gamma \sum_i w_i \lvert F_i - F_i' \rvert$ between laws of
bundles.

\emph{Semantics.} Every component asks an unconditional event about the same input and the same
label. We assume that the options of a choice query exhaust the possible labels and that a union is
asked as the explicit disjunction of its labels. These assumptions fix the semantics that the
logical structure encodes; a relation entered wrongly into $\mathcal{C}$ makes a statistically valid
certificate refute the wrong thing, while weakening a relation enlarges $\mathcal{C}$ and can only
lower the population bounds defined below.

\emph{Calibration residuals.} Calibration compares stated probabilities with frequencies, and the
comparison depends on what it conditions on. For sub-$\sigma$-algebras $\mathcal{H}_i \subseteq
\sigma(F)$ we define the \emph{information-restricted calibration residual}
\[
  \Gamma_{\mathcal{H}}(P) = \sum_{i=1}^m w_i \, \mathbb{E}\bigl\lvert \mathbb{E}[F_i - Z_i \mid
  \mathcal{H}_i] \bigr\rvert .
\]
The trivial $\sigma$-algebra gives the \emph{mean} residual $\sum_i w_i \lvert \mathbb{E}(F_i - Z_i)
\rvert$. The choice $\mathcal{H}_i = \sigma(F_i)$ gives the \emph{marginal} calibration error
\[
  \Theta_w(P) = \sum_{i=1}^m w_i \, \mathrm{ECE}_1(F_i), \qquad
  \mathrm{ECE}_1(F_i) = \mathbb{E}\bigl\lvert F_i - \Pr(Z_i = 1 \mid F_i) \bigr\rvert ,
\]
in which every component is conditioned on its own value.\footnote{Terminology varies. Our marginal
stage conditions each component on its own value; for the classes of a classifier this is class-wise
calibration, which \citet{kumar2019verified} and \citet{marx2023calibration} also call marginal
calibration. In probabilistic forecasting, marginal calibration is the agreement of the average
predictive distribution with the marginal law of the outcome
\citep{gneiting2007probabilistic,resin2026hierarchies}; for an event it is our mean stage.
\citet{andrews2026dutch} use marginal and joint for single and compound events.} The choice
$\mathcal{H}_i = \sigma(F)$ gives the \emph{joint} calibration error $\mathbb{E} \sum_i w_i \lvert
F_i - \Pr(Z_i = 1 \mid F) \rvert$, in which every component is conditioned on the whole bundle.
Because conditional expectation contracts, the three residuals, which we call the mean, marginal and
joint \emph{stages}, are ordered: mean $\le$ marginal $\le$ joint for every $P$. $\Theta_w$ is a
weighted average of per-component errors with fixed weights, not the expected calibration error of
all answers pooled into one sequence. If the pooled sequence draws the component of each answer
independently with probabilities $w_i$, Jensen's inequality makes its error at most $\Theta_w$, so a
lower bound on $\Theta_w$ is not one on the pooled error; with any other pooling the two are not
ordered.

Conditioning on an exact value makes $\Theta_w$ jump under arbitrarily small perturbations of the
answers. With the class $\mathcal{U}_H$ of stakes $u = (u_1, \dots, u_m)$ whose components $u_i
\colon [0,1] \to [-1,1]$ are $H$-Lipschitz, the \emph{smooth residual} is
\[
  \Gamma_H(P) = \sup_{u \in \mathcal{U}_H} \mathbb{E} \sum_{i=1}^m w_i \, u_i(F_i)\,(F_i - Z_i),
\]
the weight family of smooth calibration error \citep{blasiok2023unifying} applied component by
component. $\Gamma_H$ is nondecreasing in $H$, $\Gamma_0$ is the mean residual and $\Gamma_H \le
\Theta_w$.

\emph{Sharp label-free lower bounds.} A \emph{label-free lower bound} on a residual $\Gamma$ is a
functional $L$ of $P_F$, the logical structure and the weights with $L(P_F) \le \Gamma(P)$ for every
$P \in \mathcal{P}(P_F)$. It is \emph{sharp} when $L(P_F) = \inf_{P \in \mathcal{P}(P_F)}
\Gamma(P)$: no larger lower bound follows from what is observable. A certificate is such a lower
bound, or a finite-sample lower confidence bound for one; it can only show that calibration is at
least this bad.

\emph{Dutch books without outcomes.} A bettor who stakes $u_i$ on event $B_i$ at the price $F_i$
earns, as the seller, $\sum_i w_i u_i (F_i - Z_i)$. The outcome is not observed, so the bettor
counts the worst world:
\[
  d^w_u(F) = \min_{\ell \in \mathcal{W}} \sum_{i=1}^m w_i \, u_i(F) \, (F_i - z^\ell_i) \in [-1, 1]
  \qquad (\lvert u_i \rvert \le 1).
\]
The information a stake may use decides which calibration notion it tests. We write
$\mathcal{U}_{\mathcal{H}}$ for the stakes whose component $u_i$ is $\mathcal{H}_i$-measurable with
$\lvert u_i \rvert \le 1$; constant stakes ($\mathcal{U}_0$, which is also $\mathcal{U}_H$ with $H =
0$), stakes on the own value $u_i(F_i)$ ($\mathcal{U}_{\mathrm{marg}}$) and stakes on the whole
bundle $u(F)$ ($\mathcal{U}_{\mathrm{joint}}$) correspond to the mean, marginal and joint residuals.
Negative values of $d^w_u(F)$ are averaged as they are: clipping each value at zero would overstate
the expected profit, so only a final lower confidence bound is clipped.

\section{What Is Refutable Without Labels}
\label{sec:identification}

This section asks what the law of the bundle and its logical structure can refute about each
calibration residual. Section~\ref{sec:calibration-coherence} ties calibration to coherence and
marks where the tie ends. Section~\ref{sec:identification-results} identifies, for finitely
supported laws, the sharp label-free lower bound on every information-restricted residual, separates
the mean, marginal and joint stages on a small example, and shows that a bound on the bundle cannot
be attributed to a fixed question. The proofs are in Appendix~\ref{app:proofs-identification}.

\subsection{Calibration and Coherence}
\label{sec:calibration-coherence}

If every answer is marginally calibrated, the average answers are the frequencies of the queried
events, and these frequencies come from one probability measure.

\begin{lemma}[Marginal calibration implies coherence on average]
\label{lem:mean-coherence}
If $\Pr(Z_i = 1 \mid F_i) = F_i$ almost surely for every $i$, then $\mathbb{E}F \in \mathcal{C}$; in
particular every linear equality or inequality satisfied by all probability measures holds for
$(\mathbb{E}F_1, \dots, \mathbb{E}F_m)$. The converse fails.
\end{lemma}

Marginal calibration and coherence therefore meet only on average. A marginally calibrated predictor
can answer a question and its negation with probabilities that do not sum to one on an individual
input, while an identity or an inequality that fails on average, such as $\mathbb{E}F_1 >
\mathbb{E}F_2$ for $B_1 \subseteq B_2$, refutes marginal calibration. The converse fails twice over:
a predictor whose average answers are coherent can be miscalibrated under its true law
(Counterexample~\ref{cx:coherent}), and Theorem~\ref{thm:separation} gives a law of bundles with a
coherent average for which no compatible label law makes the answers marginally calibrated. In the
other direction, pointwise incoherence does not by itself imply marginal miscalibration.

\begin{counterexample}[Pointwise incoherence is not marginal miscalibration]
\label{cx:pointwise}
Let $A$ be an event and let the bundle $(F_A, F_{\neg A})$ take the values $(0, \tfrac12)$ and $(1,
\tfrac12)$ with probability $\tfrac12$ each, $A$ occurring exactly when $F_A = 1$. Then $\Theta_w =
0$, while $\mathbb{E}\operatorname{dist}^w_1(F, \mathcal{C}) = \tfrac14$ with equal weights.
\end{counterexample}

Counting coherence violations, or averaging the distance of each bundle to $\mathcal{C}$, does not
measure marginal calibration error, and an audit that reports either as the degree to which marginal
calibration is broken is refuted by this example. The pointwise distance bounds the joint
calibration error (Corollary~\ref{cor:stages}); here the joint calibration error is $\tfrac14$ as
well.

\begin{counterexample}[Coherent but miscalibrated]
\label{cx:coherent}
A predictor whose bundle always lies in $\mathcal{C}$ violates no coherence check, however unrelated
its answers are to the label. Every population certificate of this paper is then zero while
$\Theta_w$ can be large. A certificate of zero therefore means ``not refuted'', never
``calibrated''.
\end{counterexample}

Every result below is subject to this limit. A label-free check sees only the distortions that push
answers out of $\mathcal{C}$; an error in a coherent belief, however large, leaves no trace in $P_F$
and the logical structure. A positive certificate refutes, on its bundle, the calibration notion
whose residual it bounds (a finite-sample one at its level).

\subsection{Identification}
\label{sec:identification-results}

The sharp label-free lower bound on a residual $\Gamma_{\mathcal{H}}$ is, by definition, its infimum
over the label laws compatible with $P_F$: the smallest error under the label laws that a label-free
observer cannot rule out. A bettor gives lower bounds on it. If stakes that use only the information
$\mathcal{H}_i$ on $B_i$ make a positive expected profit when every input is settled in its worst
world, the answers cannot be calibrated in the sense of $\mathcal{H}$: under every compatible law
that expected profit is at most $\Gamma_{\mathcal{H}}$ (Lemma~\ref{lem:weak-duality}). The next
theorem gives the converse: a calibration violation that no admissible bet extracts is absent under
some compatible label law.

\begin{theorem}[Information-restricted minimax]
\label{thm:minimax}
If $P_F$ has finite support, then
\[
  \inf_{P \in \mathcal{P}(P_F)} \Gamma_{\mathcal{H}}(P) = \sup_{u \in \mathcal{U}_{\mathcal{H}}}
  \mathbb{E}\bigl[d^w_u(F)\bigr],
\]
both sides are attained, and they are the optimal values of a finite linear program and its dual.
\end{theorem}

The primal chooses, with $P_F$ held fixed, the label law under which the predictor is as calibrated
as it can be; the dual chooses stakes that use only the permitted information and hedge against
every world. The equality is the duality of a finite linear program. For one event put to several
agents, \citet{arieli2021feasible} characterise feasible joint posteriors by a no-trade condition
and note that, for finitely supported laws, their characterisation reduces to Farkas' lemma; the
same finite duality gives the value of the marginal stage (Corollary~\ref{cor:special}(a)). What the
theorem adds is a single minimax for every information restriction on a bundle of different,
logically related events, so that the mean, marginal and joint stages, and any stage between them,
are one object. An attaining law shows that under some label law consistent with the observations
the predictor is calibrated up to the bound; it does not say that the predictor's actual errors
resemble those under that law. At the marginal stage a stake on $B_i$ must be the same on all inputs
at which $F_i$ takes the same value, and this coupling across inputs is what separates
$L_{\mathrm{marg}}$ from the pointwise distance.

\begin{corollary}[Three stages]
\label{cor:stages}
Taking $\mathcal{H}_i$ trivial, $\sigma(F_i)$ and $\sigma(F)$ in Theorem~\ref{thm:minimax} gives the
sharp label-free lower bounds
\[
  L_0 = \operatorname{dist}^w_1(\mathbb{E}F, \mathcal{C}), \qquad
  L_{\mathrm{marg}}, \qquad
  L_{\mathrm{joint}} = \mathbb{E} \operatorname{dist}^w_1(F, \mathcal{C})
\]
on the mean, marginal and joint residuals, where $L_{\mathrm{marg}}$ is the value of the linear
program; and $L_0 \le L_{\mathrm{marg}} \le L_{\mathrm{joint}}$. In particular $L_{\mathrm{marg}}
\le \Theta_w(P)$ for every $P \in \mathcal{P}(P_F)$.
\end{corollary}

The three bounds answer three questions. $L_0$ is the best test by the supporting hyperplanes of
$\mathcal{C}$, inequalities as well as identities: if $B_1 \subseteq B_2$ and $\mathbb{E}F = (0.8,
0.2)$, no identity is violated but the inclusion is, and $L_0 = 0.3$ with equal weights.
$L_{\mathrm{joint}}$ needs no learning; it averages the distance of each bundle to $\mathcal{C}$.
$L_{\mathrm{marg}}$ lies between them and needs the linear program, because its stakes are shared
across inputs. A bound grows with the information its stakes may use, and so does the notion it
bounds: a larger bound refutes a stronger notion, not the same notion more strongly. The stages
order calibration notions by the $\sigma$-algebra they condition on, as labelled hierarchies of
calibration do \citep{resin2026hierarchies}, and the information available to a stake plays the role
of the access of a distinguisher in outcome indistinguishability \citep{dwork2021outcome};
Theorem~\ref{thm:minimax} is the label-free, refutation side of that correspondence.

Theorem~\ref{thm:minimax} also contains, as special cases, results and quantities from three other
lines of work: the feasibility of joint posterior beliefs, measures of incoherence, and the mutual
calibration of forecasters.

\begin{corollary}[Special cases]
\label{cor:special}
Let $P_F$ have finite support.
\begin{enumerate}
  \item[(a)] \emph{One event asked $m$ times.} If $B_1 = \dots = B_m = A$ with $A$ neither sure nor
  impossible, then
  \[
    L_{\mathrm{marg}} = \sup_{a} \, \mathbb{E}\Bigl[\sum_{i=1}^m a_i(F_i)\,F_i - \max\Bigl\{0,
    \sum_{i=1}^m a_i(F_i)\Bigr\}\Bigr]
  \]
  over measurable $a_i \colon [0,1] \to [-w_i, w_i]$. Hence $L_{\mathrm{marg}} = 0$ if and only if
  $F \equiv 0^m$, $F \equiv 1^m$, or $P_F$ is the law of the posterior beliefs about $A$ of $m$
  agents with a common prior $p \in (0,1)$: the no-trade characterisation of feasible joint
  posteriors \citep[Theorem~3]{arieli2021feasible}; for $m = 2$, Arieli et al.\ state an equivalent
  characterisation due to \citet{dawid1995coherent}.
  \item[(b)] \emph{The joint stage, pointwise.} For every $v \in [0,1]^m$,
  \[
    \operatorname{dist}^w_1(v, \mathcal{C}) = \max_{u \in [-1,1]^m} \min_{\ell \in \mathcal{W}}
    \sum_{i=1}^m w_i u_i (v_i - z^\ell_i),
  \]
  the identity between distance-based and Dutch-book measures of incoherence
  \citep{debona2015measuring}; $L_{\mathrm{joint}}$ is its expectation.
  \item[(c)] \emph{Two forecasters of one label.} Let $Q_1, Q_2 \in \Delta_K$ forecast a label $Y
  \in \{1, \dots, K\}$, written as a one-hot vector. Form the bundle $F = (Q_1, Q_2)$ of the $2K$
  events $\{Y = k\}$ with weights $1/(2K)$, and give coordinate $(j, k)$ the information
  $\sigma(Q_j)$. Then $\inf_{P \in \mathcal{P}(P_F)} \Gamma_{\mathcal{H}}(P) = \mathrm{MinECE}/K$,
  where
  \[
    \mathrm{MinECE} = \tfrac12 \inf_{P \in \mathcal{P}(P_F)} \mathbb{E}\bigl[\lVert Q_1 -
    \mathbb{E}[Y \mid Q_1] \rVert_1 + \lVert Q_2 - \mathbb{E}[Y \mid Q_2] \rVert_1\bigr]
  \]
  is the minimal calibration error of \citet{kong2026calibration}; so $\mathrm{MinECE} = K \sup_u
  \mathbb{E}[d^w_u(F)]$ over stakes $u_{(j,k)} = u_{jk}(Q_j) \in [-1, 1]$.
\end{enumerate}
\end{corollary}

Case~(a) turns the no-trade condition into a value: $L_{\mathrm{marg}}$ is the least marginal
calibration error that a label law compatible with forecasts of one event must leave, and its zero
set is the feasibility characterisation, apart from the two degenerate laws, which pass the no-trade
test without an interior prior. Case~(b) is the identity behind distance-based measures of
incoherence; what Corollary~\ref{cor:stages} adds is that its expectation is the sharp label-free
lower bound on the joint calibration error, and Counterexample~\ref{cx:pointwise} shows that it does
not bound the marginal one. \citet[Appendix~A]{andrews2026dutch} bound the stakes in $\ell_1$
instead and obtain the distance in the supremum norm, the dual of $\ell_1$. We bound each stake
separately, and then, under a fixed label law, the best expected profit against the realised
outcomes separates across components: the stakes $u_i = \operatorname{sign} \mathbb{E}[F_i - Z_i
\mid \mathcal{H}_i]$ make the second inequality of Lemma~\ref{lem:weak-duality} an equality, so that
this profit is the weighted sum $\Gamma_{\mathcal{H}}$ of the per-component residuals, each
conditioned on its own information. In case~(c) a coordinate is conditioned on the whole forecast
vector of its forecaster. This information contains the marginal one and is contained in the joint
one, so $L_{\mathrm{marg}} \le \mathrm{MinECE}/K \le L_{\mathrm{joint}}$; for $K = 2$ it is the
marginal stage. \citet{kong2026calibration} bound the arbitrage of zero-mean trades, each depending
only on its forecaster's own forecast and with payoffs in $[-1, 1]$, between $\mathrm{MinECE}$ and
$2\,\mathrm{MinECE}$; bounding the stakes instead of the payoffs, as $d^w_u$ does, turns the bracket
into an equality.

The next example, small enough to check by hand, separates the three stages. It mixes the two inputs
of Counterexample~\ref{cx:pointwise}, which a marginally calibrated law can produce, with two inputs
that no such law can.

\begin{theorem}[Four-point separation]
\label{thm:separation}
For an event $A$ and its complement with equal weights, let $(F_A, F_{\neg A})$ take the values $(0,
\tfrac12)$, $(1, \tfrac12)$, $(0, \tfrac34)$ and $(1, \tfrac14)$ with probability $\tfrac14$ each.
Then $0 = L_0 < L_{\mathrm{marg}} = \tfrac1{16} < L_{\mathrm{joint}} = \tfrac3{16}$.
\end{theorem}

Table~\ref{tab:four-point} lists a label law that attains $L_{\mathrm{marg}}$ and stakes that earn
it, so the separation can be checked without a solver (Figure~\ref{fig:four-point}). On these
answers the average distance to $\mathcal{C}$, $\tfrac3{16}$, is three times the marginal
calibration error that can be proved without labels, $\tfrac1{16}$, and an audit's conclusion
depends on which of the two it reports. The marginal bound can also be zero while the joint one is
not. On a fair die, let a choice query over the six faces return $0.83$ for a uniformly drawn face
and $0.034$ for each other face, beside atom queries $\{Y = k\}$ that always answer $\tfrac16$. The
choice query is overconfident, yet $L_{\mathrm{marg}} = 0$: under the label law in which the drawn
face comes up with probability $0.83$ and each other face with probability $0.034$, every component
is marginally calibrated. $L_{\mathrm{joint}} > 0$ because the two queries disagree on the same
event at every input. That the roll does not depend on the drawn face is information beyond $P_F$
and the logical structure.

A certificate bounds an error summed over the bundle. In practice one often cares about one
question, the target, and asks the others only to check it.

\begin{proposition}[No attribution to a fixed target]
\label{prop:attribution}
Split the components into targets $J$ and auxiliaries $O = \{1, \dots, m\} \setminus J$, and let
$\mathcal{P}_J(P_{F_J})$ be the analogue of $\mathcal{P}(P_F)$ for the bundle $F_J$ alone, whose
worlds are the restrictions $z^\ell_J$. With no restriction on the auxiliary answers,
\[
  \inf_{P \in \mathcal{P}(P_F)} \sum_{i \in J} w_i \, \mathrm{ECE}_1(F_i) =
  \inf_{P \in \mathcal{P}_J(P_{F_J})} \sum_{i \in J} w_i \, \mathrm{ECE}_1(F_i) :
\]
observing $F_O$ does not lower-bound the targets' error any further. If $J = \{i\}$ and component
$i$ asks an event that is neither sure nor impossible, both sides are $0$.
\end{proposition}

When the answers to ``$A$?'' and ``not $A$?'' sum to $1.19$ on average, so that $L_0 > 0$, the
bundle is refuted, but nothing in $P_F$ says which answer is wrong: a label law under which the
target is calibrated and the auxiliary answer carries all the error is always compatible. What a
lower bound $L \le \Theta_w$ says about single questions is that $\mathrm{ECE}_1(F_i) \ge L$ for at
least one component, since $\Theta_w$ is a weighted average. This covers $L_0$, $L_{\mathrm{marg}}$,
$L_H$ and $L_{H,G}$ and, on the events where they hold, finite-sample lower confidence bounds on
them; it does not cover $L_{\mathrm{joint}}$ or another bound from Theorem~\ref{thm:minimax} that
conditions on more than a component's own value, such as $\mathrm{MinECE}/K$ in
Corollary~\ref{cor:special}(c), which can be positive while every $\mathrm{ECE}_1(F_i)$ is zero
under the true law (Appendix~\ref{app:proofs-identification}). Attributing the bound to a given
target needs explicit assumptions on the auxiliaries, such as bounds on their calibration errors,
which we do not pursue. For the same reason, certificates of parts of a bundle do not add up. With
weights $\tfrac12$ each, the answer $0.8$ to ``$A$?'' alone, or to ``not $A$?'' alone, refutes
nothing, while the two answers together give $L_0 = 0.3$. Nor is the certificate a submodular
function of the set of queries, so choosing queries greedily comes with no guarantee.

\section{Stability and Its Price}
\label{sec:stability}

A certificate used to monitor a predictor should not jump when the answers move a little, by
rounding, by noise or between two runs of the same model. This section shows that the sharp bound on
the marginal calibration error can jump, traces the jump to the logical constraints of the bundle,
gives the price of stability in closed form, and shows that, for every law of the answers, the
stable bound $L_H$ is the sharp label-free lower bound on the smooth residual $\Gamma_H$ and, for
every $H > 0$, is positive exactly when the sharp bound on $\Theta_w$ is. The proofs are in
Appendix~\ref{app:proofs-stability}.

For a general $P_F$ we set
\[
  L_{\mathrm{marg}}(P_F) = \inf_{P \in \mathcal{P}(P_F)} \Theta_w(P) , \qquad
  L_H(P_F) = \sup_{u \in \mathcal{U}_H} \mathbb{E}[d^w_u(F)] ;
\]
$L_0$ and $L_{\mathrm{joint}}$ keep the formulas of Corollary~\ref{cor:stages}. Since a stake in
$\mathcal{U}_H$ is a function of its own component, Lemma~\ref{lem:weak-duality} gives $L_H(P_F) \le
\Gamma_H(P) \le \Theta_w(P)$ for every $P \in \mathcal{P}(P_F)$, so $L_H$ is a label-free lower
bound on both residuals and $L_H \le L_{\mathrm{marg}}$.

\begin{proposition}[The exact marginal certificate is discontinuous]
\label{prop:discontinuity}
For $0 \le \varepsilon \le \tfrac12$ let $F^\varepsilon$ take $(0, \tfrac12 - \varepsilon)$ and $(1,
\tfrac12 + \varepsilon)$ with probability $\tfrac12$ each, as a bundle on an event and its
complement with equal weights. Then $L_{\mathrm{marg}}(P_{F^0}) = 0$ while, for $\varepsilon > 0$,
\[
  L_{\mathrm{marg}}(P_{F^\varepsilon}) = L_{\mathrm{joint}}(P_{F^\varepsilon}) = \tfrac14 +
  \tfrac{\varepsilon}2 , \qquad W_{1,w}(P_{F^\varepsilon}, P_{F^0}) = \tfrac{\varepsilon}2 .
\]
\end{proposition}

At $\varepsilon = 0$ this is Counterexample~\ref{cx:pointwise}: $F_{\neg A}$ is $\tfrac12$ on both
inputs, and a label law can pool the two inputs and calibrate it given its own value. For
$\varepsilon > 0$ the value of each component tells the two inputs apart, so conditioning a
component on its own value is conditioning on the whole bundle. Under every compatible label law the
marginal and joint residuals then coincide, and the marginal bound jumps to the joint one. The
second component is the predictor of \citet[Lemma~4.8]{blasiok2023unifying}, which answers $\tfrac12
- \varepsilon$ and $\tfrac12 + \varepsilon$ on two equally likely inputs with different labels and
shows that the expected calibration error is discontinuous at a calibrated predictor. There the
labels are observed. Here the jump survives the minimisation over every label law compatible with
$P_F$. It is a property of conditioning on exact values, not of the linear program: for $\varepsilon
< 0.005$, rounding the answers to two decimals maps $F^\varepsilon$ to $F^0$, and the exact marginal
bound drops from above $\tfrac14$ to $0$ (Figure~\ref{fig:two-point}).

With observed labels, the answers to a single question already show such a jump. Without labels they
cannot: the next lemma shows that a single question that is neither sure nor impossible is never
refuted, and that the jump needs a logical constraint among the queried events.

\begin{lemma}[No logical constraint, no jump]
\label{lem:no-constraint}
Let $S$ be the set of components whose event is sure or impossible, $z^\ell_i = c_i$ for every
world, and suppose the remaining events are logically independent: $\{(z^\ell_i)_{i \notin S} : \ell
\in \mathcal{W}\} = \{0,1\}^{\{1, \dots, m\} \setminus S}$. Then for every $P_F$ and every $H \ge
0$,
\[
  L_0 = L_{\mathrm{marg}} = L_{\mathrm{joint}} = L_H = \sum_{i \in S} w_i \, \mathbb{E}\lvert F_i -
  c_i \rvert ,
\]
which is $1$-Lipschitz in $W_{1,w}$. In particular a bundle of one query that is neither sure nor
impossible is never refuted, and a discontinuity such as that of
Proposition~\ref{prop:discontinuity} requires logical constraints among the nontrivial events.
\end{lemma}

Without such constraints, the label law that draws each nontrivial outcome as an independent coin
with the stated probability calibrates every nontrivial component, even jointly; for one prediction
this is the observation of \citet[Section~2]{gopalan2026importance} that drawing the label as a coin
with the stated probability yields a calibrated joint law of prediction and label. What remains is
the error of the answers to sure or impossible events, which no label law can remove and which moves
continuously with $P_F$. In Proposition~\ref{prop:discontinuity} the constraint that exactly one of
$A$ and $\neg A$ occurs ties the two components to one outcome: for $\varepsilon = 0$ some label law
still makes both marginally calibrated, and for $\varepsilon > 0$ none does. The jump of
$L_{\mathrm{marg}}$ is thus produced by the coherence constraints: without them none of the bounds
jumps, and the single constraint of a complement pair suffices.

\begin{corollary}[Closed form and the impossibility of a uniform constant factor]
\label{cor:closed-form}
In the setting of Proposition~\ref{prop:discontinuity}, for every $H \ge 0$,
\[
  L_H(P_{F^\varepsilon}) = \bigl(\tfrac14 + \tfrac{\varepsilon}2\bigr) \min\{1, H\varepsilon\} .
\]
Hence $L_H / L_{\mathrm{marg}} = \min\{1, H\varepsilon\} \to 0$ as $\varepsilon \to 0$, and no
distribution-free constant $c_H > 0$ gives $L_H \ge c_H L_{\mathrm{marg}}$. More generally, no lower
bound that vanishes at $P_{F^0}$ and is continuous in $W_{1,w}$ admits such a constant.
\end{corollary}

On the two-point example a bettor profits only by staking differently on the two values $\tfrac12
\pm \varepsilon$ of $F_{\neg A}$. An $H$-Lipschitz stake can make the two stakes differ by at most
$2H\varepsilon$, against $2$ for an unrestricted one, so the profit shrinks by the factor $\min\{1,
H\varepsilon\}$. This is the price of stability, and no label-free lower bound on $\Theta_w$ that is
continuous in $W_{1,w}$ avoids it. A label-free lower bound on $\Theta_w$ is at most
$L_{\mathrm{marg}}$, so a nonnegative one vanishes at $P_{F^0}$; if it is continuous there, it is
small at $P_{F^\varepsilon}$ for small $\varepsilon$, where $L_{\mathrm{marg}} > \tfrac14$. With
observed labels, \citet[Lemmas~4.3 and~4.8]{blasiok2023unifying} together give the counterpart: a
calibration measure that vanishes on calibrated predictors and is Lipschitz in the predictor is not
bounded below by a constant multiple of the expected calibration error.
Corollary~\ref{cor:closed-form} gives the price in closed form for the label-free bound. On a finite
grid the price is bounded. If any two distinct values taken by the same component differ by at least
$\delta > 0$, then
\begin{equation}
  \min\{1, H\delta/2\} \, L_{\mathrm{marg}} \le L_H \le L_{\mathrm{marg}} ,
  \label{eq:finite-grid}
\end{equation}
so $L_H = L_{\mathrm{marg}}$ once $H \ge 2/\delta$ (Appendix~\ref{app:proofs-stability}). For
answers given to two decimals, $\delta \ge 0.01$, and $H = 200$ recovers the exact marginal bound,
at the stability constant $1 + H = 201$ of the next theorem.

\begin{theorem}[Stable certificate]
\label{thm:stable}
Let $F \in [0,1]^m$ have any law $P_F$ and fix $0 \le H < \infty$. Then
\[
  \inf_{P \in \mathcal{P}(P_F)} \Gamma_H(P) = \sup_{u \in \mathcal{U}_H} \mathbb{E}\bigl[d^w_u(F)
  \bigr] = L_H(P_F),
\]
the infimum is attained, and for two laws of bundles with the same logical structure, weights and
$H$,
\[
  \bigl\lvert L_H(P_F) - L_H(P_{F'}) \bigr\rvert \le (1 + H) \, W_{1,w}(P_F, P_{F'}) .
\]
\end{theorem}

Theorem~\ref{thm:minimax} rests on a finite linear program, which continuous answers do not give.
Restricting the stakes to $H$-Lipschitz functions keeps the minimax for every law of the answers, by
a compactness argument, and makes the bound move with the answers: a stake that cannot change
quickly earns about the same on nearby answers. For finitely supported laws, $L_H$ is again the
value of a finite linear program, in the values of the stakes at the support points under Lipschitz
constraints between neighbouring values (Lemma~\ref{lem:extension}); this is how we compute it. At
$H = 0$ the stakes are constants and $\Gamma_0$ is the mean residual, whose infimum over the
compatible label laws is $\operatorname{dist}^w_1(\mathbb{E}F, \mathcal{C})$ for every law
(Corollary~\ref{cor:stages} and its proof). So the theorem identifies $L_0$ with the value of the
best constant stake for every law, as Theorem~\ref{thm:minimax} did for finitely supported ones, and
the two meanings of $L_0$ (the formula and $L_H$ at $H = 0$) agree.

Two cautions. First, $L_H$ is sharp for the smooth residual $\Gamma_H$, not for $\Theta_w$: $L_H =
\inf_P \Gamma_H(P) \le \inf_P \Theta_w(P) = L_{\mathrm{marg}}$, with the gap of
Corollary~\ref{cor:closed-form}; this is the sense of ``sharp'' in our title. Second, the theorem
makes the population value stable. It does not say that a finite-sample bound, whose stakes are
learnt from the data, moves by at most $(1 + H)\,W_{1,w}$. With observed labels, the smooth
calibration error of one predictor, whose stakes are $1$-Lipschitz ($m = 1$, $H = 1$ in our
notation), is within a factor of two of its lower distance from calibration
\citep[Theorem~7.3]{blasiok2023unifying} and of its earth mover's distance to the set of calibrated
joint laws of prediction and label \citep[Lemma~2.5]{gopalan2026importance}; we do not carry that
factor to other $H$. The theorem instead identifies, without labels and for every $H$, the least
smooth residual over the compatible label laws; the equality itself follows from a minimax theorem.

\begin{corollary}[Stabilisation keeps what is refutable]
\label{cor:zero-set}
Let $F \in [0,1]^m$ have any law $P_F$ and fix $0 < H < \infty$. Then
\[
  L_H(P_F) = 0 \iff L_{\mathrm{marg}}(P_F) = 0 \iff \Theta_w(P) = 0 \text{ for some } P \in
  \mathcal{P}(P_F) .
\]
\end{corollary}

Corollary~\ref{cor:closed-form} says that stabilisation shrinks a bound;
Corollary~\ref{cor:zero-set} says that, for $H > 0$, it never shrinks a positive sharp bound on
$\Theta_w$ to zero. Lipschitz functions are dense among the continuous functions, so under a label
law at which no Lipschitz stake on a component's own value profits, no stake on that value profits
either. Stabilisation gives up a uniform constant factor but keeps, for every fixed $H > 0$, the set
of laws of the answers under which marginal calibration is refuted without labels in the population.
It says nothing about the sample size a refutation needs, which can be large when $L_H$ is small.
The result concerns the class $\mathcal{U}_H$ and does not give the finite class of
Section~\ref{sec:computation} the same zero set; it excludes $H = 0$; and the blind spot of
Counterexample~\ref{cx:coherent} remains, since $L_{\mathrm{marg}} = 0$ is not evidence that the
predictor is marginally calibrated under its true law. With observed labels, the smooth calibration
error of one predictor ($H = 1$) is likewise positive for every miscalibrated predictor, which
\citet[Table~1]{blasiok2023unifying} list as its soundness.

\section{Computable Certificates}
\label{sec:computation}

Theorem~\ref{thm:stable} identifies $L_H$ as a supremum over an infinite-dimensional class of stakes
under a law $P_F$ that is not known. This section makes the bound computable from a stream of
bundles, one per \emph{item}: an input drawn together with its unobserved label. The stakes are
represented by their values at fixed nodes, which lowers the population value by at most $H\Delta/2$
and keeps its stability (Lemma~\ref{lem:grid}); they are learnt online by projected gradient ascent
(Proposition~\ref{prop:ogd}); and since each item is scored with a stake chosen before the item is
seen, the scores give lower confidence bounds on $L_{H,G}$, and hence on $\Gamma_H$ and $\Theta_w$,
at a fixed horizon (Theorem~\ref{thm:fixed-t}) and at all times at once
(Proposition~\ref{prop:anytime}). The answers enter the profits unrounded. Two natural alternatives
are not Lipschitz in the answer, so Theorem~\ref{thm:stable} does not cover them: a stake of zero at
values not seen before, and a stake that is constant within bins of values, which jumps when an
answer crosses a bin edge. The proofs are in Appendix~\ref{app:proofs-computation}.

Fix nodes $0 = t_1 < \dots < t_G = 1$ with largest gap $\Delta$, and let $\phi_1, \dots, \phi_G$ be
the piecewise-linear interpolation functions on them. A matrix $\theta \in \mathbb{R}^{m \times G}$
defines the stakes $u_{\theta,i}(x) = \sum_j \theta_{ij} \phi_j(x)$, and
\begin{gather*}
  K_{H,G} = \bigl\{\theta \in [-1,1]^{m \times G} : \lvert \theta_{i,j+1} - \theta_{ij} \rvert \le
  H (t_{j+1} - t_j) \text{ for all } i, j\bigr\} , \\
  L_{H,G} = \sup_{\theta \in K_{H,G}} \mathbb{E}\bigl[d^w_{u_\theta}(F)\bigr] .
\end{gather*}
We write $\mathcal{U}_{H,G} = \{u_\theta : \theta \in K_{H,G}\}$.

\begin{lemma}[Grid approximation]
\label{lem:grid}
Every $u_\theta$ with $\theta \in K_{H,G}$ lies in $\mathcal{U}_H$, and $0 \le L_H - L_{H,G} \le
H\Delta / 2$; the supremum defining $L_{H,G}$ is attained. For fixed nodes, $H$, weights and logical
structure, $L_{H,G}$ is $(1+H)$-Lipschitz in $W_{1,w}$.
\end{lemma}

Interpolating an $H$-Lipschitz stake between the nodes moves it by at most $H\Delta/2$, and the
profit in every world moves by no more, since $\lvert F_i - z_i \rvert \le 1$ and the weights sum to
one. The bound is elementary. It is a worst case: with $H = 1$ and $G = 101$ uniform nodes it is
$0.005$, but on the example of Figure~\ref{fig:decomposition} ($H = 6$, $G = 6$) it is $0.6$, more
than $L_H = 0.240$, while the gap there is $0.011$. The last sentence of the lemma says that the
population value $L_{H,G}$ of the finite class used for computation is stable, as $L_H$ is; as in
Section~\ref{sec:stability}, this is not a statement about the finite-sample bounds below.

\begin{proposition}[Projected online gradient in the weighted metric]
\label{prop:ogd}
Let $\Pi^w_{K_{H,G}}$ be the projection onto $K_{H,G}$ in the norm of the inner product $\langle a,
b \rangle_w = \sum_i w_i \sum_j a_{ij} b_{ij}$. At time $t$, choose a worst world $\ell_t \in
\arg\min_\ell \sum_i w_i u_{\theta_t,i}(F_{t,i}) (F_{t,i} - z^\ell_i)$, set $h_{t,ij} =
\phi_j(F_{t,i}) (F_{t,i} - z^{\ell_t}_i)$, and update with a fixed step $\eta > 0$,
\[
  \theta_1 = 0, \qquad \theta_{t+1} = \Pi^w_{K_{H,G}}(\theta_t + \eta h_t) .
\]
For every sequence of bundles and every $\theta^* \in K_{H,G}$, with $f_t(\theta) =
d^w_{u_\theta}(F_t)$,
\[
  \sum_{t=1}^T \bigl\{ f_t(\theta^*) - f_t(\theta_t) \bigr\} \le \frac{G}{2\eta} +
  \frac{\eta T}{2} ,
\]
which is $\sqrt{GT}$ for $\eta = \sqrt{G/T}$.
\end{proposition}

Each profit $f_t$ is a minimum over worlds of functions linear in $\theta$, hence concave, and the
worst world at $\theta_t$ gives a supergradient; the update is projected gradient ascent
\citep{zinkevich2003online} with a fixed step. The metric matters. The profits carry the weights
$w_i$, so the supergradient in the inner product $\langle \cdot, \cdot \rangle_w$ is $h_t$ itself,
and the update direction must not be multiplied by $w_i$. Since $K_{H,G}$ constrains each component
separately, $\Pi^w_{K_{H,G}}$ projects each row of $\theta$ in the Euclidean norm, so the weights
enter the update only through the choice of the worst world $\ell_t$. Multiplying the direction by
$w_i$ still keeps the stakes in $K_{H,G}$, so it does not affect the validity of
Theorem~\ref{thm:fixed-t}(i) or Proposition~\ref{prop:anytime}, but it can lose the approach
guarantee. With $m = 200$ atoms of which exactly one holds ($\lvert \mathcal{W} \rvert = 200$), all
answered $1$ on every item, ties between worst worlds broken by the smallest index, $w_i = 1/200$,
$G = 2$, $H = 1$, $T = 10^4$ and $\alpha = 0.05$, $L_H = L_{H,G} = 0.995$ and
Theorem~\ref{thm:fixed-t}(ii) guarantees $\underline{L}_T \ge 0.4265$; the update above gives
$\underline{L}_T = 0.9643$, the weighted one $0.3228$. With a step $\eta = \sqrt{G/T_{\max}}$ fixed
for a budget $T_{\max}$, the regret bound is $\sqrt{GT}$ only at $T = T_{\max}$ and larger before it
(Figure~\ref{fig:power}(a)).

\begin{theorem}[Fixed-$T$ validity and approach]
\label{thm:fixed-t}
Fix the predictor, the bundle, the weights, the nodes, $H$, the horizon $T$ and the stake rule
before the first item is observed. Let $F_1, F_2, \dots$ be i.i.d.\ with law $P_F$, let the stakes
$u_t \in \mathcal{U}_{H,G}$ be predictable, that is, depend only on $F_1, \dots, F_{t-1}$ and on
data independent of $F_1, F_2, \dots$, write $d_t = d^w_{u_t}(F_t)$, $a_T = \sqrt{2
\log(2/\alpha)/T}$ and $\underline{L}_T = \max\{0, T^{-1} \sum_{t \le T} d_t - a_T\}$, and let $P
\in \mathcal{P}(P_F)$ be the true law.
\begin{enumerate}
  \item[(i)] \emph{Validity.} With probability at least $1 - \alpha/2$,
  \[
    \underline{L}_T \le L_{H,G} \le L_H \le \Gamma_H(P) \le \Theta_w(P) .
  \]
  \item[(ii)] \emph{Approach.} If the stakes are those of Proposition~\ref{prop:ogd} with $\eta =
  \sqrt{G/T}$, then with probability at least $1 - \alpha$ the chain of (i) holds together with
  \[
    \max\bigl\{0, \; L_H - H\Delta/2 - \sqrt{G/T} - 2 a_T\bigr\} \le \underline{L}_T .
  \]
\end{enumerate}
\end{theorem}

Part~(i) needs only i.i.d.\ items and stakes fixed before their items are seen: the expected profit
of a stake from $\mathcal{U}_{H,G}$ on a fresh item is at most $L_{H,G}$, and the bounded
differences concentrate. It therefore covers every way of learning the stakes in
Section~\ref{sec:exp-real}: the sequential $\mathcal{U}_{H,G}$ and $\mathcal{U}_0$ stakes, and the
split method, which learns its stake on the first half of the budget and scores only the second
half, so that the stake depends only on data independent of the scored items. Choosing an item's
stake after seeing that item breaks it: the stake can then depend on the whole bundle, which turns
the test into one of joint calibration, and the corresponding negative control in
Table~\ref{tab:coverage} fails. The concentration step is the Azuma--Hoeffding argument for bounded
martingale differences. Part~(ii) states the price of computing: with probability at least $1 -
\alpha$, the bound falls short of $L_H$ by at most the grid, learning and statistical terms
$H\Delta/2$, $\sqrt{G/T}$ and $2a_T$. These are worst cases. At $T = 10^4$ and $\alpha = 0.05$,
$2a_T \approx 0.054$, and $\sqrt{G/T}$ is about $0.10$ with $G = 101$ nodes and $0.033$ with $G =
11$. In Figure~\ref{fig:decomposition} ($H = 6$, $G = 6$, $T = 10^4$) the worst-case shortfall
$H\Delta/2 + \sqrt{G/T} + 2a_T$ is about $0.68$, while the bound fell short of $L_H$ by $0.048$ in
the median over 10 seeds. The figure splits the shortfall differently from the proof: its learning
gap $L_{H,G} - \rho_T$, where $\rho_T$ is the average population value of the stakes used, compares
population values, and Proposition~\ref{prop:ogd} bounds its expectation, not its value in each run
(Appendix~\ref{app:proofs-computation}). Both parts concern the bound at the horizon $T$ fixed in
advance. Computing $\underline{L}_T$ at many $T$ and keeping the largest is not covered by
Theorem~\ref{thm:fixed-t}; a bound to be read at many times is the anytime bound of
Proposition~\ref{prop:anytime}.

\begin{proposition}[Anytime-valid test and its inversion]
\label{prop:anytime} Let $F_1, F_2, \dots$ be i.i.d.\ with law $P_F$ and $\tau \in [0,1]$. Let the
stakes $u_t \in \mathcal{U}_{H,G}$ and the betting fractions $0 \le \lambda_t \le \tfrac12$ be
predictable, as in Theorem~\ref{thm:fixed-t}, and common to all $\tau$, and let $d_t =
d^w_{u_t}(F_t) \in [-1, 1]$. Under the null hypothesis $L_{H,G} \le \tau$, the wealth $M_t(\tau) =
\prod_{s \le t} \{1 + \lambda_s (d_s - \tau)\}$ is a nonnegative supermartingale, so $\Pr\{\exists t
: M_t(\tau) \ge 1/\alpha\} \le \alpha$. Consequently
\[
  \underline{L}^{\mathrm{any}}_t = \sup\Bigl(\bigl\{\tau \in [0,1] : \max_{s \le t} M_s(\tau) \ge
  1/\alpha\bigr\} \cup \{0\}\Bigr)
\]
satisfies $\Pr\{\exists t : \underline{L}^{\mathrm{any}}_t > L_{H,G}\} \le \alpha$. The same holds
for a fixed convex combination $\sum_k c_k M^{(k)}_t(\tau)$ of such wealth processes, each with its
own betting fractions and the same stakes, in place of $M_t(\tau)$.
\end{proposition}

The wealth $M_t(\tau)$ is the capital of a bettor who stakes the fraction $\lambda_t$ of it on the
profit exceeding $\tau$. Under the null hypothesis each bet is at best fair, and Ville's inequality
\citep{ville1939etude} bounds the chance that the capital ever reaches $1/\alpha$. The bound can
therefore be read at any time, including a time chosen by looking at the data, and monitoring longer
does not raise the error rate. Since $L_{H,G} \le \Gamma_H(P) \le \Theta_w(P)$, the probability that
the bound ever exceeds $\Theta_w(P)$ is at most $\alpha$ as well. Inverting on a grid of $\tau$, as
we do with step $10^{-3}$, reports the largest grid point at which the capital has reached
$1/\alpha$, which is at most $\underline{L}^{\mathrm{any}}_t$ and, as the rejected values of $\tau$
form an interval from $0$, less than one grid step below it. The capital process and its inversion
follow \citet{waudbysmith2024estimating}; what is new here is the target, the label-free $L_{H,G}$.
Proposition~\ref{prop:anytime} claims no rate of approach. Its null hypothesis concerns $L_{H,G}$
for i.i.d.\ items. If the law of the items changes, the construction still controls the error under
the null hypothesis that, given everything the stakes and betting fractions may depend on before
item $t$, the expected profit of item $t$ is at most $\tau$ for every $t$; it is not a confidence
sequence for a changing $\Theta_w$; and a change of the labels alone, with $P_F$ fixed, is invisible
to it, as to every label-free audit (Figure~\ref{fig:blind-spot}).

\section{Experiments}
\label{sec:experiments}

We first check the mechanism of the theory on synthetic examples whose answers are known
(Section~\ref{sec:exp-synthetic}), then ask whether valid certificates come out at a practical
budget on real predictors (Section~\ref{sec:exp-real}).

\subsection{Mechanism on Synthetic Data}
\label{sec:exp-synthetic}

\begin{figure}[t]
  \centering
  \includegraphics[width=\linewidth]{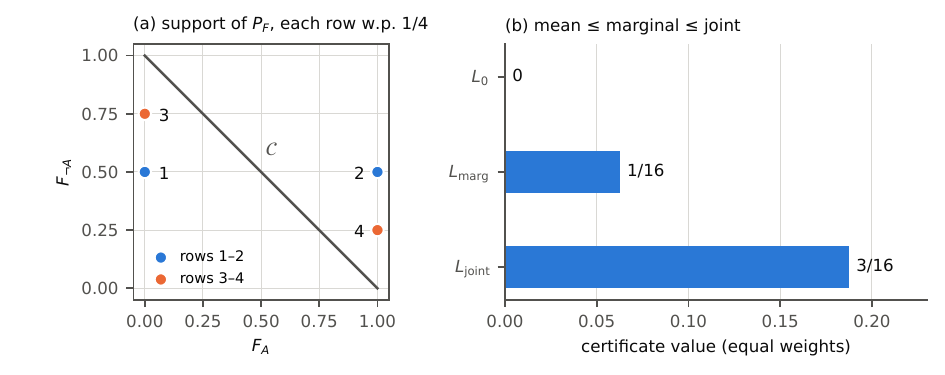}
  \caption{Four-point separation (Theorem~\ref{thm:separation}). (a) The bundle $(F_A, F_{\neg A})$
    takes four values with probability $1/4$ each; the coherent set $\mathcal{C}$ is the segment
    $F_A + F_{\neg A} = 1$. Rows~1--2 alone are Counterexample~\ref{cx:pointwise}: marginally
    calibrated under a compatible label law, yet pointwise incoherent. (b) With equal weights, the
    sharp label-free lower bounds on the mean, marginal and joint calibration residuals are $0 <
    1/16 < 3/16$.}
  \label{fig:four-point}
\end{figure}

\begin{figure}[t]
  \centering
  \includegraphics[width=\linewidth]{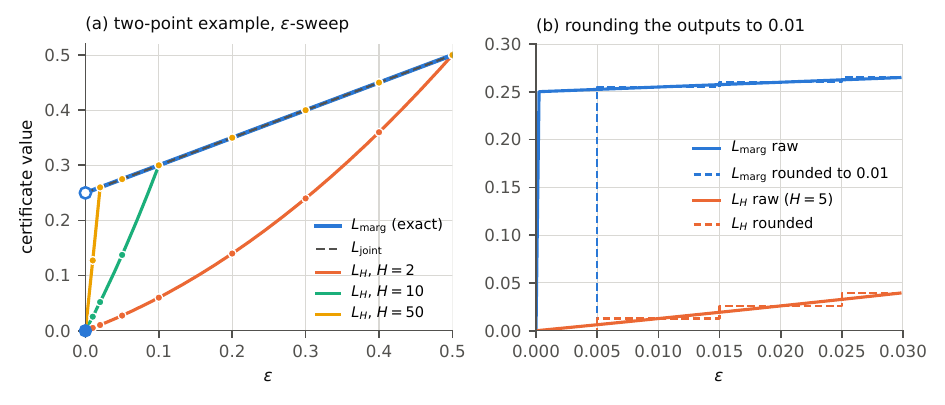}
  \caption{Sharpness versus stability on the two-point example $F^\varepsilon = (0, \tfrac12 -
    \varepsilon), (1, \tfrac12 + \varepsilon)$ (Proposition~\ref{prop:discontinuity},
    Corollary~\ref{cor:closed-form}). (a) The exact marginal bound jumps from $0$ to above
    $\tfrac14$ as soon as $\varepsilon > 0$, and equals the joint bound for $\varepsilon > 0$; the
    stabilised bound $L_H = (\tfrac14 + \tfrac\varepsilon2)\min\{1, H\varepsilon\}$ is continuous
    (lines: closed form; dots: Lipschitz-constrained LP). (b) Rounding the outputs to $0.01$ sends
    $L_{\mathrm{marg}}$ back to $0$ for $\varepsilon < 0.005$, while $L_H$ moves by at most
    $(1+H)W_{1,w}$ (Theorem~\ref{thm:stable}).}
  \label{fig:two-point}
\end{figure}

\begin{figure}[t]
  \centering
  \includegraphics[width=\linewidth]{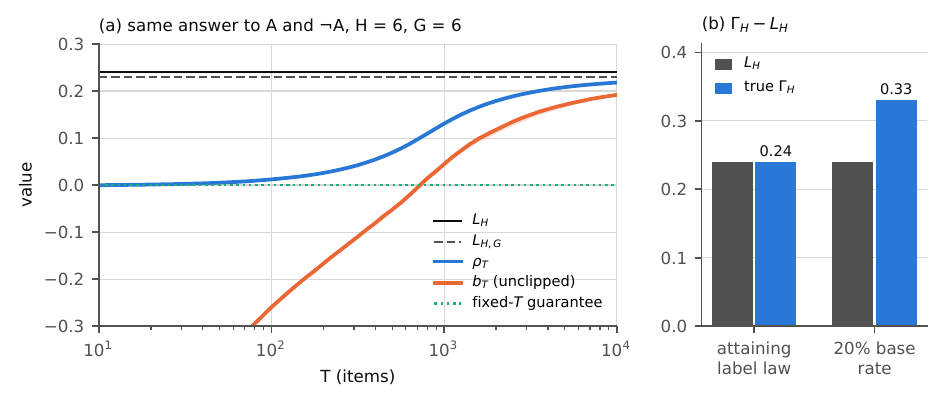}
  \caption{Where the fixed-$T$ certificate loses ground (Lemma~\ref{lem:grid},
    Proposition~\ref{prop:ogd}, Theorem~\ref{thm:fixed-t}). (a) A predictor answers ``$A$?'' and
    ``not $A$?'' with the same probability $x \in \{0.05, \dots, 0.95\}$, uniformly; its true label
    law is the one attaining $L_H$. With $H = 6$ and a uniform grid of $G = 6$ nodes, $L_H - b_T$
    splits into the grid gap $L_H - L_{H,G}$, the learning gap $L_{H,G} - \rho_T$ and the
    statistical gap $\rho_T - b_T$, where $\rho_T$ is the average population value of the stakes
    used and $b_T = T^{-1} \sum_{t \le T} d_t - a_T$; curves are medians over 10 seeds with 10--90\%
    bands, and the reported bound is $\max\{0, b_T\}$. The worst-case guarantee of
    Theorem~\ref{thm:fixed-t} is vacuous here ($H\Delta/2 > L_H$) although the bound approaches
    $L_{H,G}$. (b) The partial-identification gap $\Gamma_H - L_H$, a different decomposition: zero
    under the attaining label law, positive under a true label law with a 20\% base rate of $A$.}
  \label{fig:decomposition}
\end{figure}

On the four-point example the sharp bounds on the mean, marginal and joint calibration residuals are
$0 < 1/16 < 3/16$ (Figure~\ref{fig:four-point}); Table~\ref{tab:four-point} gives a label law and
stakes that attain the marginal value, and the sequential method approaches it
(Table~\ref{tab:decomposition}). On the two-point example the exact marginal bound jumps from $0$ to
above $\tfrac14$ under an arbitrarily small perturbation, and rounding the answers sends it back to
$0$; the stabilised bound $L_H$ is continuous, and under every perturbation we applied it moved by
at most $(1+H)W_{1,w}$, as Theorem~\ref{thm:stable} guarantees, while the exact marginal bound has
no such guarantee (Figure~\ref{fig:two-point}, Table~\ref{tab:perturbations}). On the example of
Figure~\ref{fig:decomposition}, in which a predictor gives ``$A$?'' and ``not $A$?'' the same answer
($H = 6$, $G = 6$), the figure splits the distance from $L_H$ to the fixed-$T$ bound into its grid,
learning and statistical parts. At $T = 10^4$ they are $0.011$, $0.011$ and $0.026$ against $L_H =
0.240$, although Theorem~\ref{thm:fixed-t}'s worst-case guarantee is vacuous at these $H$ and $G$.
These are checks of the theory, not empirical claims.

\begin{table}[t]
  \centering
  \begin{tabular}{rrrrrrr}
\toprule
Row & $F_A$ & $F_{\neg A}$ & $P(A\mid F)$ & $u_A(F_A)$ & $u_{\neg A}(F_{\neg A})$ & $d_u$ \\
\midrule
1 & $0$ & $\tfrac{1}{2}$ & $0$ & $-1$ & $0$ & $0$ \\
2 & $1$ & $\tfrac{1}{2}$ & $1$ & $1$ & $0$ & $0$ \\
3 & $0$ & $\tfrac{3}{4}$ & $0$ & $-1$ & $-1$ & $\tfrac{1}{8}$ \\
4 & $1$ & $\tfrac{1}{4}$ & $1$ & $1$ & $1$ & $\tfrac{1}{8}$ \\
\bottomrule
\end{tabular}

  \caption{An attaining label law and optimal stakes for Figure~\ref{fig:four-point}: under this law
    $\Theta_w = 1/16$, and these stakes earn $\mathbb{E}[d_u] = 1/16$, so the sharpness of the
    marginal bound can be checked without an LP solver.}
  \label{tab:four-point}
\end{table}

\begin{table}[t]
  \centering
  \small
  \begin{tabular}{llrrrr}
\toprule
Condition & Perturbation & $W_{1,w}$ & $(1+H)W_{1,w}$ & $|\Delta L_{\mathrm{marg}}|$ & $|\Delta L_H|$ \\
\midrule
four-point & jitter 0.001 & 0.0008 & 0.0045 & 0.0000 & 0.0002 \\
four-point & jitter 0.01 & 0.0075 & 0.0450 & 0.0000 & 0.0024 \\
four-point & jitter 0.05 & 0.0375 & 0.2250 & 0.0000 & 0.0156 \\
four-point & quantize 0.01 & 0.0000 & 0.0000 & 0.0000 & 0.0000 \\
four-point & quantize 0.1 & 0.0125 & 0.0750 & 0.0000 & 0.0000 \\
four-point & clip 0.01 & 0.0050 & 0.0300 & 0.0050 & 0.0050 \\
four-point & clip 0.05 & 0.0250 & 0.1500 & 0.0250 & 0.0250 \\
Counterexample~\ref{cx:pointwise} & jitter 0.001 & 0.0008 & 0.0045 & 0.0008 & 0.0000 \\
Counterexample~\ref{cx:pointwise} & jitter 0.01 & 0.0075 & 0.0450 & 0.0075 & 0.0003 \\
Counterexample~\ref{cx:pointwise} & jitter 0.05 & 0.0375 & 0.2250 & 0.0375 & 0.0063 \\
Counterexample~\ref{cx:pointwise} & quantize 0.01 & 0.0000 & 0.0000 & 0.0000 & 0.0000 \\
Counterexample~\ref{cx:pointwise} & quantize 0.1 & 0.0000 & 0.0000 & 0.0000 & 0.0000 \\
Counterexample~\ref{cx:pointwise} & clip 0.01 & 0.0050 & 0.0300 & 0.0000 & 0.0000 \\
Counterexample~\ref{cx:pointwise} & clip 0.05 & 0.0250 & 0.1500 & 0.0000 & 0.0000 \\
two-point, $\varepsilon=0.1$ & jitter 0.001 & 0.0008 & 0.0045 & 0.0003 & 0.0001 \\
two-point, $\varepsilon=0.1$ & jitter 0.01 & 0.0075 & 0.0450 & 0.0025 & 0.0010 \\
two-point, $\varepsilon=0.1$ & jitter 0.05 & 0.0375 & 0.2250 & 0.0125 & 0.0000 \\
two-point, $\varepsilon=0.1$ & quantize 0.01 & 0.0000 & 0.0000 & 0.0000 & 0.0000 \\
two-point, $\varepsilon=0.1$ & quantize 0.1 & 0.0000 & 0.0000 & 0.0000 & 0.0000 \\
two-point, $\varepsilon=0.1$ & clip 0.01 & 0.0050 & 0.0300 & 0.0050 & 0.0025 \\
two-point, $\varepsilon=0.1$ & clip 0.05 & 0.0250 & 0.1500 & 0.0250 & 0.0125 \\
\bottomrule
\end{tabular}

  \normalsize
  \caption{Perturbations of the synthetic conditions against the bound of Theorem~\ref{thm:stable}
    ($H = 5$): $|\Delta L_H| \le (1+H)W_{1,w}$ always holds, while $|\Delta L_{\mathrm{marg}}|$ has
    no such guarantee.}
  \label{tab:perturbations}
\end{table}

\begin{table}[t]
  \centering
  \small
  \begin{tabular}{lrrrrrrrrr}
\toprule
Condition & $H$ & $G$ & $T$ & $L_H$ & $L_{H,G}$ & $\rho_T$ & $b_T$ & learning & statistics \\
\midrule
same answer & 6 & 6 & 100 & 0.2400 & 0.2290 & 0.0120 & -0.2597 & 0.2170 & 0.2717 \\
same answer & 6 & 6 & 1{,}000 & 0.2400 & 0.2290 & 0.1306 & 0.0447 & 0.0984 & 0.0859 \\
same answer & 6 & 6 & 10{,}000 & 0.2400 & 0.2290 & 0.2180 & 0.1917 & 0.0110 & 0.0263 \\
four-point (Fig.~1) & 8 & 5 & 100 & 0.0625 & 0.0625 & 0.0020 & -0.2697 & 0.0605 & 0.2717 \\
four-point (Fig.~1) & 8 & 5 & 1{,}000 & 0.0625 & 0.0625 & 0.0220 & -0.0644 & 0.0405 & 0.0864 \\
four-point (Fig.~1) & 8 & 5 & 10{,}000 & 0.0625 & 0.0625 & 0.0578 & 0.0304 & 0.0047 & 0.0275 \\
\bottomrule
\end{tabular}

  \normalsize
  \caption{The decomposition of Figure~\ref{fig:decomposition} at three horizons (medians over 10
    seeds). On the four-point example of Figure~\ref{fig:four-point} ($H = 8$, $G = 5$), $L_{H,G} =
    L_H = L_{\mathrm{marg}} = 1/16$ and the sequential method approaches it.}
  \label{tab:decomposition}
\end{table}

\subsection{Real Data: Valid Certificates on a Budget}
\label{sec:exp-real}

\emph{Setup.} Four predictors answer the queries of a fixed bundle on two public classification data
sets: TypeSafe Jev through its API (version \texttt{jev-1.13.0}); Eikos-4B (8-bit) and its base
model, Qwen3.5-4B, both run locally; and a zero-shot NLI classifier (DeBERTa-v3-large). There is one
predictor per family, so we do not attribute differences between systems to how they were built. On
AG News ($K = 4$) the bundle asks a four-way choice query and, for every label, whether the item has
that label (an atom) and whether it does not (a complement) ($m = 12$); on DBpedia-14 ($K = 14$) it
adds five explicit unions of labels ($m = 47$). Each role carries the same total weight, split
equally among its components, and the predictors are used zero-shot. A 200-item pilot on separate
exploration items selected $H$ and $G$ ($H = 2$ on AG News, $H = 1$ on DBpedia-14, $G = 6$ uniform
nodes on both) and each system's budget $T_{\max}$ ($10{,}000$ items, or $5{,}000$ and $2{,}000$ for
the two slowest local runs on DBpedia-14). Before any confirmation item was sent, we froze and
preregistered these choices, the betting rule (the mixture of $\lambda_k = 2^{-k}$, $k = 1, \dots,
8$) and the confirmatory family: one anytime lower bound (Proposition~\ref{prop:anytime}) per system
and data set, with $\mathcal{U}_{H,G}$ stakes learnt sequentially, each at $\alpha = 0.05/7$. Jev on
DBpedia-14 did not meet the preregistered pilot criterion for collecting confirmation data and has
no confirmatory certificate; its pilot bound was $0$. The pilot also included preregistered
controls, which vary the option names, the option order and whether Jev receives the whole bundle in
one request; they entered neither the selection nor the confirmatory family, and the statistics by
which we compare them were not preregistered (Appendix~\ref{app:controls}). Appendix~\ref{app:study}
lists the data, the bundles, how each predictor's probabilities are read and the records of the
pilot and of the confirmatory collection, and Appendix~\ref{app:preregistration} the two
registrations, the analyses that were not preregistered and how the figures and tables are
reproduced.

\begin{figure}[t]
  \centering
  \includegraphics[width=\linewidth]{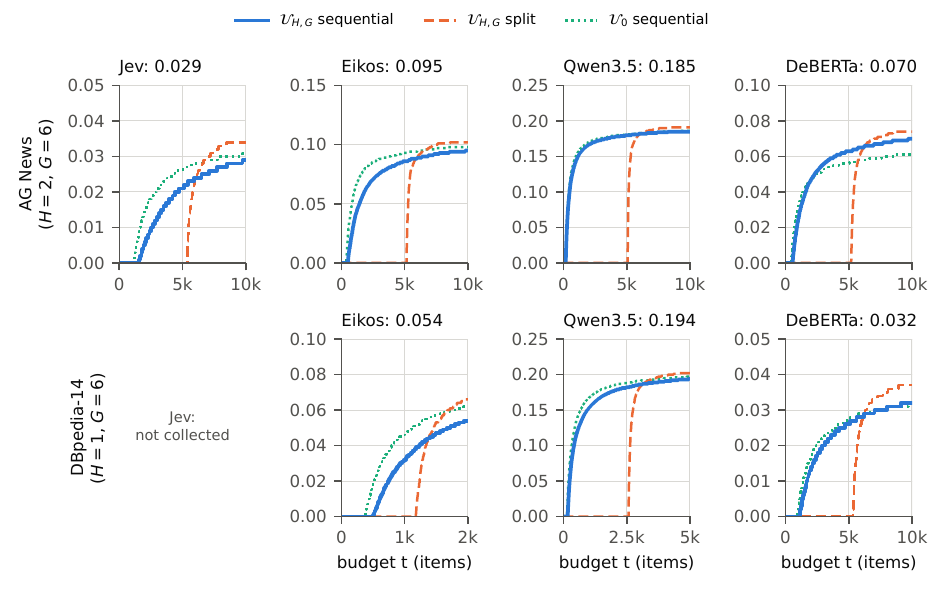}
  \caption{Anytime-valid lower bounds (Proposition~\ref{prop:anytime}) on the preregistered
    confirmatory streams, at $\alpha = 0.05/7$ per certificate. Each panel is one system on one data
    set, over the budget $t$ up to its $T_{\max}$ (so the $x$ scales differ), with its own $y$
    scale. The three ways of learning the stakes share the stream (the same items in the same
    arrival order), the bundle, the betting rule and the total budget. The $\mathcal{U}_{H,G}$
    bounds are on $L_{H,G}$; the $\mathcal{U}_0$ stakes are constants, a mean-only audit (they
    target $L_0$). The split method learns on the first $\lfloor T_{\max}/2 \rfloor$ items without
    scoring them, so its bound is $0$ there. The $\mathcal{U}_{H,G}$-sequential bound is the
    confirmatory certificate; its value at $T_{\max}$ is in each panel's title. Every bound is on
    the bundle total, not on any one question. Jev on DBpedia-14 did not meet the preregistered
    pilot criterion for collecting confirmation data (``not collected''); its pilot anytime bound,
    on the 200 exploration items, was $0$.}
  \label{fig:real-data}
\end{figure}

\emph{Confirmatory result.} All seven confirmatory certificates were positive, from $0.029$ (Jev on
AG News) to $0.194$ (Qwen3.5 on DBpedia-14; Figure~\ref{fig:real-data}, Table~\ref{tab:real-data}).
Each first turned positive after $170$ to $1{,}580$ items. Under the assumptions of
Proposition~\ref{prop:anytime}, each certificate refutes that its system is marginally calibrated on
its bundle ($\Theta_w = 0$), and the seven refutations hold together at family-wise level $0.05$:
each certificate bounds $L_{H,G}$ from below, and $L_{H,G} \le \Theta_w$
(Theorem~\ref{thm:fixed-t}). A certificate bounds the bundle total from below and does not say which
question is miscalibrated (Proposition~\ref{prop:attribution}). For both $\mathcal{U}_{H,G}$ methods
the anytime bound at $T_{\max}$ was at least the fixed-$T$ bound, which was $0$ for two of the seven
sequential runs.

\emph{Three ways of learning the stakes.} Figure~\ref{fig:real-data} also compares three ways of
learning the stakes on the same streams; these are secondary results, outside the confirmatory
family. At $T_{\max}$ the mean-only audit ($\mathcal{U}_0$ learnt sequentially) was above the
confirmatory method on five pairs (by $0.001$ to $0.008$), equal on one and below on one (DeBERTa on
AG News, $0.061$ against $0.070$), and it turned positive first on all seven; $\mathcal{U}_{H,G}$
learnt by the split method ended highest on all seven. Constant stakes lie in $K_{H,G}$, so $L_0 \le
L_{H,G}$ for every distribution, as $L_0$ is the value of the best constant stake
(Theorem~\ref{thm:stable} with $H = 0$): the mean-only audit targets no more, and it can come out
ahead only at a finite budget, through the learning and statistical terms, which the two classes pay
differently. A sequential bound pays the learning gap of Theorem~\ref{thm:fixed-t} from the first
item, and Proposition~\ref{prop:ogd} bounds that gap by a term that grows with the number of nodes
$G$; for constant stakes the same bound holds with $G = 1$ (Appendix~\ref{app:proofs-computation}).
An anytime bound also depends on the spread of the profits $d_t$, not only on their mean. The split
method carries no learning regret into the half it scores, at the price of half the budget and of
the estimation error of the stake it learnt, which Theorem~\ref{thm:fixed-t} does not bound. That
the split method ended higher is consistent with the learning gap of Figure~\ref{fig:decomposition};
that it turned positive later is by construction, since it scores only the second half. To separate
the stake class from the learning method we added, after seeing Figure~\ref{fig:real-data}, a split
arm for $\mathcal{U}_0$ (Appendix~\ref{app:u0-split}; post hoc, not preregistered). With the
learning method held fixed, the predictor-dependent stakes improved on the constant ones for the two
DeBERTa pairs (by $0.008$ and $0.002$) and by one grid step for Qwen3.5 on AG News, and tied on the
other four. On these streams, at the frozen $H$, $G$ and budgets, the predictor-dependent stakes
added little to what the constant stakes found. These explanations are consistent with the data; we
did not test them.

\emph{Labelled metrics and plug-in values.} With the items' labels we also estimated $\Theta_w$ by
binned ECE, and computed $L_0$, $L_{\mathrm{marg}}$ and $L_{\mathrm{joint}}$ on the empirical
distribution of each stream as uncorrected plug-ins (Appendix~\ref{app:descriptive},
Table~\ref{tab:descriptive}). None of them is a certificate, and their estimation conditions were
not preregistered. The labelled estimates exceeded the certificates on every pair. That is
consistent with $L_{H,G} \le \Theta_w$, but being estimates they do not check the certificates: the
inequality concerns the population $\Theta_w$. The certificates were between about one half and nine
tenths of the estimates. The shortfall mixes four things: the part of $\Theta_w$ no label-free bound
can reach ($\Theta_w - L_{\mathrm{marg}}$), the price of stability and of the grid
($L_{\mathrm{marg}} - L_{H,G}$), the learning and statistical slack of the sequential bound
($L_{H,G}$ minus the certificate), and the bias of the binned estimate itself, which can go either
way (binning averages miscalibration out within a bin; sampling noise adds to it). The plug-ins show
the instability of the exact marginal bound (Proposition~\ref{prop:discontinuity}) on real outputs.
In every coordinate, DeBERTa's outputs took at least $0.997\,T$ distinct values over the $T$ items
of a stream, and on its streams the empirical marginal bound coincided with the joint one to within
$10^{-6}$, as it must when each coordinate's value tells the items apart; on DBpedia-14 it even
exceeded every labelled estimate of $\Theta_w$, although $L_{\mathrm{marg}} \le \Theta_w$ in the
population. Each of the other three systems (Jev, Eikos and Qwen3.5) had a coordinate with at most
$0.02\,T$ distinct values; on their streams the empirical marginal bound stayed below the joint one
and within $0.007$ of the empirical mean bound. The stabilised certificate is a valid lower bound in
both regimes.

\begin{table}[t]
  \centering
  \small
  \begin{tabular}{lrrrrrr}
\toprule
 & & \multicolumn{3}{c}{anytime} & \multicolumn{2}{c}{fixed-$T$} \\
\cmidrule(lr){3-5}\cmidrule(lr){6-7}
System & $T_{\max}$ & $\mathcal{U}_{H,G}$ seq. & $\mathcal{U}_{H,G}$ split & $\mathcal{U}_0$ seq. & $\mathcal{U}_{H,G}$ seq. & $\mathcal{U}_{H,G}$ split \\
\midrule
\multicolumn{7}{l}{AG News ($H = 2$, $G = 6$)} \\
Jev & 10{,}000 & \textbf{0.029} & 0.034 & 0.031 & 0 & 0 \\
Eikos & 10{,}000 & \textbf{0.095} & 0.102 & 0.098 & 0.0641 & 0.0587 \\
Qwen3.5 & 10{,}000 & \textbf{0.185} & 0.191 & 0.186 & 0.1543 & 0.1473 \\
DeBERTa & 10{,}000 & \textbf{0.070} & 0.074 & 0.061 & 0.0386 & 0.0308 \\
\midrule
\multicolumn{7}{l}{DBpedia-14 ($H = 1$, $G = 6$)} \\
Eikos & 2{,}000 & \textbf{0.054} & 0.066 & 0.062 & 0 & 0 \\
Qwen3.5 & 5{,}000 & \textbf{0.194} & 0.202 & 0.197 & 0.1499 & 0.1415 \\
DeBERTa & 10{,}000 & \textbf{0.032} & 0.037 & 0.032 & 0.0005 & 0 \\
\bottomrule
\end{tabular}

  \normalsize
  \caption{The certificates of Figure~\ref{fig:real-data} at $T_{\max}$, all at the same $\alpha$:
    the confirmatory anytime bound (bold) and the secondary results. Every anytime bound, the
    confirmatory one included, is on a grid of $10^{-3}$; the fixed-$T$ bounds
    (Theorem~\ref{thm:fixed-t}) are rounded to four decimals, and $0$ means the bound was clipped at
    zero. Under the assumptions of Proposition~\ref{prop:anytime} (the items of a stream i.i.d.\ in
    arrival order, predictable stakes), each positive confirmatory bound refutes that its system is
    marginally calibrated ($\Theta_w = 0$) on that bundle, the seven together at family-wise level
    $0.05$; a zero is never read as evidence of calibration.}
  \label{tab:real-data}
\end{table}

\section{Related Work}
\label{sec:related}

Table~\ref{tab:related} compares the closest works with ours from four viewpoints: the quantity
defined, the information observed, the object optimised over and the kind of guarantee.

\begin{table}[p]
  \centering
  \scriptsize
  \setlength{\tabcolsep}{3pt}
  \begin{tabular}{@{}>{\raggedright\arraybackslash}p{2.3cm}>{\raggedright\arraybackslash}p{3.1cm}>{\raggedright\arraybackslash}p{2.3cm}>{\raggedright\arraybackslash}p{2.8cm}>{\raggedright\arraybackslash}p{3.85cm}@{}}
    \toprule
    Work & Quantity & Observed & Optimised over & Guarantee \\
    \midrule
    This paper & least information-restricted calibration residual over the compatible label
      laws (mean, marginal, joint, smooth) & law of the bundle and logic; no labels & stakes
      restricted in information and Lipschitz constant, settled in the worst world & sharpness
      (finite support for the general minimax); strict separation on an example;
      no attribution to a given question; discontinuity and no uniform constant factor;
      $(1+H)$-stability; zero set kept for $H > 0$ (population); for i.i.d.\ items and
      predictable stakes, fixed-$T$ validity, approach for the stakes of
      Proposition~\ref{prop:ogd}, and anytime validity without a rate \\
    \citet{debona2015measuring}; \citet{schervish2003measures} & incoherence: distance to
      coherent assessments, sure loss & assessments and logic; no labels & bounded stakes, or
      coherent assessments & identities by LP duality; postulates; no link to calibration \\
    \citet{andrews2026dutch} & worst-case sure profit per item & predictions and logic; no
      labels & stakes bounded in $\ell_1$ on the whole bundle & identity with a supremum-norm
      distance; bootstrap intervals; no link to calibration \\
    \citet{paleka2025consistency} & arbitrage in a logarithmic scoring market & forecasts and
      relations; no labels & arbitrageur's prices & values per checked tuple; correlation with
      Brier scores, empirical \\
    \citet{arieli2021feasible} & feasibility of joint posteriors under a common prior & joint law
      of posteriors on one binary state; no state & trades in $[-1,1]$ on one's own posterior &
      feasible under a prior in $(0,1)$ if and only if no
      trade profits, apart from the laws $F \equiv 0^m$ and $F \equiv 1^m$ \\
    \citet{kong2026calibration} & least calibration error of two forecasters over the compatible
      label laws & joint law of two forecasts of one label; no labels & zero-mean bounded trades
      on one's own forecast & bracket within a factor of two; zero set; population only \\
    \citet{blasiok2023unifying}; \citet{gopalan2026importance} & distances from calibration;
      smooth calibration error & predictions and labels & couplings with calibrated laws;
      Lipschitz weights & equivalence within a factor of two; continuity, completeness,
      soundness; ECE discontinuous \\
    \citet{resin2026hierarchies}; \citet{dwork2021outcome} & hierarchies of calibration notions
      & predictions and labels & -- (classes of distinguishers for \citealt{dwork2021outcome}) &
      implications and counterexamples; distinguisher access matched to calibration notions \\
    \citet{arnold2023sequentially}; \citet{antonov2026bet}; \citet{henzi2024safe} & exact
      calibration as the null & outcomes observed & bets learnt online & anytime validity, or
      validity at a fixed sample size \citep{henzi2024safe}; power or growth up to a regret term
      \citep{antonov2026bet,henzi2024safe} \\
    \citet{waudbysmith2024estimating} & mean of a bounded variable & the variable observed &
      predictable bet sizes & confidence sequences and fixed-$T$ intervals \\
    \bottomrule
  \end{tabular}
  \normalsize
  \caption{The closest works from four viewpoints. In the table, ``labels'' means that realised
    outcomes are observed; ``logic'' means that relations among different events are used.}
  \label{tab:related}
\end{table}

\emph{Incoherence and Dutch books.} Incoherence of probability assessments has long been measured by
their distance to the coherent assessments, or by the sure loss of a bettor with bounded stakes
\citep[see][]{schervish2003measures,staffel2015measuring}, and computed by linear programs
\citep{potyka2014linear}. \citet{debona2015measuring} prove that the weighted $\ell_1$ distance to
the coherent set equals the largest sure loss under box-bounded stakes, the identity of
Corollary~\ref{cor:special}(b). \citet{andrews2026dutch} apply Dutch books to language models, with
stakes bounded in $\ell_1$, the dual distance in the supremum norm and bootstrap confidence
intervals, and \citet{dragonevskiy2026understanding} proposes bounded Dutch books as a training
objective, while \citet{andrews2026revealed} grounds label-free evaluation in representation
theorems and notes that coherence does not restrict which objective rationalises the behaviour, a
limitation that applies to our bounds as well. \citet{paleka2025consistency} measure the arbitrage
of a logarithmic scoring market on tuples of related forecasts, after the consistency checks of
\citet{fluri2024evaluating}, and find it correlated with Brier scores known only later.
\citet{zhu2024incoherent} and \citet{matta2026rethinking} measure violations of the laws of
probability by language models, and \citet{li2026beyond} and \citet{joy2026system} by typed decision
models; \citet[Section~5.4]{li2026beyond} observe that a systematic violation over a partition is
also an error of calibration in the large, which Corollary~\ref{cor:stages} makes exact and sharp
for every bundle ($L_0$). None of these works bounds calibration error by incoherence in general.
The weighted $\ell_1$ form of the joint-stage measure, the distance of a bundle to $\mathcal{C}$, is
that of \citet{debona2015measuring}. What we add is the reading of its expectation, for finitely
supported laws, as the sharp label-free lower bound on the joint calibration error, which does not
bound the marginal one (Counterexample~\ref{cx:pointwise}), and the marginal and mean stages below
it.

\emph{Coherence, feasibility and calibration without labels.} Coherence and calibration have long
been studied together in the foundations of probability
\citep{dawid1982well,seidenfeld1985calibration}. For one event put to several agents,
\citet{arieli2021feasible} characterise by a no-trade condition the laws of posterior beliefs that a
common prior in $(0,1)$ can produce, apart from two degenerate laws
(Corollary~\ref{cor:special}(a)), and for two agents they attribute an equivalent characterisation
to \citet{dawid1995coherent}; for finitely supported laws, Corollary~\ref{cor:special}(a) gives the
condition a value. \citet{kong2026calibration} take the least calibration error of two forecasters
of one label over the label laws compatible with their joint law, and bracket it by the arbitrage of
bounded trades; Corollary~\ref{cor:special}(c) gives the exact dual. Label-free partial
identification of calibration error is therefore not new; what is new here is a single minimax for
bundles of different, logically related events under every information restriction, and what follows
from it on such bundles: the strict separation, the impossibility of attribution, the stabilisation
and the certificates. \citet{kong2026calibration} also improve a forecaster without labels under the
assumption that a reference forecaster is calibrated, which parallels
Proposition~\ref{prop:attribution}: without labels, a statement about one forecaster needs an
assumption on the others. In the opposite direction, \citet{chen2026partial} assume that the reports
of a language model are calibrated and partially identify the prevalence of a latent label.
\citet{wang2026verification} shows that without labels the calibration error of a single predictor
cannot be estimated well in the worst case, in line with Lemma~\ref{lem:no-constraint}, by which a
single question about an event that is neither sure nor impossible is never refuted.
\citet{popordanoska2023estimating} estimate calibration error under label shift without target
labels, from source labels and importance weights rather than logical structure. Our bounds are
sharp in the sense of partial identification \citep{manski2003partial}. Lipschitz restrictions
narrow identified sets when they are placed on unobserved response functions
\citep{kim2018identification}; ours are placed on the bettor's stakes, and what they buy is
continuity in the observed law. A rich enough class of instrument functions leaves an identified set
unchanged \citep[Lemma~2]{andrews2013inference}, as Lipschitz stakes with $H > 0$ leave the zero set
unchanged (Corollary~\ref{cor:zero-set}), and in martingale optimal transport the value of a
coupling problem with conditional-mean constraints can be unstable in the observed marginal laws
\citep{bruckerhoff2022instability}, a parallel to Proposition~\ref{prop:discontinuity}.

\emph{Calibration measures with labels.} With labels observed, \citet{blasiok2023unifying} develop a
theory of the distance from calibration, show that the expected calibration error is discontinuous
and relate the smooth calibration error to the lower distance from calibration within a factor of
two; \citet{gopalan2026importance} relate it to an earth mover's distance to the calibrated laws,
and \citet{blasiok2024smooth} show that a kernel-smoothed expected calibration error with bandwidth
$\sigma$ is $(1 + \sigma^{-1})$-Lipschitz in the Wasserstein distance, a labelled counterpart of the
stability in Theorem~\ref{thm:stable}. The weight family of smooth calibration goes back to
\citet{kakade2008deterministic} \citep[see also][]{foster2018smooth,gopalan2025calibration};
\citet{hartline2025smooth} relate smooth calibration to decision making, and \citet{hu2024testing}
compute the smooth calibration error by a linear program with Lipschitz constraints between
neighbouring values, as we compute $L_H$ (Lemma~\ref{lem:extension}). The access of a distinguisher
orders calibration notions in outcome indistinguishability \citep{dwork2021outcome}, and the
conditioning $\sigma$-algebra orders them in \citet{resin2026hierarchies}; our stages are the
label-free, refutation side of these hierarchies. All of these works measure calibration on observed
outcomes; our bounds are the least values over the label laws compatible with the answers.

\emph{Sequential inference.} The capital process and its inversion follow
\citet{waudbysmith2024estimating}. Learning a witness function or a bet online, by projected
gradient ascent or online convex optimisation, is the approach of \citet{shekhar2024nonparametric}
and \citet{casgrain2024anytime}. Null hypotheses with a tolerance are tested by betting without
labels in audits of statistical parity \citep{chugg2023auditing} and of the behaviour of language
models \citep{richter2024auditing}, but not for calibration. Sequential tests of calibration observe
the outcomes and test exact calibration: \citet{arnold2023sequentially} for probability forecasts
and \citet{antonov2026bet} for conditional quantile forecasts with nulls indexed by information
sets. Closest to Theorem~\ref{thm:fixed-t} is the safe Hosmer--Lemeshow test of
\citet{henzi2024safe}, a test at a fixed sample size that bets with a calibration curve learnt
online and bounds its growth up to a regret term. Sequential comparisons of forecasters
\citep{henzi2022valid,choe2024comparing} also observe the outcomes, and they compare forecasters
rather than test calibration. The certificates here use no label, and they bound the size of a
calibration error from below instead of testing exact calibration.

\emph{Audits of typed decision models.} Typed decision models have been evaluated against human
judgements as judges, with their confidence used to decide when to escalate \citep{li2026jevjudge};
against labels as annotators in computational social science, including the calibration of their
confidence \citep{ibrahim2026evaluating}; and against coded fields and human judgements as coders of
crash narratives, with recalibration on those labels \citep{rafe2026calibrated}. Their decisions
also flip when the names of options are reassigned to other definitions \citep{sun2026typesafe}.
These audits rest on labels or on controlled changes of the questions; the certificates here bound
calibration error from the answers to unchanged questions, without labels.

\section{Limitations}
\label{sec:limitations}

\emph{Blind spots.} A label-free certificate sees only the distortions that push answers out of the
coherent set. A predictor whose answers are coherent leaves no trace however miscalibrated it is
(Counterexample~\ref{cx:coherent}); a law with $L_{\mathrm{marg}} = 0$ stays unrefuted for every $H$
(Corollary~\ref{cor:zero-set}); and a change of the law of the labels given the answers, with $P_F$
fixed, is invisible to every label-free audit (Figure~\ref{fig:blind-spot}). A rise of a certificate
is not a rise of the calibration error either, since two lower bounds do not order the errors they
bound. Certificates therefore do not replace labelled audits. They show where a labelled audit is
needed.

\emph{Semantics.} A certificate refutes calibration relative to the logical structure entered in
$\mathcal{C}$. A relation entered wrongly makes a valid certificate refute the wrong thing: in
Table~\ref{tab:blind-spot}, a calibrated and coherent predictor audited under a structure that
forgets one label receives a positive certificate although $\Theta_w = 0$, while dropping a doubtful
relation only lowers the bound. On real data the structure rests on semantics that we fixed in
advance and did not verify. We use unconditional events only, and unions are asked as explicit
disjunctions of labels. A choice query answered by a typed decision model is relative, settling
which of the listed options holds \citep{typesafe2026jaggedness}, and we read the probabilities of
every choice query as the unconditional probabilities of the labels only because the options exhaust
the label set, which we assume for both data sets without checking it. Conditional events, ordinal
scores and choice queries over sub-menus change what calibration is about, and we leave them out.

\emph{Number of worlds.} The theory holds for any finite set of worlds, but the computations
enumerate them. In this paper every event is a function of one label, so there are as many worlds as
labels: finding the worst world of an item takes $O(m\lvert\mathcal{W}\rvert)$ operations, and the
linear program for $L_{\mathrm{marg}}$ on $N$ support points has $N\lvert\mathcal{W}\rvert$
variables for the label law. For general propositions in natural language the worlds are the
consistent truth assignments, whose number can grow exponentially in the number of propositions. A
certificate there needs an implicit description of the coherent set, or a superset of the worlds,
which keeps every certificate valid at the price of a weaker bound.

\emph{Assumptions and scope.} The certificates assume that the items of a stream are i.i.d.\ in
arrival order and that each stake is fixed before its item is seen. Under a changing law the anytime
construction controls the error only for the conditional null hypothesis stated after
Proposition~\ref{prop:anytime}, and it is not a confidence sequence for a changing $\Theta_w$. The
approach guarantee of Theorem~\ref{thm:fixed-t} is a worst case and can be vacuous, as it is in
Figure~\ref{fig:decomposition}, and the stability of Theorem~\ref{thm:stable} and
Lemma~\ref{lem:grid} concerns population values, not the finite-sample bounds. A certificate bounds
the bundle total, and attributing it to one question needs assumptions on the others
(Proposition~\ref{prop:attribution}), which we do not pursue. The real-data study has one predictor
per family, two data sets and zero-shot use, so it compares configurations without identifying why
they differ; its comparisons of the ways of learning the stakes are secondary, and one of them is
post hoc.

\emph{Repair.} A positive certificate does not say how to correct the predictor.
Appendix~\ref{app:repair} treats repair without labels for strictly proper scoring rules, oriented
as penalties, whose generator is strictly convex and continuously differentiable on a neighbourhood
of the cube, such as the Brier score but not the logarithmic score. Projecting the whole bundle onto
$\mathcal{C}$ in the Bregman divergence of such a score, to the closest coherent bundle
$\Pi_\Phi(F)$, never raises the score in any world and lowers it wherever the bundle is incoherent
\citep{predd2009probabilistic}. The calibration regret of the score, its calibration part, then
falls by at least the expected divergence between $F$ and $\Pi_\Phi(F)$, which for the Brier score,
$\mathbb{E}\lVert F - \Pi_\Phi(F) \rVert_w^2$, is computable without labels
(Theorem~\ref{thm:repair}). Both hold only as far as the logical structure entered in $\mathcal{C}$
is right. The guarantee is for the bundle and that score, not for the residuals that certificates
bound: a projected predictor is coherent, so every population certificate of it is zero, while its
marginal calibration error can be higher than before (Counterexample~\ref{cx:repair}). A question
that is neither sure nor impossible cannot be repaired in the same sense. No post-processing that
changes its answers with positive probability keeps its expected Brier score from rising under every
compatible law, while its calibration error can be lowered without labels by replacing its answers
with their mean, which gives up resolution (Proposition~\ref{prop:focal-repair}). The inequalities
are known, and Appendix~\ref{app:repair} gives their sources; what this paper adds is what they
imply for certificates.

\section{Conclusion}
\label{sec:conclusion}

Without labels one can show that a predictor is miscalibrated, but never that it is calibrated. For
finitely supported laws, the law of a predictor's answers to logically related questions refutes
exactly the calibration error that a bettor can extract by using the permitted information and
settling every input in its worst world. For marginal calibration this sharp bound is unstable: the
logical constraints that make refutation possible can also make it jump. Restricting the stakes to
Lipschitz functions buys stability at a price, which we computed in closed form on a two-point
example, and for every positive Lipschitz constant keeps what is refutable in the population; on
i.i.d.\ streams, a finite-dimensional online procedure turns the stable bound into fixed-horizon and
anytime-valid certificates. On four real predictors every confirmatory certificate was positive. A
positive certificate refutes, on its bundle and at its level, the calibration notion it bounds and
shows an auditor where to spend labels; a certificate of zero refutes nothing, and the labelled
audits it leaves in place are the ones that can find coherent errors and shifts of the labels alone.

\clearpage
\appendix

\section{Proofs for Section~\ref{sec:identification}}
\label{app:proofs-identification}

Throughout, $P \in \mathcal{P}(P_F)$ is a joint law of $(F, Z)$ on $[0,1]^m \times \mathcal{Z}$,
where $\mathcal{Z} = \{z^\ell : \ell \in \mathcal{W}\}$, and $\mathbb{E}$ and $\Pr$ refer to $P$.
Since $F$ is the first coordinate, every set in $\sigma(F)$ is $\{F \in B\}$ for a unique Borel set
$B \subseteq [0,1]^m$, and its probability $P_F(B)$ is the same under every $P \in
\mathcal{P}(P_F)$; in particular the null sets of $\mathcal{H}_i \subseteq \sigma(F)$ do not depend
on $P$. A stake $u \in \mathcal{U}_{\mathcal{H}}$ is a function of $F$, $u_i = g_i(F)$ with $g_i$
Borel, so $d^w_u(F)$ is a random variable with values in $[-1, 1]$; stakes that agree almost surely
have the same expected value and are identified. For a complement pair $(A, \neg A)$ with weights
$(w_A, w_{\neg A})$, $\mathcal{C} = \{c \in [0,1]^2 : c_A + c_{\neg A} = 1\}$ and
\begin{equation}
  \operatorname{dist}^w_1(v, \mathcal{C}) = \min\{w_A, w_{\neg A}\} \, \lvert v_A + v_{\neg A} - 1
  \rvert \qquad (v \in [0,1]^2):
  \label{eq:complement-distance}
\end{equation}
for $c \in \mathcal{C}$, $w_A \lvert v_A - c_A \rvert + w_{\neg A} \lvert v_{\neg A} - c_{\neg A}
\rvert \ge \min\{w_A, w_{\neg A}\} \lvert v_A + v_{\neg A} - 1 \rvert$ by the triangle inequality,
and if $w_A \le w_{\neg A}$ the point $c = (1 - v_{\neg A}, v_{\neg A}) \in \mathcal{C}$ attains it
(symmetrically otherwise).

\begin{lemma}[Weak duality]
\label{lem:weak-duality}
Let $P_F$ be any law on $[0,1]^m$, let $\mathcal{H}_i \subseteq \sigma(F)$ be any
sub-$\sigma$-algebras and $u \in \mathcal{U}_{\mathcal{H}}$. Then for every $P \in
\mathcal{P}(P_F)$,
\[
  \mathbb{E}\bigl[d^w_u(F)\bigr] \le \mathbb{E} \sum_{i=1}^m w_i \, u_i \, (F_i - Z_i) \le
  \Gamma_{\mathcal{H}}(P) .
\]
\end{lemma}

\begin{proof}
Since $Z \in \mathcal{Z}$, the minimum over worlds in $d^w_u(F)$ is at most its value at the
realised world, so $d^w_u(F) \le \sum_i w_i u_i (F_i - Z_i)$ pointwise; both sides are bounded, and
taking expectations gives the first inequality. Since $u_i$ is bounded and
$\mathcal{H}_i$-measurable, $\mathbb{E}[u_i (F_i - Z_i)] = \mathbb{E}\bigl[u_i \, \mathbb{E}[F_i -
Z_i \mid \mathcal{H}_i]\bigr] \le \mathbb{E}\bigl\lvert \mathbb{E}[F_i - Z_i \mid \mathcal{H}_i]
\bigr\rvert$, because $\lvert u_i \rvert \le 1$. Multiplying by $w_i > 0$ and summing gives the
second.
\end{proof}

\begin{lemma}[Finite support]
\label{lem:finite-support}
Let $P_F$ have finite support $V = \{v^1, \dots, v^N\}$ with $\pi_s = P_F(\{v^s\}) > 0$.
\begin{enumerate}
  \item[(i)] For every $i$ there is a partition $\{A_{i1}, \dots, A_{iK_i}\}$ of $\{1, \dots, N\}$
  such that $\mathcal{H}_i$ and the $\sigma$-algebra $\mathcal{G}_i$ generated by the events $E_{ia}
  = \{F \in \{v^s : s \in A_{ia}\}\}$ agree up to null sets: each set of one differs from a set of
  the other by a null set. We write $a_i(s)$ for the cell that contains $s$.
  \item[(ii)] Up to almost sure equality, the stakes $u_i$ that are $\mathcal{H}_i$-measurable with
  $\lvert u_i \rvert \le 1$ are the functions of the cell, $u_i = u_{i a_i(s)}$ on $\{F = v^s\}$,
  with $(u_{ia})_a \in [-1, 1]^{K_i}$.
  \item[(iii)] Let $X = \{x \in \mathbb{R}_{\ge 0}^{N \times \mathcal{W}} : \sum_\ell x_{s\ell} =
  \pi_s \text{ for all } s\}$. For $x \in X$, the law $P_x$ with $P_x(F = v^s, Z = \zeta) =
  \sum_{\ell : z^\ell = \zeta} x_{s\ell}$ lies in $\mathcal{P}(P_F)$, every $P \in \mathcal{P}(P_F)$
  is some $P_x$, and
  \[
    \Gamma_{\mathcal{H}}(P_x) = \sum_{i=1}^m w_i \sum_{a=1}^{K_i} \lvert c_{ia}(x) \rvert ,
    \qquad c_{ia}(x) = \sum_{s \in A_{ia}} \sum_{\ell \in \mathcal{W}} x_{s\ell} \, (v^s_i -
    z^\ell_i) .
  \]
\end{enumerate}
\end{lemma}

\begin{proof}
(i) A set $\{F \in B\}$ agrees almost surely with $\{F \in B \cap V\}$. The traces $B \cap V$ of the
sets $\{F \in B\} \in \mathcal{H}_i$ are closed under complements and unions, as $\mathcal{H}_i$ is,
so they form an algebra of subsets of the finite set $V$. It is generated by its atoms, which
partition $V$; let $A_{ia}$ index the atoms. Each atom is the trace of a set in $\mathcal{H}_i$,
which agrees almost surely with $E_{ia}$, and each set in $\mathcal{H}_i$ agrees almost surely with
the union of the $E_{ia}$ over the atoms in its trace. (ii) By (i), for every Borel set $C$ the
event $\{u_i \in C\}$ agrees almost surely with a union of $E_{ia}$'s. As each $E_{ia}$ has positive
probability, $\Pr(u_i \in C \mid E_{ia}) \in \{0, 1\}$ for every $C$, so $u_i$ is almost surely
constant on $E_{ia}$, with a value $u_{ia} \in [-1, 1]$. Conversely, by (i) a function of the cell
agrees almost surely with an $\mathcal{H}_i$-measurable function, which can be clipped to $[-1, 1]$.
(iii) A law with first marginal $P_F$ is carried by $V \times \mathcal{Z}$ and is determined by its
masses $p_{s\zeta} \ge 0$, with $\sum_\zeta p_{s\zeta} = \pi_s$; choosing one world $\ell(\zeta)$
with $z^{\ell(\zeta)} = \zeta$ for each $\zeta$ and setting $x_{s\ell(\zeta)} = p_{s\zeta}$, and
$x_{s\ell} = 0$ otherwise, gives $x \in X$ with $P_x = P$. Conversely $P_x$ has first marginal
$P_F$. By (i), for integrable $Y$, $\mathbb{E}[Y \mid \mathcal{H}_i] = \mathbb{E}[Y \mid
\mathcal{G}_i] = \sum_a \mathbf{1}_{E_{ia}} \, \mathbb{E}[Y \mathbf{1}_{E_{ia}}] / \Pr(E_{ia})$
almost surely, so $\mathbb{E} \lvert \mathbb{E}[Y \mid \mathcal{H}_i] \rvert = \sum_a \lvert
\mathbb{E}[Y \mathbf{1}_{E_{ia}}] \rvert$. For $Y = F_i - Z_i$ under $P_x$, $\mathbb{E}[Y
\mathbf{1}_{E_{ia}}] = c_{ia}(x)$.
\end{proof}

\begin{proof}\textbf{of Theorem~\ref{thm:minimax}.}
Use the partitions of Lemma~\ref{lem:finite-support}. By part~(iii), $\inf_{P \in \mathcal{P}(P_F)}
\Gamma_{\mathcal{H}}(P)$ is the optimal value of the linear program
\[
  \text{(P)} \qquad \min_{x \in X, \; t} \; \sum_{i=1}^m w_i \sum_{a=1}^{K_i} t_{ia} \quad
  \text{subject to} \quad t_{ia} - c_{ia}(x) \ge 0, \quad t_{ia} + c_{ia}(x) \ge 0 ,
\]
since for fixed $x$ the best $t_{ia}$ is $\lvert c_{ia}(x) \rvert$. (P) is feasible and its
objective is nonnegative on the feasible set, so it has an optimal solution $(x^*, t^*)$, and
$P_{x^*}$ attains the infimum. By the duality theorem of linear programming its dual has an optimal
solution with the same value. With multipliers $\alpha_{ia}, \beta_{ia} \ge 0$ for the two families
of inequalities and $r_s$ for the equalities $\sum_\ell x_{s\ell} = \pi_s$, the dual is
\[
  \begin{aligned}
    \text{(D)} \qquad \max_{\alpha, \beta \ge 0, \; r} \; \sum_{s=1}^N \pi_s r_s \quad
    \text{subject to} \quad & \alpha_{ia} + \beta_{ia} = w_i , \\
    & r_s \le \sum_{i=1}^m (\alpha_{i a_i(s)} - \beta_{i a_i(s)}) (v^s_i - z^\ell_i) \text{
    for all } s, \ell :
  \end{aligned}
\]
the equality is the dual constraint of the free variable $t_{ia}$, and the inequality that of
$x_{s\ell} \ge 0$, whose objective coefficient is zero. As $(\alpha_{ia}, \beta_{ia})$ ranges over
the nonnegative pairs with sum $w_i$, $u_{ia} = (\alpha_{ia} - \beta_{ia}) / w_i$ ranges over $[-1,
1]$, with inverse $\alpha_{ia} = w_i (1 + u_{ia})/2$, $\beta_{ia} = w_i (1 - u_{ia})/2$. For given
$u$ the largest feasible $r_s$ is
\[
  \min_{\ell \in \mathcal{W}} \sum_{i=1}^m w_i \, u_{i a_i(s)} \, (v^s_i - z^\ell_i) =
  d^w_u(v^s) ,
\]
where $u$ also denotes the stake that takes the value $u_{i a_i(s)}$ on $\{F = v^s\}$. So the
optimal value of (D) is the maximum of $\sum_s \pi_s d^w_u(v^s) = \mathbb{E}[d^w_u(F)]$ over
$(u_{ia}) \in \prod_i [-1, 1]^{K_i}$, which by Lemma~\ref{lem:finite-support}(ii) is $\sup_{u \in
\mathcal{U}_{\mathcal{H}}} \mathbb{E}[d^w_u(F)]$, and an optimal solution of (D) gives a stake that
attains it.
\end{proof}

\begin{proof}\textbf{of Lemma~\ref{lem:mean-coherence}.}
Since $\Pr(Z_i = 1 \mid F_i) = F_i$, the tower property gives $\mathbb{E}F_i = \Pr(Z_i = 1) =
\mathbb{E}Z_i$, and $\mathbb{E}Z = \sum_{\zeta \in \mathcal{Z}} \Pr(Z = \zeta) \, \zeta \in
\mathcal{C}$. A probability measure $\mu$ on $\mathcal{W}$ gives $(\mu(B_i))_i = \sum_\ell
\mu(\{\ell\}) z^\ell$, and every point of $\mathcal{C}$ arises in this way, so a linear equality or
inequality that holds for $(\mu(B_i))_i$ under every $\mu$ holds on $\mathcal{C}$, in particular at
$\mathbb{E}F$. The converse fails in two ways: in Counterexample~\ref{cx:coherent}, $\mathbb{E}F \in
\mathcal{C}$ while the true law has $\Theta_w > 0$; and in Theorem~\ref{thm:separation},
$\mathbb{E}F \in \mathcal{C}$ while $L_{\mathrm{marg}} > 0$, so $\Theta_w(P) \ge \tfrac1{16}$ for
every compatible $P$.
\end{proof}

\begin{proof}\textbf{of Counterexample~\ref{cx:pointwise}.}
Under the stated law $\Pr(A \mid F_A) = F_A$, so $\mathrm{ECE}_1(F_A) = 0$; and $F_{\neg A} \equiv
\tfrac12 = \Pr(\neg A)$, so $\mathrm{ECE}_1(F_{\neg A}) = 0$ and $\Theta_w = 0$. At both support
points $\lvert F_A + F_{\neg A} - 1 \rvert = \tfrac12$, so \eqref{eq:complement-distance} with equal
weights gives $\operatorname{dist}^w_1(F, \mathcal{C}) = \tfrac14$ at both. Under the same law
$\Pr(\neg A \mid F) = 1 - F_A$, so the joint calibration error is $\tfrac12 \, \mathbb{E} \lvert
\tfrac12 - (1 - F_A) \rvert = \tfrac14$.
\end{proof}

\begin{proof}\textbf{of Counterexample~\ref{cx:coherent}.}
Suppose $F \in \mathcal{C}$ almost surely, and let $u$ be any stake that is a function of $F$ with
$\lvert u_i \rvert \le 1$. Writing $F = \sum_\ell \lambda_\ell z^\ell$ with $\lambda$ a probability
vector, the $\lambda$-average over worlds of $\sum_i w_i u_i(F) (F_i - z^\ell_i)$ is $\sum_i w_i
u_i(F) (F_i - F_i) = 0$, so $d^w_u(F) \le 0$. Hence $L_H$ and $L_{H,G}$
(Sections~\ref{sec:stability} and~\ref{sec:computation}), suprema of $\mathbb{E}[d^w_u(F)]$ over
classes that contain $u = 0$, are $0$. So are $L_0 = \operatorname{dist}^w_1(\mathbb{E}F,
\mathcal{C})$, as $\mathbb{E}F \in \mathcal{C}$, and $L_{\mathrm{joint}} =
\mathbb{E}\operatorname{dist}^w_1(F, \mathcal{C})$. For $L_{\mathrm{marg}}(P_F) = \inf_{P}
\Theta_w(P)$, let $\lambda \colon \mathcal{C} \to \Delta(\mathcal{W})$ be a Borel map with
$\sum_\ell \lambda_\ell(c) z^\ell = c$; for instance the barycentric coordinates in a triangulation
of $\mathcal{C}$ whose vertices are among the $z^\ell$, which are affine on the relative interior of
each of its finitely many simplices. Let $P$ draw $Z = z^\ell$ with probability $\lambda_\ell(F)$
given $F$. Then $\mathbb{E}[Z \mid F] = F$, so $\Pr(Z_i = 1 \mid F_i) =
\mathbb{E}\bigl[\mathbb{E}[Z_i \mid F] \bigm| F_i\bigr] = F_i$ and $\Theta_w(P) = 0$; more generally
$\mathbb{E}[F_i - Z_i \mid \mathcal{H}_i] = 0$ for every $\mathcal{H}_i \subseteq \sigma(F)$, so
every information-restricted residual of $P$ is $0$, and so is every bound of
Theorem~\ref{thm:minimax}. Yet $\Theta_w$ can be large under the true law. For a complement pair
with equal weights, let $F$ take $(0.3, 0.7)$ and $(0.7, 0.3)$ with probability $\tfrac12$ each
while $A$ is a fair coin independent of $F$: both values lie in $\mathcal{C}$, and $\Theta_w = 0.2$.
If $F \equiv (1, 0)$ while the true label is never in $A$, then $\Theta_w = 1$, its largest possible
value, since $\mathrm{ECE}_1 \le 1$ and $\sum_i w_i = 1$.
\end{proof}

\begin{proof}\textbf{of Corollary~\ref{cor:stages}.}
\emph{Mean stage.} With trivial $\mathcal{H}_i$, $\Gamma_{\mathcal{H}}(P) = \sum_i w_i \lvert
\mathbb{E}F_i - \mathbb{E}Z_i \rvert$. As $P$ ranges over $\mathcal{P}(P_F)$, $\mathbb{E}Z$ ranges
over all of $\mathcal{C}$: $\mathbb{E}Z \in \mathcal{C}$ as in the proof of
Lemma~\ref{lem:mean-coherence}, and for $c = \sum_\ell \lambda_\ell z^\ell$ the law under which $Z =
z^\ell$ with probability $\lambda_\ell$, independently of $F$, has $\mathbb{E}Z = c$. So the infimum
is $\min_{c \in \mathcal{C}} \sum_i w_i \lvert \mathbb{E}F_i - c_i \rvert =
\operatorname{dist}^w_1(\mathbb{E}F, \mathcal{C})$; this part does not use the finite support.
\emph{Joint stage.} With $\mathcal{H}_i = \sigma(F)$, $\Gamma_{\mathcal{H}}(P) = \mathbb{E} \sum_i
w_i \lvert F_i - \mathbb{E}[Z_i \mid F] \rvert$, and $\mathbb{E}[Z \mid F = v^s] = \sum_\ell
(x_{s\ell}/\pi_s) z^\ell \in \mathcal{C}$ in the notation of Lemma~\ref{lem:finite-support}, so
$\Gamma_{\mathcal{H}}(P) \ge \mathbb{E} \operatorname{dist}^w_1(F, \mathcal{C})$. Choosing for each
$s$ a nearest point $c^s = \sum_\ell \lambda^s_\ell z^\ell$ of the compact set $\mathcal{C}$ and
$x_{s\ell} = \pi_s \lambda^s_\ell$ attains equality. \emph{Marginal stage.} $\Gamma_{\mathcal{H}}$
with $\mathcal{H}_i = \sigma(F_i)$ is $\Theta_w$, since $\mathbb{E}[F_i - Z_i \mid F_i] = F_i -
\Pr(Z_i = 1 \mid F_i)$, and its infimum is the value of the linear program of
Theorem~\ref{thm:minimax}. \emph{Order.} If $\mathcal{G} \subseteq \mathcal{H}$, then for integrable
$Y$, conditional Jensen gives $\mathbb{E}\lvert \mathbb{E}[Y \mid \mathcal{G}] \rvert = \mathbb{E}
\bigl\lvert \mathbb{E}\bigl[\mathbb{E}[Y \mid \mathcal{H}] \bigm| \mathcal{G}\bigr] \bigr\rvert \le
\mathbb{E} \lvert \mathbb{E}[Y \mid \mathcal{H}] \rvert$. The trivial $\sigma$-algebra,
$\sigma(F_i)$ and $\sigma(F)$ are nested, so the mean, marginal and joint residuals are ordered for
every $P$, and so are their infima. Each bound is the infimum, so it is sharp by definition, and it
is attained by Theorem~\ref{thm:minimax}; in particular $L_{\mathrm{marg}} \le \Theta_w(P)$ for
every $P \in \mathcal{P}(P_F)$.
\end{proof}

\begin{proof}\textbf{of Corollary~\ref{cor:special}.}
(a) As $A$ is neither sure nor impossible, some world lies in $A$ and some outside it, so
$\mathcal{Z} = \{0^m, 1^m\}$. For $u \in \mathcal{U}_{\mathrm{marg}}$, $a_i = w_i u_i$ is a
measurable function of $F_i$ with values in $[-w_i, w_i]$, every such function arises, and
\[
  d^w_u(F) = \min\Bigl\{\sum_i a_i(F_i) F_i, \; \sum_i a_i(F_i) (F_i - 1)\Bigr\} = \sum_i
  a_i(F_i) F_i - \max\Bigl\{0, \sum_i a_i(F_i)\Bigr\} ,
\]
so Theorem~\ref{thm:minimax} with $\mathcal{H}_i = \sigma(F_i)$ gives the formula. Suppose
$L_{\mathrm{marg}} = 0$. By Theorem~\ref{thm:minimax} the infimum is attained, so $\Theta_w(P) = 0$
for some $P \in \mathcal{P}(P_F)$, and as $w_i > 0$, $\Pr(Z_i = 1 \mid F_i) = F_i$ for every $i$.
Under $P$ all $Z_i$ equal one indicator $Z_A$, and $p = \Pr(Z_A = 1) = \mathbb{E}F_i$. If $p = 0$,
then $F_i = \Pr(Z_A = 1 \mid F_i) = 0$ for every $i$, so $F \equiv 0^m$; likewise $p = 1$ gives $F
\equiv 1^m$. If $0 < p < 1$, the information structure with state $\omega = Z_A$, prior $p$ and
signals $s_i = F_i$ gives agent $i$ the posterior $\Pr(\omega = 1 \mid s_i) = F_i$, so $P_F$ is the
law of the posterior beliefs of $m$ agents with the common prior $p$. Conversely, if $F \equiv 0^m$
(or $1^m$), the law with $Z \equiv 0^m$ (or $1^m$) has $\Theta_w = 0$. If $F_i = \Pr(\omega = 1 \mid
s_i)$ for a binary state $\omega$ with $\Pr(\omega = 1) = p$ and signals $s_1, \dots, s_m$ on some
probability space, with $F \sim P_F$, the law $P$ of $(F, (\omega, \dots, \omega))$ lies in
$\mathcal{P}(P_F)$, and since $F_i$ is a function of $s_i$, $\Pr(\omega = 1 \mid F_i) =
\mathbb{E}[\Pr(\omega = 1 \mid s_i) \mid F_i] = F_i$, so $\Theta_w(P) = 0$. In every case
$L_{\mathrm{marg}} \le \Theta_w(P) = 0$. The integrand above is positively homogeneous in $a$ and
vanishes at $a = 0$, and $w_i \le 1$, so $L_{\mathrm{marg}}$, the supremum of its expectation over
$\lvert a_i \rvert \le w_i$, is $0$ exactly when that expectation is nonpositive for every $a$ with
$a_i \colon [0,1] \to [-1, 1]$ (a trading scheme in their terms), which is condition~(8) of
\citet[Theorem~3]{arieli2021feasible}. With the characterisation just proved, this gives their
Theorem~3 for finitely supported $P_F$ other than $\delta_{0^m}$ and $\delta_{1^m}$; these two
satisfy~(8) but have no prior in $(0, 1)$.

(b) Apply Theorem~\ref{thm:minimax} to $P_F = \delta_v$. As $F = v$ almost surely, every
$\mathcal{H}_i \subseteq \sigma(F)$ is trivial up to null sets, the stakes are the constants $u \in
[-1, 1]^m$, and $\mathbb{E}[d^w_u(F)] = d^w_u(v)$. A law in $\mathcal{P}(\delta_v)$ is the product
of $\delta_v$ with a law of $Z$, whose mean ranges over $\mathcal{C}$, and $\Gamma_{\mathcal{H}}(P)
= \sum_i w_i \lvert v_i - \mathbb{E}Z_i \rvert$, so the infimum is $\operatorname{dist}^w_1(v,
\mathcal{C})$, and the maximum over $u$ is attained. By Corollary~\ref{cor:stages},
$L_{\mathrm{joint}}$ is the expectation of this distance.

(c) The coordinate $(j, k)$ asks $\{Y = k\}$ with $F_{(j,k)} = Q_{jk}$, and the world $y$ has
$z^y_{(j,k)} = \mathbf{1}\{y = k\}$, so $Z_{(j,k)} = \mathbf{1}\{Y = k\} = Y_k$ for the one-hot $Y$.
With weights $1/(2K)$ and $\mathcal{H}_{(j,k)} = \sigma(Q_j) \subseteq \sigma(F)$,
\[
  \Gamma_{\mathcal{H}}(P) = \frac{1}{2K} \sum_{j=1}^2 \sum_{k=1}^K \mathbb{E} \bigl\lvert
  \mathbb{E}[Q_{jk} - Y_k \mid Q_j] \bigr\rvert = \frac{1}{2K} \sum_{j=1}^2 \mathbb{E} \bigl\lVert
  Q_j - \mathbb{E}[Y \mid Q_j] \bigr\rVert_1 .
\]
A law in $\mathcal{P}(P_F)$ is a joint law of $(Q_1, Q_2, Y)$ whose $(Q_1, Q_2)$-marginal is $P_F$,
which is the set over which $\mathrm{MinECE}$ is minimised. Hence the infimum is
$2\,\mathrm{MinECE}/(2K) = \mathrm{MinECE}/K$, and Theorem~\ref{thm:minimax} gives the dual form. As
$\sigma(Q_{jk}) \subseteq \sigma(Q_j) \subseteq \sigma(F)$, the order argument in the proof of
Corollary~\ref{cor:stages} gives $L_{\mathrm{marg}} \le \mathrm{MinECE}/K \le L_{\mathrm{joint}}$;
for $K = 2$, $\sigma(Q_j) = \sigma(Q_{j1})$ since $Q_{j2} = 1 - Q_{j1}$. That this is an equality,
where \citet{kong2026calibration} have a bracket within a factor of $2$, comes from the norm on the
bets. For one report $q$ and a law $p$ of $Y$, stakes $u \in [-1, 1]^K$ earn $\sum_k u_k (q_k -
p_k)$ in expectation, whose maximum is $\lVert q - p \rVert_1$; the same bets written as zero-mean
payoffs $\langle u, q \rangle - u_y$ can be as large as $2$ in absolute value, and restricting the
payoffs to $[-1, 1]$ instead leaves a maximum between $\tfrac12 \lVert q - p \rVert_1$ and $\lVert q
- p \rVert_1$ \citep[Lemma~D.4]{kong2026calibration}. Finally, this bound need not be a lower bound
on $\Theta_w$. Let $K = 3$ and let $(Q_1, Q_2, Y)$ take the values $(e_2, (0, \tfrac12, \tfrac12),
2)$, $(e_3, (\tfrac12, 0, \tfrac12), 3)$ and $(e_1, (\tfrac12, \tfrac12, 0), 1)$ with probability
$\tfrac13$ each. Under this law every scalar component is marginally calibrated, so $\Theta_w = 0$.
Each forecast vector identifies the state, so under any compatible law $\mathbb{E}[Y \mid Q_1]$ and
$\mathbb{E}[Y \mid Q_2]$ are the same vector $p$ at each state, and $\lVert Q_1 - p \rVert_1 +
\lVert Q_2 - p \rVert_1 \ge \lVert Q_1 - Q_2 \rVert_1 = 1$, with equality at $p = Q_1$; hence
$\mathrm{MinECE} = \tfrac12$ and the bound is $\tfrac16$.
\end{proof}

\begin{proof}\textbf{of Theorem~\ref{thm:separation}.}
The weights are $\tfrac12$ and $\mathcal{Z} = \{(1, 0), (0, 1)\}$. Since $\mathbb{E}F = (\tfrac12,
\tfrac12) \in \mathcal{C}$, $L_0 = 0$. By~\eqref{eq:complement-distance} the four values have
$\operatorname{dist}^w_1(F, \mathcal{C}) = \tfrac12 \lvert F_A + F_{\neg A} - 1 \rvert = \tfrac14,
\tfrac14, \tfrac18, \tfrac18$, so $L_{\mathrm{joint}} = \tfrac3{16}$. For the upper bound on
$L_{\mathrm{marg}}$, let $A$ occur exactly when $F_A = 1$. Then $F_A$ is calibrated. $F_{\neg A} =
\tfrac12$ at the first two values, where $\neg A$ occurs with conditional probability $\tfrac12$;
$F_{\neg A} = \tfrac34$ at the third, where $\neg A$ occurs; $F_{\neg A} = \tfrac14$ at the fourth,
where it does not. So $\mathrm{ECE}_1(F_{\neg A}) = \tfrac14 \cdot \tfrac14 + \tfrac14 \cdot
\tfrac14 = \tfrac18$ and $\Theta_w = \tfrac1{16}$. For the lower bound, take $u_A(0) = -1$, $u_A(1)
= 1$, $u_{\neg A}(\tfrac12) = 0$, $u_{\neg A}(\tfrac34) = -1$ and $u_{\neg A}(\tfrac14) = 1$, each a
function of its own component. At $(0, \tfrac12)$ the two worlds give $\tfrac12$ (world $A$) and $0$
(world $\neg A$), and at $(1, \tfrac12)$ they give $0$ and $\tfrac12$, so $d^w_u = 0$ there. At $(0,
\tfrac34)$ world $A$ gives $\tfrac12\{(-1)(0 - 1) + (-1)(\tfrac34 - 0)\} = \tfrac18$ and world $\neg
A$ gives $\tfrac12 (-1)(\tfrac34 - 1) = \tfrac18$; at $(1, \tfrac14)$ both give $\tfrac18$ in the
same way. So $\mathbb{E}[d^w_u(F)] = \tfrac14 (0 + 0 + \tfrac18 + \tfrac18) = \tfrac1{16}$, and by
Lemma~\ref{lem:weak-duality}, $\Theta_w(P) \ge \tfrac1{16}$ for every $P \in \mathcal{P}(P_F)$.
Hence $L_{\mathrm{marg}} = \tfrac1{16}$.
\end{proof}

\begin{proof}\textbf{of Proposition~\ref{prop:attribution}.}
Write $\Theta_J(P) = \sum_{i \in J} w_i \, \mathrm{ECE}_1(F_i)$. Since $\mathrm{ECE}_1(F_i) =
\mathbb{E} \lvert F_i - \Pr(Z_i = 1 \mid F_i) \rvert$ depends only on the law of $(F_i, Z_i)$,
$\Theta_J(P)$ depends only on the law of $(F_J, Z_J)$. For $P \in \mathcal{P}(P_F)$ that law has
first marginal $P_{F_J}$ and $Z_J \in \{z^\ell_J\}$, so it lies in $\mathcal{P}_J(P_{F_J})$ with the
same value; hence the left side is at least the right. Conversely let $Q \in
\mathcal{P}_J(P_{F_J})$. As $Z_J$ takes finitely many values, there is a kernel $\kappa(f, \zeta) =
Q(Z_J = \zeta \mid F_J = f)$, Borel in $f$. For each restricted vector $\zeta$ choose a world
$\ell(\zeta)$ with $z^{\ell(\zeta)}_J = \zeta$, and let $P$ draw $F \sim P_F$, then $\zeta$ from
$\kappa(F_J, \cdot)$, and set $Z = z^{\ell(\zeta)}$. Then $P \in \mathcal{P}(P_F)$, $Z_J = \zeta$,
and $(F_J, Z_J)$ has law $Q$ under $P$, so $\Theta_J(P)$ is the value of $Q$; hence the left side is
at most the right. If $J = \{i\}$ and component $i$ asks an event that is neither sure nor
impossible, the restricted worlds are $\{0, 1\}$, and the law under which $Z_i = 1$ with probability
$F_i$ given $F_i$ has $\Pr(Z_i = 1 \mid F_i) = F_i$, so both sides are $0$. The argument uses only
that the objective depends on the law of $(F_J, Z_J)$, so it applies unchanged to the mean residual
of the targets, to their residual conditioned on $F_J$, and to their smooth residual.
\end{proof}

\section{Proofs for Section~\ref{sec:stability}}
\label{app:proofs-stability}

We keep the conventions of Appendix~\ref{app:proofs-identification}. For $u \in \mathcal{U}_H$ and
$P \in \mathcal{P}(P_F)$ let
\[
  \Psi(P, u) = \mathbb{E} \sum_{i=1}^m w_i \, u_i(F_i) \, (F_i - Z_i) ,
\]
so that $\Gamma_H(P) = \sup_{u \in \mathcal{U}_H} \Psi(P, u)$. The component $u_i(F_i)$ of a stake
in $\mathcal{U}_H$ is $\sigma(F_i)$-measurable with $\lvert u_i \rvert \le 1$, and $\sigma(F_i)
\subseteq \sigma(F)$, so $\mathcal{U}_H \subseteq \mathcal{U}_{\mathrm{marg}}$, and
Lemma~\ref{lem:weak-duality} with $\mathcal{H}_i = \sigma(F_i)$ gives, for any $P_F$,
\begin{equation}
  \mathbb{E}\bigl[d^w_u(F)\bigr] \le \Psi(P, u) \le \Gamma_H(P) \le \Theta_w(P) \qquad (u \in
  \mathcal{U}_H, \; P \in \mathcal{P}(P_F)) ,
  \label{eq:smooth-weak-duality}
\end{equation}
the last inequality because $\Psi(P, u) \le \Theta_w(P)$ for every $u \in \mathcal{U}_H$. Taking the
supremum over $u$, $L_H(P_F) \le \Gamma_H(P) \le \Theta_w(P)$ for every $P$, so $L_H \le
L_{\mathrm{marg}}$; and $L_H \ge 0$, as $u = 0$ lies in $\mathcal{U}_H$. The classes $\mathcal{U}_H$
grow with $H$, so $\Gamma_H$ and $L_H$ are nondecreasing in $H$. Every constant stake lies in every
$\mathcal{U}_H$, and for a constant stake $\Psi(P, u) = \sum_i w_i u_i \, \mathbb{E}(F_i - Z_i)$, so
$\Gamma_0$ is the mean residual.

\begin{lemma}[Lipschitz extension]
\label{lem:extension}
Let $D \subseteq [0,1]$ be finite and let $g \colon D \to [-1, 1]$ satisfy $\lvert g(x) - g(y)
\rvert \le H \lvert x - y \rvert$ for $x, y \in D$. Then $g$ extends to an $H$-Lipschitz function
from $[0,1]$ to $[-1, 1]$. Consequently, if $P_F$ has finite support $\{v^1, \dots, v^N\}$ with
masses $\pi_s$, and $D_i$ is the set of values that $F_i$ takes on it, then $L_H(P_F)$ is the
optimal value of the finite linear program that maximises $\sum_s \pi_s r_s$ over $u_i(x) \in [-1,
1]$ ($x \in D_i$) and $r_s \in \mathbb{R}$ subject to
\begin{gather*}
  \lvert u_i(x') - u_i(x) \rvert \le H (x' - x) \quad \text{for consecutive values } x < x'
  \text{ of } D_i , \\
  r_s \le \sum_{i=1}^m w_i \, u_i(v^s_i) \, (v^s_i - z^\ell_i) \quad \text{for all } s, \ell ,
\end{gather*}
and the supremum defining $L_H$ is attained.
\end{lemma}

\begin{proof}
If $D$ is empty, take the zero function. Otherwise let $x_1 < \dots < x_k$ be its points, and let
$\bar g$ equal $g(x_1)$ on $[0, x_1]$ and $g(x_k)$ on $[x_k, 1]$ and interpolate linearly between
consecutive points. Its values lie between values of $g$, hence in $[-1, 1]$, and it is continuous
and piecewise linear with slopes of absolute value at most $H$, hence $H$-Lipschitz. For the
consequence, $\mathbb{E}[d^w_u(F)] = \sum_s \pi_s d^w_u(v^s)$ depends on $u$ only through the values
$u_i(x)$, $x \in D_i$. The values of a stake in $\mathcal{U}_H$ satisfy the constraints. Conversely,
values that satisfy them for consecutive points satisfy $\lvert u_i(x') - u_i(x) \rvert \le H \lvert
x' - x \rvert$ for all $x, x' \in D_i$ by the triangle inequality, and extend to a stake in
$\mathcal{U}_H$ by the first part. For given values the largest feasible $r_s$ is $d^w_u(v^s)$, so
the optimal value is $L_H(P_F)$; it is attained, because $\sum_s \pi_s d^w_u(v^s)$ is continuous in
the values, which range over a compact set.
\end{proof}

\begin{proof}\textbf{of Proposition~\ref{prop:discontinuity}.}
The weights are $\tfrac12$ and $\mathcal{Z} = \{(1, 0), (0, 1)\}$. At $\varepsilon = 0$ the law of
Counterexample~\ref{cx:pointwise} has $\Theta_w = 0$, so $L_{\mathrm{marg}}(P_{F^0}) = 0$. Let
$\varepsilon > 0$, and let $v^1 = (0, \tfrac12 - \varepsilon)$ and $v^2 = (1, \tfrac12 +
\varepsilon)$ be the support points. Each component takes different values at $v^1$ and $v^2$, so
$\{F_i = v^s_i\}$ and $\{F = v^s\}$ agree almost surely for both $i$ and $s$, and $\sigma(F_i)$
agrees with $\sigma(F)$ up to null sets. Under every $P \in \mathcal{P}(P_F)$, therefore, $\Pr(Z_i =
1 \mid F_i) = \Pr(Z_i = 1 \mid F)$ almost surely, the marginal and joint residuals of $P$ are equal,
and so are their infima $L_{\mathrm{marg}}$ and $L_{\mathrm{joint}}$ (Corollary~\ref{cor:stages}).
By~\eqref{eq:complement-distance}, $\operatorname{dist}^w_1(v^s, \mathcal{C}) = \tfrac12 \lvert
v^s_A + v^s_{\neg A} - 1 \rvert = \tfrac12 (\tfrac12 + \varepsilon)$ at both points, so
$L_{\mathrm{joint}} = \tfrac14 + \tfrac\varepsilon2$. For the distance, $F^0_{\neg A} \equiv
\tfrac12$, so under every coupling $\mathbb{E}\lvert F^\varepsilon_{\neg A} - F^0_{\neg A} \rvert =
\mathbb{E}\lvert F^\varepsilon_{\neg A} - \tfrac12 \rvert = \varepsilon$, and $W_{1,w} \ge
\tfrac\varepsilon2$. The coupling that pairs the points with the same $F_A$ moves only $F_{\neg A}$,
by $\varepsilon$, so $W_{1,w} = \tfrac\varepsilon2$.
\end{proof}

\begin{proof}\textbf{of Lemma~\ref{lem:no-constraint}.}
Let $R = \{1, \dots, m\} \setminus S$ and $\kappa = \sum_{i \in S} w_i \, \mathbb{E}\lvert F_i - c_i
\rvert$. By assumption $\mathcal{Z}$ is the set of vectors $z$ with $z_i = c_i$ for $i \in S$ and
$z_R \in \{0, 1\}^R$, so $\mathcal{C} = \{x \in [0,1]^m : x_i = c_i \text{ for } i \in S\}$ and
$\operatorname{dist}^w_1(v, \mathcal{C}) = \sum_{i \in S} w_i \lvert v_i - c_i \rvert$. Hence
$L_{\mathrm{joint}} = \kappa$. As $c_i \in \{0, 1\}$ and $F_i \in [0,1]$, the sign of $F_i - c_i$
does not depend on $F_i$, so $\lvert \mathbb{E}F_i - c_i \rvert = \mathbb{E}\lvert F_i - c_i \rvert$
and $L_0 = \operatorname{dist}^w_1(\mathbb{E}F, \mathcal{C}) = \kappa$. The constant stake with $u_i
= 1 - 2c_i$ for $i \in S$ and $u_i = 0$ for $i \in R$ earns $\sum_{i \in S} w_i (1 - 2c_i)(F_i -
c_i) = \sum_{i \in S} w_i \lvert F_i - c_i \rvert$ in every world, so $\mathbb{E}[d^w_u(F)] =
\kappa$, and as it lies in $\mathcal{U}_H$ for every $H \ge 0$, $L_H \ge \kappa$. Conversely, let
$P^*$ draw $F \sim P_F$, set $Z_i = c_i$ for $i \in S$ and, given $F$, draw the $Z_i$ for $i \in R$
independently with $\Pr(Z_i = 1 \mid F) = F_i$. Every vector so drawn lies in $\mathcal{Z}$, so $P^*
\in \mathcal{P}(P_F)$; and $\Pr(Z_i = 1 \mid F)$ is $F_i$ for $i \in R$ and $c_i$ for $i \in S$, so
the joint residual of $P^*$ is $\kappa$. By~\eqref{eq:smooth-weak-duality} and the order of the
residuals (proof of Corollary~\ref{cor:stages}, valid for every $P$), $\kappa \le L_H \le
L_{\mathrm{marg}} \le \Theta_w(P^*) \le \kappa$. So all four bounds equal $\kappa$. For a coupling
$\gamma$ of $P_F$ and $P_{F'}$, $\lvert \kappa(P_F) - \kappa(P_{F'}) \rvert \le \mathbb{E}_\gamma
\sum_{i \in S} w_i \lvert F_i - F'_i \rvert$, and the infimum over $\gamma$ gives the Lipschitz
bound. A bundle of one query that is neither sure nor impossible has $S = \emptyset$ and
$\mathcal{Z} = \{0, 1\}$, so every bound is $0$.
\end{proof}

\begin{proof}\textbf{of Corollary~\ref{cor:closed-form}.}
Keep the notation of the proof of Proposition~\ref{prop:discontinuity}, now with $0 \le \varepsilon
\le \tfrac12$, and let $\beta = \tfrac14 + \tfrac\varepsilon2$. \emph{Upper bound.} Let $P^\dagger$
be the law under which $A$ occurs exactly when $F_A = 1$, so that $Z = (0, 1)$ at $v^1$ and $Z = (1,
0)$ at $v^2$. Then $F_A - Z_A = 0$, and $F_{\neg A} - Z_{\neg A}$ is $-(\tfrac12 + \varepsilon)$ at
$v^1$ and $\tfrac12 + \varepsilon$ at $v^2$. For $u \in \mathcal{U}_H$,
by~\eqref{eq:smooth-weak-duality},
\[
  \mathbb{E}\bigl[d^w_u(F)\bigr] \le \Psi(P^\dagger, u) = \tfrac14 \bigl(\tfrac12 +
  \varepsilon\bigr) \bigl\{u_{\neg A}(\tfrac12 + \varepsilon) - u_{\neg A}(\tfrac12 -
  \varepsilon)\bigr\} \le \beta \min\{1, H\varepsilon\} ,
\]
since the difference is at most $2$ and at most $2H\varepsilon$. \emph{Lower bound.} Let $c =
\min\{1, H\varepsilon\}$ and set $u_A(0) = u_{\neg A}(\tfrac12 - \varepsilon) = -c$ and $u_A(1) =
u_{\neg A}(\tfrac12 + \varepsilon) = c$ (for $\varepsilon = 0$, $c = 0$ and the two values at
$\tfrac12$ agree). As $\varepsilon \le \tfrac12$, $2c \le 2H\varepsilon \le H$, so both functions
are $H$-Lipschitz on their two points, and by Lemma~\ref{lem:extension} they extend to a stake $u
\in \mathcal{U}_H$. At $v^1$, world $A$ gives $\tfrac12 \{(-c)(0 - 1) + (-c)(\tfrac12 -
\varepsilon)\} = \tfrac c2 (\tfrac12 + \varepsilon)$ and world $\neg A$ gives $\tfrac12 (-c)
(\tfrac12 - \varepsilon - 1) = \tfrac c2 (\tfrac12 + \varepsilon)$. At $v^2$, world $A$ gives
$\tfrac12 c (\tfrac12 + \varepsilon)$ and world $\neg A$ gives $\tfrac12 \{c + c(\tfrac12 +
\varepsilon - 1)\} = \tfrac c2 (\tfrac12 + \varepsilon)$. So $d^w_u = c\beta$ at both points, and
$L_H = \beta \min\{1, H\varepsilon\}$. A stake with $u_{\neg A}(\tfrac12 \pm \varepsilon) = \pm c$
attains the upper bound, which is therefore $\Gamma_H(P^\dagger)$, so $P^\dagger$ attains the
infimum of Theorem~\ref{thm:stable}.

For $\varepsilon > 0$, $L_{\mathrm{marg}} = \beta > 0$ by Proposition~\ref{prop:discontinuity}, so
$L_H / L_{\mathrm{marg}} = \min\{1, H\varepsilon\}$. Given $H$ and $c_H > 0$, any $\varepsilon \in
(0, \tfrac12]$ with $H\varepsilon < c_H$ gives $L_H < c_H L_{\mathrm{marg}}$. More generally, let
$B$ be a functional of the laws of bundles with this logical structure and these weights such that
$B(P_{F^0}) = 0$ and $B$ is continuous at $P_{F^0}$ in $W_{1,w}$. As $W_{1,w}(P_{F^\varepsilon},
P_{F^0}) = \tfrac\varepsilon2 \to 0$, $B(P_{F^\varepsilon}) \to 0$, while
$L_{\mathrm{marg}}(P_{F^\varepsilon}) > \tfrac14$ for $\varepsilon > 0$, so $B \ge c \,
L_{\mathrm{marg}}$ fails for every $c > 0$. A label-free lower bound $L$ on $\Theta_w$ satisfies
$L(P_F) \le \Theta_w(P)$ for every $P \in \mathcal{P}(P_F)$, hence $L \le L_{\mathrm{marg}}$, and a
nonnegative one vanishes at $P_{F^0}$.
\end{proof}

\begin{proof}\textbf{of~\eqref{eq:finite-grid}.}
The right inequality is~\eqref{eq:smooth-weak-duality}. If any two distinct values of $F_i$ differ
by at least $\delta$, then $F_i$ takes at most $1 + 1/\delta$ values, so $P_F$ has finite support.
The cells of $\sigma(F_i)$ in Lemma~\ref{lem:finite-support} are the sets of support points with the
same value of $F_i$, so by Theorem~\ref{thm:minimax} with $\mathcal{H}_i = \sigma(F_i)$ and
Lemma~\ref{lem:finite-support}(ii) there are functions $u^*_i \colon D_i \to [-1, 1]$ on the values
$D_i$ of $F_i$ such that the stake $u^* = (u^*_i(F_i))_i$ has $\mathbb{E}[d^w_{u^*}(F)] =
L_{\mathrm{marg}}$. For distinct $x, x' \in D_i$, $\lvert u^*_i(x) - u^*_i(x') \rvert \le 2 \le
(2/\delta) \lvert x - x' \rvert$, so with $\lambda = \min\{1, H\delta/2\}$ the function $\lambda
u^*_i$ is $H$-Lipschitz on $D_i$ with values in $[-1, 1]$, and it extends to $[0,1]$ by
Lemma~\ref{lem:extension}. The minimum over worlds is positively homogeneous, $d^w_{\lambda u^*} =
\lambda \, d^w_{u^*}$, so $L_H \ge \lambda L_{\mathrm{marg}}$. If $H \ge 2/\delta$, then $\lambda =
1$.
\end{proof}

\begin{proof}\textbf{of Theorem~\ref{thm:stable}.}
Let $\Omega = [0,1]^m \times \mathcal{Z}$, a compact metric space, and give the probability measures
on $\Omega$ the weak topology, induced by the continuous functions on $\Omega$. They form a compact
convex subset (Prokhorov's theorem) of the finite signed measures on $\Omega$, a locally convex
space under the same topology. $\mathcal{P}(P_F)$ is nonempty, as $P_F \otimes \delta_{z^\ell}$ lies
in it, and convex. It is closed, since $P \mapsto \int g(f) \, P(df, dz)$ is continuous for every
continuous $g$ on $[0,1]^m$ and a law on $[0,1]^m$ is determined by the integrals of continuous
functions; so it is compact. For $u \in \mathcal{U}_H$ the integrand $(f, z) \mapsto \sum_i w_i
u_i(f_i)(f_i - z_i)$ is continuous on $\Omega$, because each $u_i$ is Lipschitz, so $P \mapsto
\Psi(P, u)$ is affine and continuous. $\mathcal{U}_H$ is a convex subset of $C([0,1])^m$ with the
supremum norm, and $u \mapsto \Psi(P, u)$ is linear with $\lvert \Psi(P, u) - \Psi(P, u') \rvert \le
\sum_i w_i \lVert u_i - u'_i \rVert_\infty$. Sion's minimax theorem
\citep[Corollary~3.3]{sion1958general}, applied with $\mathcal{P}(P_F)$ as the compact side,
therefore gives
\[
  \min_{P \in \mathcal{P}(P_F)} \sup_{u \in \mathcal{U}_H} \Psi(P, u) = \sup_{u \in
  \mathcal{U}_H} \min_{P \in \mathcal{P}(P_F)} \Psi(P, u) ,
\]
with both minima attained: on the left because $\Gamma_H = \sup_u \Psi(\cdot, u)$ is lower
semicontinuous, as a supremum of continuous functions, on a compact set, and on the right because
$\Psi(\cdot, u)$ is continuous. The left side is $\min_P \Gamma_H(P)$. On the right, fix $u$.
By~\eqref{eq:smooth-weak-duality}, $\Psi(P, u) \ge \mathbb{E}[d^w_u(F)]$ for every $P$. Conversely,
list the worlds as $\ell_1, \dots, \ell_n$ and let $\ell^*(f)$ be the first of them that minimises
$\sum_i w_i u_i(f_i)(f_i - z^\ell_i)$; these finitely many functions of $f$ are continuous, so
$\ell^*$ is Borel. The stakes are functions of $F$, so the worst world can be chosen as a function
of $F$: the law of $(F, z^{\ell^*(F)})$ with $F \sim P_F$ lies in $\mathcal{P}(P_F)$, and under it
$\Psi(P, u) = \mathbb{E}[d^w_u(F)]$. So $\min_P \Psi(P, u) = \mathbb{E}[d^w_u(F)]$, the right side
is $L_H(P_F)$, and the first claim follows.

For the second, let $u \in \mathcal{U}_H$, $f, f' \in [0,1]^m$ and $z \in \mathcal{Z}$. Then
\[
  \bigl\lvert u_i(f_i)(f_i - z_i) - u_i(f'_i)(f'_i - z_i) \bigr\rvert \le \lvert u_i(f_i) \rvert
  \, \lvert f_i - f'_i \rvert + \lvert u_i(f_i) - u_i(f'_i) \rvert \, \lvert f'_i - z_i \rvert \le
  (1 + H) \lvert f_i - f'_i \rvert ,
\]
so the profits in the worlds, and hence their minimum $d^w_u$, change by at most $(1 + H)$ times
$\sum_i w_i \lvert f_i - f'_i \rvert$. For a coupling $\gamma$ of $P_F$ and $P_{F'}$ it follows that
\[
  \bigl\lvert \mathbb{E}[d^w_u(F)] - \mathbb{E}[d^w_u(F')] \bigr\rvert \le (1 + H) \,
  \mathbb{E}_\gamma \sum_i w_i \lvert F_i - F'_i \rvert .
\]
The infimum over $\gamma$ bounds this by $(1 + H) \, W_{1,w}(P_F, P_{F'})$, and the suprema over $u$
of two families differ by at most the supremum of their differences.
\end{proof}

\begin{proof}\textbf{of Corollary~\ref{cor:zero-set}.}
If $\Theta_w(P) = 0$ for some $P \in \mathcal{P}(P_F)$, then $0 \le L_{\mathrm{marg}} \le
\Theta_w(P) = 0$; and if $L_{\mathrm{marg}} = 0$, then $0 \le L_H \le L_{\mathrm{marg}} = 0$
by~\eqref{eq:smooth-weak-duality}. Suppose $L_H(P_F) = 0$. By Theorem~\ref{thm:stable} some $P^* \in
\mathcal{P}(P_F)$ has $\Gamma_H(P^*) = 0$. Fix $i$ and a Lipschitz function $g \colon [0,1] \to
\mathbb{R}$ other than zero, with Lipschitz constant $K_g$. As $H > 0$, for $s = \min\{1 / \lVert g
\rVert_\infty, H / K_g\}$ (read $H / K_g$ as $\infty$ if $K_g = 0$) the two stakes with $\pm s g$ in
component $i$ and $0$ elsewhere lie in $\mathcal{U}_H$, so $\pm s \, w_i \,
\mathbb{E}_{P^*}[g(F_i)(F_i - Z_i)] \le \Gamma_H(P^*) = 0$; as $w_i > 0$, $\mathbb{E}_{P^*}[g(F_i)
(F_i - Z_i)] = 0$. Lipschitz functions, polynomials among them, are dense in $C([0,1])$ with the
supremum norm, and $\lvert \mathbb{E}[(g - h)(F_i)(F_i - Z_i)] \rvert \le \lVert g - h
\rVert_\infty$, so the same holds for every continuous $h$. The finite signed measure $\nu_i(B) =
\mathbb{E}_{P^*}[(F_i - Z_i) \mathbf{1}\{F_i \in B\}]$ on $[0,1]$ therefore integrates every
continuous function to zero, so $\nu_i = 0$. That is, $\mathbb{E}_{P^*}[F_i - Z_i \mid F_i] = 0$
almost surely, $\Pr(Z_i = 1 \mid F_i) = F_i$ and $\mathrm{ECE}_1(F_i) = 0$. This holds for every
$i$, so $\Theta_w(P^*) = 0$.
\end{proof}

\section{Proofs for Section~\ref{sec:computation}}
\label{app:proofs-computation}

We keep the conventions of Appendices~\ref{app:proofs-identification}
and~\ref{app:proofs-stability}. On $[t_j, t_{j+1}]$ the interpolation functions $\phi_j$ and
$\phi_{j+1}$ are the two barycentric coordinates and the others vanish, so $\phi_j \ge 0$, $\sum_j
\phi_j = 1$, and $u_{\theta,i}$ is the piecewise-linear function with $u_{\theta,i}(t_j) =
\theta_{ij}$. The profit $\sum_i w_i u_{\theta,i}(f_i)(f_i - z^\ell_i)$ in each world is continuous
in $(\theta, f)$, and so is $d^w_{u_\theta}(f)$, the minimum of these profits; and $L_{H,G} \in [0,
1]$, since $\theta = 0$ lies in $K_{H,G}$ and $d^w_u \le 1$. In the sequential results
$\mathcal{F}_t = \mathcal{F}_0 \vee \sigma(F_1, \dots, F_t)$, where $\mathcal{F}_0$ is generated by
data independent of $F_1, F_2, \dots$ (for the split method, the bundles of the half of the budget
on which its stake is learnt; otherwise the trivial $\sigma$-algebra), and a sequence is predictable
when its $t$-th term is $\mathcal{F}_{t-1}$-measurable. Then $F_t$ is independent of
$\mathcal{F}_{t-1}$ for every $t$, which is all the proofs use.

\begin{proof}\textbf{of Lemma~\ref{lem:grid}.}
On $[t_j, t_{j+1}]$ the function $u_{\theta,i}$ is linear with the values $\theta_{ij},
\theta_{i,j+1} \in [-1, 1]$ at the ends, so its values lie in $[-1, 1]$ and its slope has absolute
value at most $H$; being continuous and piecewise linear with such slopes, it is $H$-Lipschitz. So
$\mathcal{U}_{H,G} \subseteq \mathcal{U}_H$ and $L_{H,G} \le L_H$. Conversely, let $u \in
\mathcal{U}_H$ and $\theta_{ij} = u_i(t_j)$, so that $\theta \in K_{H,G}$. For $x = (1 - \lambda)
t_j + \lambda t_{j+1}$ with $\lambda \in [0, 1]$ and $g_j = t_{j+1} - t_j$,
\[
  \lvert u_i(x) - u_{\theta,i}(x) \rvert = \bigl\lvert (1 - \lambda)\{u_i(x) - u_i(t_j)\} +
  \lambda \{u_i(x) - u_i(t_{j+1})\} \bigr\rvert \le 2 \lambda (1 - \lambda) H g_j \le
  \frac{H\Delta}{2} ,
\]
since $\lvert u_i(x) - u_i(t_j) \rvert \le H\lambda g_j$ and $\lvert u_i(x) - u_i(t_{j+1}) \rvert
\le H(1 - \lambda) g_j$. As $\lvert f_i - z^\ell_i \rvert \le 1$ and $\sum_i w_i = 1$, the profits
in all worlds change by at most $H\Delta/2$ when $u$ is replaced by $u_\theta$, and so does their
minimum: $d^w_{u_\theta}(f) \ge d^w_u(f) - H\Delta/2$ for every $f$. Taking expectations and then
the supremum over $u$ gives $L_{H,G} \ge L_H - H\Delta/2$. The Lipschitz bound is the second part of
the proof of Theorem~\ref{thm:stable}, with the supremum taken over the subclass
$\mathcal{U}_{H,G}$. Finally, $\theta \mapsto \mathbb{E}[d^w_{u_\theta}(F)]$ is continuous on the
compact set $K_{H,G}$, so the supremum defining $L_{H,G}$ is attained.
\end{proof}

\begin{proof}\textbf{of Proposition~\ref{prop:ogd}.}
Write $\langle a, b \rangle_w$ also for the inner product on $\mathbb{R}^{m \times G}$ and $\lVert
\cdot \rVert_w$ for its norm. For each world $\ell$, the profit $\sum_i w_i
u_{\theta,i}(F_{t,i})(F_{t,i} - z^\ell_i)$ equals $\langle h^\ell_t, \theta \rangle_w$ with
$h^\ell_{t,ij} = \phi_j(F_{t,i})(F_{t,i} - z^\ell_i)$, so $f_t(\theta) = \min_\ell \langle h^\ell_t,
\theta \rangle_w$ and $h_t = h^{\ell_t}_t$. Hence $f_t(\theta) \le \langle h_t, \theta \rangle_w$
for every $\theta$, with equality at $\theta_t$, and
\[
  f_t(\theta^*) - f_t(\theta_t) \le \langle h_t, \theta^* - \theta_t \rangle_w :
\]
$h_t$ is a supergradient of $f_t$ at $\theta_t$ in this inner product. The projection onto the
closed convex set $K_{H,G}$ in the norm $\lVert \cdot \rVert_w$ satisfies $\lVert \Pi^w_{K_{H,G}}(a)
- \theta^* \rVert_w \le \lVert a - \theta^* \rVert_w$ for $\theta^* \in K_{H,G}$, so
\[
  \lVert \theta_{t+1} - \theta^* \rVert_w^2 \le \lVert \theta_t + \eta h_t - \theta^* \rVert_w^2 =
  \lVert \theta_t - \theta^* \rVert_w^2 - 2\eta \langle h_t, \theta^* - \theta_t \rangle_w + \eta^2
  \lVert h_t \rVert_w^2 ,
\]
that is, $\langle h_t, \theta^* - \theta_t \rangle_w \le (\lVert \theta_t - \theta^* \rVert_w^2 -
\lVert \theta_{t+1} - \theta^* \rVert_w^2)/(2\eta) + \eta \lVert h_t \rVert_w^2 / 2$. Summed over $t
\le T$, the first terms telescope to at most $\lVert \theta_1 - \theta^* \rVert_w^2 / (2\eta) =
\lVert \theta^* \rVert_w^2 / (2\eta) \le G/(2\eta)$, since $\lvert \theta^*_{ij} \rvert \le 1$ and
$\sum_i w_i = 1$. As $\phi_j \ge 0$ and $\sum_j \phi_j = 1$, $\sum_j \phi_j(x)^2 \le 1$, so $\lVert
h_t \rVert_w^2 = \sum_i w_i (F_{t,i} - z^{\ell_t}_i)^2 \sum_j \phi_j(F_{t,i})^2 \le 1$. Hence
$\sum_{t \le T} \{f_t(\theta^*) - f_t(\theta_t)\} \le G/(2\eta) + \eta T/2$, which is $\sqrt{GT}$ at
$\eta = \sqrt{G/T}$. An update in the direction $(w_i h_{t,ij})_{ij}$, the supergradient in the
Euclidean inner product, is gradient ascent in the Euclidean metric, in which $\lVert \theta^*
\rVert^2$ can be as large as $mG$; the bound above does not hold for it, and in the example of
Section~\ref{sec:computation} with $m = 200$ atoms its regret per item is $0.645$, compared with
$\sqrt{G/T} \approx 0.014$. The same argument applies to constant stakes $\theta \in [-1, 1]^m$,
updated in the direction $h_{t,i} = F_{t,i} - z^{\ell_t}_i$ and projected onto $[-1, 1]^m$: there
$\lVert \theta^* \rVert_w^2 \le 1$ and $\lVert h_t \rVert_w^2 \le 1$, and the bound holds with $G =
1$.
\end{proof}

\begin{proof}\textbf{of Theorem~\ref{thm:fixed-t}.}
(i) Write the stake at time $t$ as $u_t = u_{\theta_t}$ with $\theta_t$ an
$\mathcal{F}_{t-1}$-measurable random element of $K_{H,G}$, and let $\varphi(\theta) =
\mathbb{E}[d^w_{u_\theta}(F)]$ for $F \sim P_F$, a continuous function with $\varphi \le L_{H,G}$ on
$K_{H,G}$. As $F_t$ is independent of $\mathcal{F}_{t-1}$, $\mu_t = \mathbb{E}[d_t \mid
\mathcal{F}_{t-1}] = \varphi(\theta_t) \le L_{H,G}$ almost surely. Given $\mathcal{F}_{t-1}$, $d_t -
\mu_t$ has mean zero and lies in the interval $[-1 - \mu_t, 1 - \mu_t]$ of length $2$, so
Hoeffding's lemma \citep{hoeffding1963probability} gives $\mathbb{E}[\exp\{s(d_t - \mu_t)\} \mid
\mathcal{F}_{t-1}] \le \exp(s^2/2)$ for $s \ge 0$. By the tower property, $\mathbb{E} \exp\{s\sum_{t
\le T}(d_t - \mu_t)\} \le \exp(Ts^2/2)$, and Markov's inequality with $s = a_T$ gives
\[
  \Pr\Bigl\{\sum_{t \le T} (d_t - \mu_t) \ge T a_T\Bigr\} \le \exp(-T a_T^2/2) = \frac{\alpha}{2} .
\]
Off this event $T^{-1} \sum_t d_t - a_T < T^{-1} \sum_t \mu_t \le L_{H,G}$, and as $L_{H,G} \ge 0$,
$\underline{L}_T \le L_{H,G}$. The rest of the chain holds surely: $L_{H,G} \le L_H$ by
Lemma~\ref{lem:grid}, and $L_H \le \Gamma_H(P) \le \Theta_w(P)$ by~\eqref{eq:smooth-weak-duality}.

(ii) The stakes of Proposition~\ref{prop:ogd} are predictable, as $\theta_{t+1}$ is computed from
$F_1, \dots, F_t$, and lie in $\mathcal{U}_{H,G}$, so (i) applies to them. Let $\theta^*$ attain
$L_{H,G}$ (Lemma~\ref{lem:grid}). The scores $e_t = d^w_{u_{\theta^*}}(F_t)$ are i.i.d.\ in $[-1,
1]$ with mean $L_{H,G}$, and the computation of (i) applied to $L_{H,G} - e_t$ gives $\Pr\{T^{-1}
\sum_t e_t \le L_{H,G} - a_T\} \le \alpha/2$. Proposition~\ref{prop:ogd} holds for every sequence of
bundles and gives $\sum_t d_t \ge \sum_t e_t - \sqrt{GT}$. Off both exceptional events, whose
probabilities add up to at most $\alpha$,
\[
  \frac1T \sum_{t \le T} d_t - a_T > L_{H,G} - \sqrt{G/T} - 2a_T \ge L_H - H\Delta/2 -
  \sqrt{G/T} - 2a_T
\]
by Lemma~\ref{lem:grid}, and $\underline{L}_T$, the larger of $0$ and the left side, is at least the
larger of $0$ and the right side.

\emph{The learning gap in expectation.} Let $\rho_T = T^{-1} \sum_{t \le T} \varphi(\theta_t)$ be
the average population value of the stakes used; Figure~\ref{fig:decomposition} plots $L_{H,G} -
\rho_T$. For the stakes of Proposition~\ref{prop:ogd} with any fixed $\eta$,
$\mathbb{E}[f_t(\theta_t) \mid \mathcal{F}_{t-1}] = \varphi(\theta_t)$ as in (i), and
$\mathbb{E}[f_t(\theta^*)] = L_{H,G}$ for $\theta^*$ attaining $L_{H,G}$, so taking expectations in
Proposition~\ref{prop:ogd} gives $\mathbb{E}[L_{H,G} - \rho_T] \le G/(2\eta T) + \eta/2$. This
bounds the expected gap, not the gap in each run.
\end{proof}

\begin{proof}\textbf{of Proposition~\ref{prop:anytime}.}
As $d_s - \tau \in [-2, 1]$ and $0 \le \lambda_s \le \tfrac12$, each factor satisfies $1 +
\lambda_s(d_s - \tau) \ge 1 - 2\lambda_s \ge 0$, so $M_t(\tau) \ge 0$, and $M_t(\tau) \le (3/2)^t$.
As in the proof of Theorem~\ref{thm:fixed-t}, $\mathbb{E}[d_t \mid \mathcal{F}_{t-1}] \le L_{H,G}$,
so under the null hypothesis $L_{H,G} \le \tau$, with $\lambda_t \ge 0$ predictable,
\[
  \mathbb{E}[M_t(\tau) \mid \mathcal{F}_{t-1}] = M_{t-1}(\tau) \bigl\{1 + \lambda_t
  (\mathbb{E}[d_t \mid \mathcal{F}_{t-1}] - \tau)\bigr\} \le M_{t-1}(\tau) ,
\]
and $(M_t(\tau))_{t \ge 0}$, with $M_0(\tau) = 1$, is a nonnegative supermartingale. The i.i.d.\
assumption enters only through $\mathbb{E}[d_t \mid \mathcal{F}_{t-1}] \le L_{H,G}$; if instead
$\mathbb{E}[d_t \mid \mathcal{F}_{t-1}] \le \tau$ at every $t$, the same display holds, which gives
the conditional null hypothesis stated after Proposition~\ref{prop:anytime}. For Ville's inequality,
let $N$ be the first $t$ with $M_t(\tau) \ge 1/\alpha$. For each $n$, optional stopping at the
bounded time $N \wedge n$ gives $\mathbb{E}\, M_{N \wedge n}(\tau) \le 1$, while $M_{N \wedge
n}(\tau) \ge \alpha^{-1} \mathbf{1}\{N \le n\}$; so $\Pr\{N \le n\} \le \alpha$, and letting $n \to
\infty$, $\Pr\{\exists t : M_t(\tau) \ge 1/\alpha\} \le \alpha$.

For the inversion, fix the outcomes. Each factor is nonnegative and nonincreasing in $\tau$, because
$d_s$ and $\lambda_s$ do not depend on $\tau$, so $M_s(\tau)$ and $\max_{s \le t} M_s(\tau)$ are
nonincreasing in $\tau$, and the set $R_t$ of $\tau \in [0, 1]$ with $\max_{s \le t} M_s(\tau) \ge
1/\alpha$ contains $[0, \tau]$ whenever it contains $\tau$. Let $\tau_0 = L_{H,G} \in [0, 1]$, at
which the null hypothesis holds. If $\underline{L}^{\mathrm{any}}_t > \tau_0$, then $R_t$ contains
some $\tau > \tau_0$, hence $\tau_0$, and $M_s(\tau_0) \ge 1/\alpha$ for some $s \le t$. Therefore
\[
  \Pr\bigl\{\exists t : \underline{L}^{\mathrm{any}}_t > L_{H,G}\bigr\} \le \Pr\bigl\{\exists s :
  M_s(\tau_0) \ge 1/\alpha\bigr\} \le \alpha .
\]
A fixed convex combination $\sum_k c_k M^{(k)}_t(\tau)$ of such processes, each with predictable
fractions in $[0, \tfrac12]$ and the same stakes, is nonnegative, starts at $1$, is a
supermartingale under the null hypothesis as a convex combination of supermartingales for the same
filtration, and is nonincreasing in $\tau$; the same argument applies to it.
\end{proof}

\clearpage
\section{Repair Without Labels}
\label{app:repair}

A positive certificate shows that a predictor is miscalibrated but not how to correct it, and the
labels that would correct it are missing. This appendix collects what is known about repairing
answers without labels and what it implies for the certificates. None of the inequalities below is
new. The dominance of an incoherent forecast by its Bregman projection is due to
\citet{predd2009probabilistic}, the decomposition of an expected score to
\citet{brocker2009reliability}, the argument against improving one forecast without labels to
\citet{kong2026calibration}, and the effect of a constant forecast on calibration measures to
\citet{hartline2025truthful}. We restate them in our notation, with short proofs for completeness,
because of two consequences for certificates: a repair of the whole bundle can raise the residuals
that certificates bound while it sets every population certificate to zero
(Counterexample~\ref{cx:repair}), and its guarantee, which concerns the score of the whole bundle,
does not extend to a single question (Proposition~\ref{prop:focal-repair}).

\emph{Scores and the calibration regret.} Let $\Phi$ be a strictly convex function that is
continuously differentiable on an open set containing $[0,1]^m$, and let
\[
  D_\Phi(a, b) = \Phi(a) - \Phi(b) - \langle \nabla\Phi(b), a - b \rangle ,
\]
which is nonnegative and zero only when $a = b$. In world $\ell$ a bundle $v$ receives the score
$D_\Phi(z^\ell, v)$, a proper scoring rule in Bregman form, oriented as a penalty (lower is better).
With $\Phi(c) = \sum_i w_i c_i^2$ this is the weighted Brier score $\lVert v - z^\ell \rVert_w^2 =
\sum_i w_i (v_i - z^\ell_i)^2$. The generator of the logarithmic score is not differentiable on the
boundary of the cube, and we do not treat it. For a random bundle $G = g(F)$, with $g$ Borel and
values in $[0,1]^m$, let $\xi_G = \mathbb{E}[Z \mid G]$; since $Z$ takes finitely many values in
$\mathcal{C}$, a version of $\xi_G$ takes values in $\mathcal{C}$. The \emph{calibration regret} of
$G$ is
\[
  \mathrm{Cal}_\Phi(G) = \mathbb{E}\, D_\Phi(\xi_G, G) ,
\]
which for the weighted Brier score is $\sum_i w_i \, \mathbb{E}\bigl(G_i - \Pr(Z_i = 1 \mid
G)\bigr)^2$: every component is conditioned on the whole bundle, as in the joint residual, and the
deviation is squared.

\begin{lemma}[Bregman projection; \citealp{predd2009probabilistic}]
\label{lem:bregman-projection}
For every $v \in [0,1]^m$, $c \mapsto D_\Phi(c, v)$ has a unique minimiser $\Pi_\Phi(v)$ over
$\mathcal{C}$. The map $\Pi_\Phi \colon [0,1]^m \to \mathcal{C}$ is continuous, $\Pi_\Phi(v) = v$ if
and only if $v \in \mathcal{C}$, and for every $c \in \mathcal{C}$
\begin{equation}
  D_\Phi(c, v) \ge D_\Phi\bigl(c, \Pi_\Phi(v)\bigr) + D_\Phi\bigl(\Pi_\Phi(v), v\bigr) .
  \label{eq:pythagoras}
\end{equation}
\end{lemma}

\begin{proof}
For fixed $v$, $D_\Phi(c, v) = \Phi(c) - \langle \nabla\Phi(v), c \rangle + \text{const}$ is
continuous and strictly convex in $c$ on the compact convex set $\mathcal{C}$, so it has a unique
minimiser, and $\Pi_\Phi(v) = v$ exactly when $v \in \mathcal{C}$, because $D_\Phi(c, v) = 0$ only
for $c = v$. For continuity, let $v_n \to v$. Since $\Phi$ and $\nabla\Phi$ are continuous, $D_\Phi$
is jointly continuous on $[0,1]^m \times [0,1]^m$. Every subsequence of $\Pi_\Phi(v_n)$ has a
further subsequence converging to some $c^* \in \mathcal{C}$, and letting $n \to \infty$ along it in
$D_\Phi(\Pi_\Phi(v_n), v_n) \le D_\Phi(c, v_n)$ gives $D_\Phi(c^*, v) \le D_\Phi(c, v)$ for all $c
\in \mathcal{C}$, so $c^* = \Pi_\Phi(v)$; hence $\Pi_\Phi(v_n) \to \Pi_\Phi(v)$.
For~\eqref{eq:pythagoras}, expanding the definitions gives, for all $a, b, c \in [0,1]^m$, the
three-point identity
\begin{equation}
  D_\Phi(c, b) = D_\Phi(c, a) + D_\Phi(a, b) + \langle \nabla\Phi(a) - \nabla\Phi(b), c - a
  \rangle .
  \label{eq:three-point}
\end{equation}
Take $b = v$ and $a = p = \Pi_\Phi(v)$. For $c \in \mathcal{C}$ and $t \in (0, 1]$, $p + t(c - p)
\in \mathcal{C}$, so $D_\Phi(p + t(c - p), v) \ge D_\Phi(p, v)$; dividing by $t$ and letting $t
\downarrow 0$ gives $\langle \nabla\Phi(p) - \nabla\Phi(v), c - p \rangle \ge 0$, since $\Phi$ is
differentiable at $p$. Inserting this in~\eqref{eq:three-point} gives~\eqref{eq:pythagoras}.
\end{proof}

\citet{predd2009probabilistic} state~\eqref{eq:pythagoras} for generators whose gradient need only
extend continuously to the boundary of the cube; the continuity of $\Pi_\Phi$, which we use only to
make $\Pi_\Phi(F)$ a random variable, is not stated there.

\begin{theorem}[Repairing the whole bundle]
\label{thm:repair}
Let $P$ be any law of $(F, Z)$ on $[0,1]^m \times \mathcal{Z}$, so any element of $\mathcal{P}(P_F)$
for any $P_F$, and let $\bar F = \Pi_\Phi(F)$. Then
\begin{enumerate}
  \item[(i)] $D_\Phi(z^\ell, F) \ge D_\Phi(z^\ell, \bar F) + D_\Phi(\bar F, F)$ for every world
  $\ell$ and every value of $F$;
  \item[(ii)] $\mathrm{Cal}_\Phi(\bar F) \le \mathrm{Cal}_\Phi(F) - \mathbb{E}\, D_\Phi(\bar F, F) -
  \mathbb{E}\, D_\Phi(\xi_F, \xi_{\bar F})$.
\end{enumerate}
For the weighted Brier score the guaranteed improvement $\mathbb{E}\lVert F - \bar F \rVert_w^2$
depends only on $P_F$ and $\mathcal{C}$, and it is positive if and only if $F \notin \mathcal{C}$
with positive probability.
\end{theorem}

\begin{proof}
(i) is~\eqref{eq:pythagoras} with $c = z^\ell \in \mathcal{C}$. (ii) Since $\Pi_\Phi$ is continuous,
$\sigma(\bar F) \subseteq \sigma(F)$. Since $\xi_F \in \mathcal{C}$, \eqref{eq:pythagoras} at $v =
F$ and $c = \xi_F$ gives $D_\Phi(\xi_F, F) \ge D_\Phi(\xi_F, \bar F) + D_\Phi(\bar F, F)$ pointwise.
All terms are bounded, as $D_\Phi$ is continuous on a compact set, and taking expectations gives
$\mathrm{Cal}_\Phi(F) \ge \mathbb{E}\, D_\Phi(\xi_F, \bar F) + \mathbb{E}\, D_\Phi(\bar F, F)$.
By~\eqref{eq:three-point} with $(c, a, b) = (\xi_F, \xi_{\bar F}, \bar F)$,
\[
  D_\Phi(\xi_F, \bar F) = D_\Phi(\xi_F, \xi_{\bar F}) + D_\Phi(\xi_{\bar F}, \bar F) + \langle
  \nabla\Phi(\xi_{\bar F}) - \nabla\Phi(\bar F), \xi_F - \xi_{\bar F} \rangle .
\]
The vector $\nabla\Phi(\xi_{\bar F}) - \nabla\Phi(\bar F)$ is bounded and $\sigma(\bar
F)$-measurable, and $\mathbb{E}[\xi_F \mid \bar F] = \mathbb{E}\bigl[\mathbb{E}[Z \mid F] \bigm|
\bar F\bigr] = \xi_{\bar F}$ by the tower property, so the inner product has expectation zero and
$\mathbb{E}\, D_\Phi(\xi_F, \bar F) = \mathbb{E}\, D_\Phi(\xi_F, \xi_{\bar F}) +
\mathrm{Cal}_\Phi(\bar F)$. Combining the two inequalities gives (ii). For the weighted Brier score,
$D_\Phi(\bar F, F) = \lVert F - \bar F \rVert_w^2$ is a function of $F$ that is positive exactly
when $F \notin \mathcal{C}$.
\end{proof}

(i) is the dominance theorem of \citet{predd2009probabilistic}, whose proof projects onto the
coherent set. In forecast reconciliation, projecting forecasts onto a coherent subspace, or onto a
closed convex set, likewise never increases the squared error, for every realisation
\citep{panagiotelis2021forecast,vanerven2015game}. For the answers of language models the Brier form
of (i) appears in \citet[Proposition~11]{dragonevskiy2026understanding}, and
\citet{kotawala2026locally} bounds the squared distance to the true marginals when those lie in the
coherent set. The first step of (ii) is the expected form of~\eqref{eq:pythagoras}, which
\citet[Theorem~8]{mohri2025coherence} state for a projection onto a closed convex set of conditional
distributions that is assumed to contain the optimal one, and its second step is the decomposition
of an expected score by sufficiency in \citet{brocker2009reliability}; equivalently, (ii) is (i)
averaged over the label and split into calibration regret and refinement. What (ii) adds is an
interpretation. The calibration regret falls by at least the score gain, and further by the
resolution $\mathbb{E}\, D_\Phi(\xi_F, \xi_{\bar F})$ that the projection gives up when it merges
bundles; and the optimal conditional law, which \citet{mohri2025coherence} assume to lie in the
constraint set, lies in $\mathcal{C}$ here without assumption, because $Z$ takes its values among
the worlds. It holds only as far as the semantics do: labels that violate the logical structure
entered in $\mathcal{C}$ void both (i) and (ii). Nor does (ii) alone single out the projection. For
the weighted Brier score, the constant $\Pi_\Phi(\mathbb{E}F)$ also has calibration regret $\lVert
\mathbb{E}Z - \Pi_\Phi(\mathbb{E}F) \rVert_w^2 \le \lVert \mathbb{E}Z - \mathbb{E}F \rVert_w^2 \le
\mathrm{Cal}_\Phi(F)$, by~\eqref{eq:pythagoras} and Jensen's inequality, but it gives up all
resolution; what singles out the projection is that by (i) its score is no worse in any world.

\begin{counterexample}[Repair can raise the residuals that certificates bound]
\label{cx:repair}
Take three labels, the bundle of their atoms ($m = 3$, $\mathcal{C}$ the simplex, weights
$\tfrac13$), the constant bundle $F = (0.6, 0.6, 0.1)$ and a label with law $q = (0.6, 0.2, 0.2)$.
The projection for the weighted Brier score is $\bar F = (0.5, 0.5, 0)$, and the calibration regret
falls from $0.17/3$ to $0.14/3$, by exactly $\lVert F - \bar F \rVert_w^2 = 0.03/3$. Since both
bundles are constant, the mean, marginal and joint residuals and $\Gamma_H$ all equal $\sum_i w_i
\lvert v_i - q_i \rvert$ for the constant value $v$, which rises from $0.5/3$ to $0.6/3$. Every
population certificate of $F$ equals $\operatorname{dist}^w_1(F, \mathcal{C}) = 0.1$, and every
population certificate of $\bar F$ is $0$, as for every coherent predictor
(Counterexample~\ref{cx:coherent}).
\end{counterexample}

This shows a limit of the projection's guarantee; it does not show that no repair without labels can
lower $\Theta_w$, and we do not study repairs in the $\ell_1$ residuals.

\begin{proposition}[No label-free repair of one question's score]
\label{prop:focal-repair}
Let component $j$ ask an event that is neither sure nor impossible, and let $g \colon [0,1]^m \to
[0,1]$ be Borel, chosen with knowledge of $P_F$ and $\mathcal{C}$.
\begin{enumerate}
  \item[(i)] If $P_F(g(F) \ne F_j) > 0$, there is $P \in \mathcal{P}(P_F)$ under which
  $\mathbb{E}\bigl(g(F) - Z_j\bigr)^2 - \mathbb{E}(F_j - Z_j)^2 = \mathbb{E}\bigl(g(F) - F_j\bigr)^2
  > 0$.
  \item[(ii)] For $\mathrm{ECE}_1$ this fails: the constant $g = \mathbb{E} F_j$ has
  $\mathrm{ECE}_1(g) \le \mathrm{ECE}_1(F_j)$ under every $P \in \mathcal{P}(P_F)$, and if $F_j$
  takes the values $\tfrac14$ and $\tfrac34$ with probability $\tfrac12$ each, some $P \in
  \mathcal{P}(P_F)$ has $\mathrm{ECE}_1(F_j) = \tfrac34$ and $\mathrm{ECE}_1(g) = 0$.
\end{enumerate}
\end{proposition}

\begin{proof}
(i) Choose worlds $\ell_1$ with $z^{\ell_1}_j = 1$ and $\ell_0$ with $z^{\ell_0}_j = 0$, and let $P$
draw, given $F$, $Z = z^{\ell_1}$ with probability $F_j$ and $Z = z^{\ell_0}$ otherwise; the kernel
is Borel in $F$, so $P \in \mathcal{P}(P_F)$, and $\Pr(Z_j = 1 \mid F) = F_j$. Given $F$, with $g =
g(F)$,
\[
  \mathbb{E}\bigl[(g - Z_j)^2 - (F_j - Z_j)^2 \bigm| F\bigr] = g^2 - F_j^2 - 2 (g - F_j) F_j = (g -
  F_j)^2 ,
\]
and taking expectations gives the identity; the right side is positive because $g(F) \ne F_j$ with
positive probability. (ii) A constant answer is conditioned on nothing, so $\mathrm{ECE}_1(g) =
\lvert \mathbb{E} F_j - \Pr(Z_j = 1) \rvert = \bigl\lvert \mathbb{E}\, \mathbb{E}[F_j - Z_j \mid
F_j] \bigr\rvert \le \mathrm{ECE}_1(F_j)$ by Jensen's inequality. In the example, let $Z =
z^{\ell_1}$ exactly when $F_j = \tfrac14$: then $\mathrm{ECE}_1(F_j) = \tfrac12 \cdot \tfrac34 +
\tfrac12 \cdot \tfrac34$, and $\mathbb{E} F_j = \tfrac12 = \Pr(Z_j = 1)$.
\end{proof}

The argument of (i) is that of \citet[proof of Theorem~3.3]{kong2026calibration}: a forecast that is
calibrated under some law allowed by what is known cannot be improved under every such law. Here no
assumption on the other answers is needed to allow that law, which is the one behind
Proposition~\ref{prop:attribution}: the focal question calibrated given the whole bundle, the other
answers carrying the error. \citet{kong2026calibration} obtain an improvement without labels by
assuming that a reference forecaster is calibrated; when the two forecasters are not mutually
calibrated, this assumption excludes every law under which the improved one is calibrated. The proof
uses only that the Brier score is strictly proper, so for every strictly proper scoring rule of one
event with finite expected scores the same law makes $g(F)$ strictly worse; the excess is then the
expected divergence of that score between $g(F)$ and $F_j$, which is $\mathbb{E}(g(F) - F_j)^2$ only
for the Brier score. Under the same law, the projection of Theorem~\ref{thm:repair} still lowers the
calibration regret of the bundle while, by (i), it worsens the focal question unless $\bar F_j =
F_j$ almost surely: the gain is on the other answers. (ii) is the case of one event and
$\mathrm{ECE}_1$ of \citet[Theorem~A.4]{hartline2025truthful}, who show that replacing a forecast by
its mean raises no $\ell_\alpha$ expected calibration error ($\alpha \ge 1$) and no smooth
calibration error; see also \citet[Proposition~3.3]{kumar2019verified} for one bin. Under the law in
(i), the expected Brier score of the mean is worse by $\operatorname{Var} F_j$, which is
$\tfrac1{16}$ in the example. So, without labels, no post-processing is guaranteed to keep the score
of a question that is neither sure nor impossible from worsening, while its calibration error can be
lowered by giving up resolution. For the whole bundle, Theorem~\ref{thm:repair} guarantees that the
score is no worse in any world and that the calibration regret of that score falls; it guarantees
nothing about $\Theta_w$ (Counterexample~\ref{cx:repair}).

\clearpage
\section{Anytime-Valid Certificates on Synthetic Data}
\label{app:synthetic}

\begin{figure}[t]
  \centering
  \includegraphics[width=\linewidth]{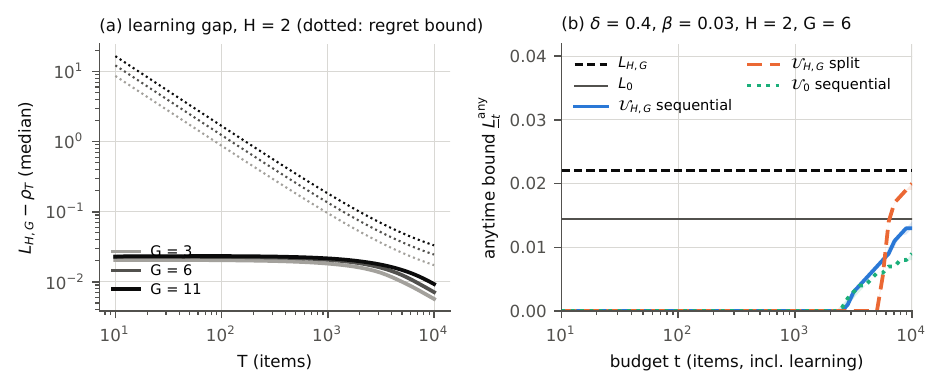}
  \caption{Power design on a synthetic bundle shaped like the main one ($K = 4$: a choice query,
    atoms and complements, with equal weights for the three roles). The choice query is calibrated;
    the yes/no answers are overconfident by $\delta$ and yes-biased by $\beta$. (a) The empirical
    learning gap $L_{H,G} - \rho_T$ of projected OGD (median over 5 seeds), $H = 2$, against
    $G/(2\eta T) + \eta/2$ (dotted), the bound that Proposition~\ref{prop:ogd} gives on its
    expectation, not on each seed, for the fixed step $\eta = \sqrt{G/T_{\max}}$
    (Appendix~\ref{app:proofs-computation}); it meets $\sqrt{G/T}$ only at $T = T_{\max} = 10^4$.
    (b) Anytime-valid lower bounds (Proposition~\ref{prop:anytime}, $\alpha = 0.05$, the mixture of
    $\lambda_k = 2^{-k}$, $k = 1, \dots, 8$) of the three ways of learning the stakes on the same
    streams: constants learnt online ($\mathcal U_0$), $\mathcal U_{H,G}$ learnt on the first half
    of the budget and held fixed, and $\mathcal U_{H,G}$ learnt online; medians over 10 seeds with
    10--90\% bands. The budget counts the learning items too. With only 64 support points the split
    method's batch fit is nearly exact.}
  \label{fig:power}
\end{figure}

\begin{figure}[t]
  \centering
  \includegraphics[width=\linewidth]{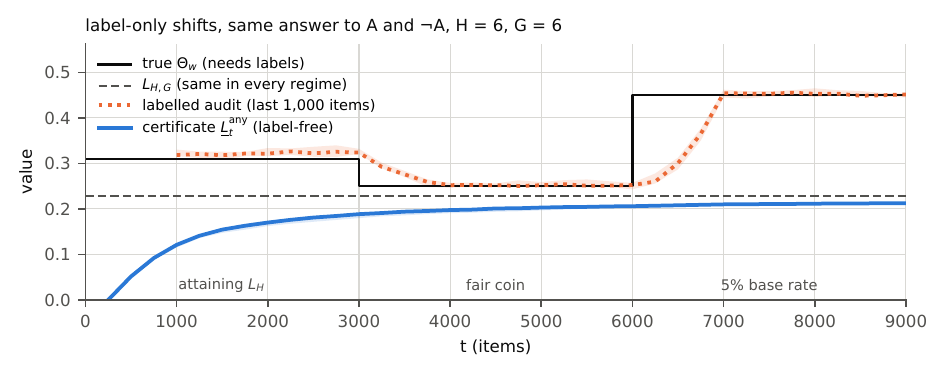}
  \caption{A blind spot of every label-free audit. One stream of bundles from the predictor of
    Figure~\ref{fig:decomposition}(a), which gives ``$A$?'' and ``not $A$?'' the same answer; the
    law of $Y$ given $F$ changes twice while $P_F$ stays put. The certificate's trajectory is the
    same in every regime, while the true $\Theta_w$ falls and then rises; a labelled audit (plug-in
    $\Theta_w$ over the last 1{,}000 items) follows it. Medians over 10 seeds with 10--90\% bands.}
  \label{fig:blind-spot}
\end{figure}

This appendix collects the remaining synthetic results. Figure~\ref{fig:power} and
Table~\ref{tab:power} give the power design on a synthetic bundle shaped like the main one, from
which the pilot's thresholds for going on to confirmatory data collection were taken
(Appendix~\ref{app:study}). Table~\ref{tab:coverage} checks by simulation that the fixed-$T$ and
anytime bounds hold at their levels and that the negative controls fail. Figure~\ref{fig:blind-spot}
and Table~\ref{tab:blind-spot} show two limits that Section~\ref{sec:limitations} discusses: a
change of the law of the labels alone, which no label-free audit sees, and a logical structure
entered wrongly, which makes a valid certificate refute the wrong thing.

\begin{table}[t]
  \centering
  \small
  \begin{tabular}{rrrr}
\toprule
$n$ & one-sided $\sqrt{2\log(1/\alpha)/n}$ & Theorem~\ref{thm:fixed-t} $a_T$ & planned \\
\midrule
200 & 0.1731 & 0.1921 & -- \\
500 & 0.1095 & 0.1215 & 0.110 \\
1,000 & 0.0774 & 0.0859 & 0.077 \\
2,000 & 0.0547 & 0.0607 & -- \\
5,000 & 0.0346 & 0.0384 & 0.035 \\
10,000 & 0.0245 & 0.0272 & 0.025 \\
15,000 & 0.0200 & 0.0222 & 0.020 \\
\bottomrule
\end{tabular}

\medskip

\begin{tabular}{rrrrrrrr}
\toprule
$H$ & $G$ & $L_H$ & $L_{H,G}$ & grid gap & $H\Delta/2$ & learning gap & $\sqrt{G/T}$ \\
\midrule
1 & 3 & 0.0185 & 0.0172 & 0.0013 & 0.250 & 0.0046 & 0.017 \\
1 & 6 & 0.0185 & 0.0183 & 0.0003 & 0.100 & 0.0062 & 0.024 \\
1 & 11 & 0.0185 & 0.0184 & 0.0001 & 0.050 & 0.0080 & 0.033 \\
2 & 3 & 0.0226 & 0.0200 & 0.0026 & 0.500 & 0.0057 & 0.017 \\
2 & 6 & 0.0226 & 0.0221 & 0.0005 & 0.200 & 0.0071 & 0.024 \\
2 & 11 & 0.0226 & 0.0224 & 0.0002 & 0.100 & 0.0093 & 0.033 \\
4 & 3 & 0.0280 & 0.0219 & 0.0060 & 1.000 & 0.0072 & 0.017 \\
4 & 6 & 0.0280 & 0.0262 & 0.0017 & 0.400 & 0.0093 & 0.024 \\
4 & 11 & 0.0280 & 0.0277 & 0.0003 & 0.200 & 0.0123 & 0.033 \\
\bottomrule
\end{tabular}

\medskip

\begin{tabular}{rrrrlrrr}
\toprule
$\delta$ & $\beta$ & $L_0$ & $L_{H,G}$ & method & $N_{\mathrm{any}}$ & $N_{\mathrm{any}}^{90\%}$ & $N_{\mathrm{fixed}}$ \\
\midrule
0 & 0.02 & 0.0126 & 0.0126 & $\mathcal{U}_{H,G}$ seq. & 5\,623 & 5\,623 & $>10^4$ \\
0 & 0.02 & 0.0126 & 0.0126 & $\mathcal{U}_{H,G}$ split & 6\,310 & 6\,310 & $>10^4$ \\
0 & 0.02 & 0.0126 & 0.0126 & $\mathcal{U}_0$ seq. & 3\,981 & 3\,981 & $>10^4$ \\
0 & 0.05 & 0.0304 & 0.0304 & $\mathcal{U}_{H,G}$ seq. & 1\,995 & 1\,995 & $>10^4$ \\
0 & 0.05 & 0.0304 & 0.0304 & $\mathcal{U}_{H,G}$ split & 5\,623 & 5\,623 & $>10^4$ \\
0 & 0.05 & 0.0304 & 0.0304 & $\mathcal{U}_0$ seq. & 1\,413 & 1\,413 & $>10^4$ \\
0 & 0.1 & 0.0583 & 0.0583 & $\mathcal{U}_{H,G}$ seq. & 891 & 891 & 4\,467 \\
0 & 0.1 & 0.0583 & 0.0583 & $\mathcal{U}_{H,G}$ split & 5\,623 & 5\,623 & 7\,943 \\
0 & 0.1 & 0.0583 & 0.0583 & $\mathcal{U}_0$ seq. & 708 & 708 & 3\,548 \\
0.4 & 0 & 0.0119 & 0.0161 & $\mathcal{U}_{H,G}$ seq. & 3\,548 & 3\,548 & $>10^4$ \\
0.4 & 0 & 0.0119 & 0.0161 & $\mathcal{U}_{H,G}$ split & 6\,310 & 6\,310 & $>10^4$ \\
0.4 & 0 & 0.0119 & 0.0161 & $\mathcal{U}_0$ seq. & 4\,467 & 4\,467 & $>10^4$ \\
0.8 & 0 & 0.0239 & 0.0288 & $\mathcal{U}_{H,G}$ seq. & 1\,778 & 1\,778 & $>10^4$ \\
0.8 & 0 & 0.0239 & 0.0288 & $\mathcal{U}_{H,G}$ split & 5\,623 & 5\,623 & $>10^4$ \\
0.8 & 0 & 0.0239 & 0.0288 & $\mathcal{U}_0$ seq. & 1\,995 & 1\,995 & $>10^4$ \\
0.4 & 0.03 & 0.0145 & 0.0221 & $\mathcal{U}_{H,G}$ seq. & 2\,818 & 2\,818 & $>10^4$ \\
0.4 & 0.03 & 0.0145 & 0.0221 & $\mathcal{U}_{H,G}$ split & 5\,623 & 5\,623 & $>10^4$ \\
0.4 & 0.03 & 0.0145 & 0.0221 & $\mathcal{U}_0$ seq. & 2\,512 & 2\,818 & $>10^4$ \\
\bottomrule
\end{tabular}

  \normalsize
  \caption{Power design. Top: Hoeffding widths at $\alpha = 0.05$; ``planned'' is what the study
    plan assumed. Middle: grid and learning gaps at $T = 10^4$ (medians over 5 seeds; $\delta =
    0.4$, $\beta = 0.03$) against $H\Delta/2$, the worst case of the grid gap
    (Lemma~\ref{lem:grid}), and $\sqrt{G/T}$, the bound on the expected learning gap
    (Proposition~\ref{prop:ogd}, Appendix~\ref{app:proofs-computation}). Bottom: the budget at which
    the bound first turns positive in the median seed ($N_{\mathrm{any}}$, $N_{\mathrm{fixed}}$;
    about half the seeds detect) and in at least 90\% of the seeds ($N_{\mathrm{any}}^{90\%}$), read
    off runs with $T_{\max} = 10^4$ ($H = 2$, $G = 6$, 10 seeds).}
  \label{tab:power}
\end{table}

\begin{table}[t]
  \centering
  \scriptsize
  \begin{tabular}{lllrrlrrr}
\toprule
condition & procedure & role & $H$ & $G$ & mode & target & $\hat p$ & pass line \\
\midrule
Counterexample~\ref{cx:pointwise} & $\mathcal{U}_{H,G}$ seq. & valid & 6 & 6 & fixed-T & 0.000 & 0.000 & 0.040 \\
Counterexample~\ref{cx:pointwise} & $\mathcal{U}_{H,G}$ seq. & valid & 6 & 6 & anytime & 0.000 & 0.000 & 0.071 \\
Counterexample~\ref{cx:pointwise} & $\mathcal{U}_0$ seq. & valid & 6 & 6 & fixed-T & 0.000 & 0.000 & 0.040 \\
Counterexample~\ref{cx:pointwise} & $\mathcal{U}_0$ seq. & valid & 6 & 6 & anytime & 0.000 & 0.000 & 0.071 \\
Counterexample~\ref{cx:pointwise} & $\mathcal{U}_{H,G}$ split & valid & 6 & 6 & fixed-T & 0.000 & 0.000 & 0.040 \\
Counterexample~\ref{cx:pointwise} & $\mathcal{U}_{H,G}$ split & valid & 6 & 6 & anytime & 0.000 & 0.004 & 0.071 \\
Counterexample~\ref{cx:pointwise} & fixed stake & valid & 6 & 6 & fixed-T & 0.000 & 0.000 & 0.040 \\
Counterexample~\ref{cx:pointwise} & fixed stake & valid & 6 & 6 & anytime & 0.000 & 0.005 & 0.071 \\
Counterexample~\ref{cx:pointwise} & fixed stake, clipped & control & 6 & 6 & fixed-T & 0.000 & 1.000 & 0.040 \\
Counterexample~\ref{cx:pointwise} & fixed stake, clipped & control & 6 & 6 & anytime & 0.000 & 1.000 & 0.071 \\
Counterexample~\ref{cx:pointwise} & look-ahead & control & 6 & 6 & fixed-T & 0.000 & 1.000 & 0.040 \\
Counterexample~\ref{cx:pointwise} & look-ahead & control & 6 & 6 & anytime & 0.000 & 1.000 & 0.071 \\
Counterexample~\ref{cx:pointwise} & empirical LP & control & 6 & 6 & empirical & 0.000 & 0.979 & 0.071 \\
same answer & fixed optimum & valid & 6 & 6 & fixed-T & 0.229 & 0.000 & 0.040 \\
same answer & fixed optimum & valid & 6 & 6 & anytime & 0.229 & 0.001 & 0.071 \\
same answer & $\mathcal{U}_{H,G}$ seq. & valid & 6 & 6 & fixed-T & 0.229 & 0.000 & 0.040 \\
same answer & $\mathcal{U}_{H,G}$ seq. & valid & 6 & 6 & anytime & 0.229 & 0.000 & 0.071 \\
same answer & $\mathcal{U}_{H,G}$ split & valid & 6 & 6 & fixed-T & 0.229 & 0.000 & 0.040 \\
same answer & $\mathcal{U}_{H,G}$ split & valid & 6 & 6 & anytime & 0.229 & 0.000 & 0.071 \\
same answer & $\mathcal{U}_0$ seq. & valid & 6 & 6 & fixed-T & 0.000 & 0.000 & 0.040 \\
same answer & $\mathcal{U}_0$ seq. & valid & 6 & 6 & anytime & 0.000 & 0.000 & 0.071 \\
yes to both & $\mathcal{U}_{H,G}$ seq. & valid & 1 & 11 & fixed-T & 0.400 & 0.000 & 0.040 \\
yes to both & $\mathcal{U}_{H,G}$ seq. & valid & 1 & 11 & anytime & 0.400 & 0.000 & 0.071 \\
yes to both & $\mathcal{U}_{H,G}$ seq. & valid & 1 & 11 & fixed-T chain & 0.400 & 0.000 & 0.071 \\
\bottomrule
\end{tabular}

  \normalsize
  \caption{Coverage by simulation ($R = 1{,}000$ replicates of $T = 1{,}000$ items, $\alpha =
    0.05$). $\hat p$ is the share of replicates in which the checked event fails: the fixed-$T$
    bound above its target (level $\alpha/2$), Theorem~\ref{thm:fixed-t}'s whole chain failing
    (level $\alpha$), or the anytime bound above its target at some $t \le T$ (level $\alpha$). A
    valid procedure passes when $\hat p$ is at most the pass line; a negative control (clipping each
    item at zero, choosing the stake after seeing the item, the uncorrected empirical LP) must
    exceed it. ``same answer'' is the predictor of Figure~\ref{fig:decomposition}(a) under its
    attaining label law, on which the population-optimal stake (``fixed optimum'') has expected
    profit exactly $L_{H,G}$; ``yes to both'' answers close to yes to both $A$ and $\neg A$, so that
    at $H = 1$ and $G = 11$ the whole chain of Theorem~\ref{thm:fixed-t} can be checked (``fixed-T
    chain''); ``fixed stake'' is one stake fixed in advance whose profit has mean exactly $0$.}
  \label{tab:coverage}
\end{table}

\begin{table}[t]
  \centering
  \small
  \begin{tabular}{lrrrr}
\toprule
$Y\mid F$ & $L_H$ & $L_{H,G}$ & true $\Gamma_H$ & true $\Theta_w$ \\
\midrule
attaining $L_H$ & 0.240 & 0.229 & 0.240 & 0.310 \\
fair coin & 0.240 & 0.229 & 0.240 & 0.250 \\
5\% base rate & 0.240 & 0.229 & 0.450 & 0.450 \\
\bottomrule
\end{tabular}

\medskip

\begin{tabular}{lrrrrrr}
\toprule
structure & worlds & $L_0$ & $L_{\mathrm{marg}}$ & $L_{\mathrm{joint}}$ & $L_H$ & true $\Theta_w$ \\
\midrule
true & 3 & 0.000 & 0.000 & 0.000 & 0.000 & 0.000 \\
claimed & 2 & 0.200 & 0.200 & 0.200 & 0.200 & 0.000 \\
widened & 4 & 0.000 & 0.000 & 0.000 & 0.000 & 0.000 \\
\bottomrule
\end{tabular}

  \normalsize
  \caption{Top: the regimes of Figure~\ref{fig:blind-spot}. Bottom: a calibrated, coherent predictor
    of ``$A$?'' and ``$B$?'' on the labels $\{A, B, C\}$, audited under the true structure, under a
    structure that forgets $C$ (``claimed''), and with the doubtful relation dropped (``widened'').
    A wrong relation yields a positive certificate although $\Theta_w = 0$; widening only lowers the
    bound.}
  \label{tab:blind-spot}
\end{table}

\clearpage
\section{Labelled Descriptive Metrics on Real Data}
\label{app:descriptive}

Table~\ref{tab:descriptive} lists, for each confirmatory stream, the labelled estimates of
$\Theta_w$ and the plug-in values discussed in Section~\ref{sec:exp-real}, two of which the protocol
names as negative controls. The estimation conditions were not preregistered;
Appendix~\ref{app:preregistration} records when they were fixed. The three sensitivity estimates
differ from the main one by less than $0.002$ on every pair.

\begin{table}[t]
  \centering
  \small
  \setlength{\tabcolsep}{2.5pt}
  \begin{tabular}{lrrrrrrrrr}
\toprule
 & & \multicolumn{4}{c}{$\widehat\Theta_w$ (binned ECE)} & & \multicolumn{2}{c}{negative controls} & reference \\
\cmidrule(lr){3-6}\cmidrule(lr){8-9}\cmidrule(lr){10-10}
System & certificate & 15 eq.-w. & 10 eq.-w. & 20 eq.-w. & 15 eq.-m. & Brier & $\hat L_0$ & $\hat L_{\mathrm{marg}}$ & $\hat L_{\mathrm{joint}}$ \\
\midrule
\multicolumn{10}{l}{AG News} \\
Jev & 0.029 & \textbf{0.059} & 0.059 & 0.060 & 0.059 & 0.061 & 0.038 & 0.040 & 0.053 \\
Eikos & 0.095 & \textbf{0.128} & 0.127 & 0.128 & 0.129 & 0.096 & 0.106 & 0.106 & 0.113 \\
Qwen3.5 & 0.185 & \textbf{0.217} & 0.216 & 0.217 & 0.217 & 0.190 & 0.194 & 0.201 & 0.232 \\
DeBERTa & 0.070 & \textbf{0.121} & 0.121 & 0.122 & 0.121 & 0.126 & 0.069 & 0.103 & 0.103 \\
\midrule
\multicolumn{10}{l}{DBpedia-14} \\
Eikos & 0.054 & \textbf{0.095} & 0.094 & 0.095 & 0.094 & 0.030 & 0.081 & 0.082 & 0.084 \\
Qwen3.5 & 0.194 & \textbf{0.215} & 0.215 & 0.216 & 0.215 & 0.161 & 0.207 & 0.211 & 0.225 \\
DeBERTa & 0.032 & \textbf{0.050} & 0.050 & 0.051 & 0.050 & 0.031 & 0.038 & 0.052 & 0.052 \\
\bottomrule
\end{tabular}

  \normalsize
  \caption{Labelled descriptive metrics and negative controls on the confirmatory streams of
    Figure~\ref{fig:real-data}; none of them is a certificate. ``certificate'' is the confirmatory
    anytime lower bound on $L_{H,G}$ (Table~\ref{tab:real-data}). $\widehat\Theta_w$ estimates
    $\Theta_w = \sum_i w_i\,\mathrm{ECE}_1(F_i)$ with the items' labels: per coordinate, the binned
    ECE (15 equal-width bins, in bold; 10 and 20 equal-width and 15 equal-mass bins for
    sensitivity), aggregated with the bundle's fixed weights. Brier is $\sum_i w_i$ times the mean
    squared error of coordinate $i$. A binned ECE estimates $\Theta_w$; it is not $\Theta_w$. The
    estimation conditions were not preregistered. The negative controls are $L_0$ and
    $L_{\mathrm{marg}}$ computed on the empirical distribution of the stream without correction:
    they overstate, are never valid bounds and are not compared with the certificate. $\hat
    L_{\mathrm{joint}}$, the empirical mean of $\mathrm{dist}^w_1(F, \mathcal{C})$, is shown for
    reference; it concerns the joint calibration error, not $\Theta_w$.}
  \label{tab:descriptive}
\end{table}

\clearpage
\section{Stake Class Against Learning Method on Real Data (Post Hoc)}
\label{app:u0-split}

Figure~\ref{fig:real-data} learns the constant stakes of $\mathcal{U}_0$ sequentially only, so its
comparison cannot separate the stake class from the way the stakes are learnt.
Table~\ref{tab:u0-split} adds the missing arm, $\mathcal{U}_0$ learnt by the split method: the split
method of $\mathcal{U}_{H,G}$ with $H = 0$, whose admissible stakes are the constants in $[-1, 1]$.
This arm was added after Figure~\ref{fig:real-data} had been seen. It was not preregistered and is
not part of the confirmatory family; its conditions were fixed before any of its values was
computed.

With the learning method held fixed, the split arm of $\mathcal{U}_{H,G}$ was above that of
$\mathcal{U}_0$ at $T_{\max}$ on three of the seven pairs (by $0.008$ and $0.002$ for DeBERTa on AG
News and DBpedia-14, and by one grid step of $10^{-3}$ for Qwen3.5 on AG News), equal on the other
four, and below on none; the $\mathcal{U}_0$ split arm first turned positive at the same budget on
four pairs and 8 to 26 items later on the other three. With the stake class held fixed, the split
arm was above the sequential one on all seven pairs, for $\mathcal{U}_0$ (by $0.003$ to $0.005$) as
for $\mathcal{U}_{H,G}$ (by $0.004$ to $0.012$). On the four pairs where the split arms tie, the
split arm's advantage in Figure~\ref{fig:real-data} goes with the learning method alone; on the two
DeBERTa pairs the predictor-dependent stakes also improved on the constant stakes, on AG News by as
much as the learning method did. We did not establish why. Each comparison is a description of one
stream per pair, not a test.

\begin{table}[t]
  \centering
  \small
  \setlength{\tabcolsep}{4pt}
  \begin{tabular}{lrrrrrrrrr}
\toprule
 & \multicolumn{5}{c}{anytime at $T_{\max}$} & \multicolumn{2}{c}{fixed-$T$} & \multicolumn{2}{c}{first positive $t$} \\
\cmidrule(lr){2-6}\cmidrule(lr){7-8}\cmidrule(lr){9-10}
 & \multicolumn{2}{c}{sequential} & \multicolumn{3}{c}{split} & \multicolumn{2}{c}{split} & \multicolumn{2}{c}{split} \\
\cmidrule(lr){2-3}\cmidrule(lr){4-6}\cmidrule(lr){7-8}\cmidrule(lr){9-10}
System & $\mathcal{U}_{H,G}$ & $\mathcal{U}_0$ & $\mathcal{U}_{H,G}$ & $\mathcal{U}_0$ & difference & $\mathcal{U}_{H,G}$ & $\mathcal{U}_0$ & $\mathcal{U}_{H,G}$ & $\mathcal{U}_0$ \\
\midrule
\multicolumn{10}{l}{AG News} \\
Jev & \textbf{0.029} & 0.031 & 0.034 & 0.034 & 0 & 0 & 0 & 5{,}381 & 5{,}389 \\
Eikos & \textbf{0.095} & 0.098 & 0.102 & 0.102 & 0 & 0.0587 & 0.0587 & 5{,}150 & 5{,}150 \\
Qwen3.5 & \textbf{0.185} & 0.186 & 0.191 & 0.190 & +0.001 & 0.1473 & 0.1463 & 5{,}076 & 5{,}076 \\
DeBERTa & \textbf{0.070} & 0.061 & 0.074 & 0.066 & +0.008 & 0.0308 & 0.0217 & 5{,}194 & 5{,}220 \\
\midrule
\multicolumn{10}{l}{DBpedia-14} \\
Eikos & \textbf{0.054} & 0.062 & 0.066 & 0.066 & 0 & 0 & 0 & 1{,}178 & 1{,}178 \\
Qwen3.5 & \textbf{0.194} & 0.197 & 0.202 & 0.202 & 0 & 0.1415 & 0.1411 & 2{,}573 & 2{,}573 \\
DeBERTa & \textbf{0.032} & 0.032 & 0.037 & 0.035 & +0.002 & 0 & 0 & 5{,}354 & 5{,}370 \\
\bottomrule
\end{tabular}

  \normalsize
  \caption{Stake class against learning method on the confirmatory streams of
    Figure~\ref{fig:real-data}, every bound at $\alpha = 0.05/7$. The two split arms learn on the
    first $\lfloor T_{\max}/2 \rfloor$ items and score the rest with the learnt stake; they differ
    only in $H$ (the frozen value for $\mathcal{U}_{H,G}$, $0$ for $\mathcal{U}_0$), with the same
    bundle, weights, nodes, betting rule and arrival order. ``difference'' is the
    $\mathcal{U}_{H,G}$ split bound minus the $\mathcal{U}_0$ split bound; ``first positive $t$'' is
    the budget at which a split arm's anytime bound first turned positive. The sequential columns
    and the $\mathcal{U}_{H,G}$ split arm repeat Table~\ref{tab:real-data}; recomputed by the same
    code as the $\mathcal{U}_0$ split arm, the $\mathcal{U}_{H,G}$ split arm was reproduced exactly.
    The $\mathcal{U}_0$ split arm is post hoc and not preregistered.}
  \label{tab:u0-split}
\end{table}

\clearpage
\section{Controls on the Pilot Items}
\label{app:controls}

The pilot also included preregistered controls, each of which changes one factor of a system's
\emph{main condition}, its configuration in Section~\ref{sec:exp-real}, on the same 200 exploration
items of each data set: the options of the choice query named by their labels instead of neutral
letters; the options listed in the reverse order, with the neutral letters given by position, so
that every label also changes its letter; and Jev asked one question per request instead of the
whole bundle in one. Qwen3.5 reads option letters only and takes no named control, and DeBERTa asks
the same hypotheses whatever the options' names or order and takes no option control. A control
changes the wording of the questions or a setting of the predictor, so it is a different audited
predictor, but it keeps the components and weights of the main bundle, so its residual is the same
functional $\Theta_w$; the controls entered neither the selection of $H$ and $G$ nor the decision
taken after the pilot (Appendix~\ref{app:study}). Collecting them was preregistered; the quantities
compared were not. They were fixed before any control's value or any comparison was computed; the
main conditions' bounds and plug-in values and the noise floors (Appendix~\ref{app:pilot}) are the
pilot's. Table~\ref{tab:controls} lists all twelve comparisons. For each condition it gives the
anytime lower bound on $L_{H,G}$ computed as in the pilot (Proposition~\ref{prop:anytime} at $\alpha
= 0.05$, $\mathcal{U}_{H,G}$ learnt sequentially with $\eta = \sqrt{G/200}$, at the selected $H$ and
$G$), the uncorrected plug-in value of $L_{H,G}$ on the 200 items, an effect size and never a bound,
and the accuracy of the choice query, computed with the items' labels. For each pair it gives the
item-matched shift $\frac{1}{T} \sum_{t \le T} \sum_i w_i \lvert F^{\mathrm{c}}_{t,i} -
F^{\mathrm{main}}_{t,i} \rvert$, over all components and over the choice query's alone. The shift is
the cost of one coupling of the two empirical laws, so it is at least their $W_{1,w}$, and by
Theorem~\ref{thm:stable} and Lemma~\ref{lem:grid} the plug-in values of $L_H$ and $L_{H,G}$ under
the two conditions differ by at most $(1 + H)$ times it. The bound says nothing about the anytime
bounds, which are finite-sample values.

The plug-in values changed by at most $0.0033$, below $(1 + H)$ times the shift on every pair and
small next to the values themselves ($0.013$ to $0.212$). This is a statement about the empirical
$L_{H,G}$, not about the answers, their accuracy or their calibration error: the inequality bounds
the change of $L_{H,G}$ by the shift of the answers, not the other way round. The answers did move,
and for Eikos with named options on AG News the plug-in value was exactly that of the main condition
while the answers and the accuracy of its choice query changed. We had fixed in advance a reading
under which a small change would mean answers robust to the factor; for this reason we do not adopt
it. The anytime bounds were $0$ under both conditions on eight pairs, so those pairs are
uninformative, and changed by at most $0.002$ on the other four. Eikos and Qwen3.5 answer each
question in its own forward pass, so under an option control their yes/no questions are asked
exactly as under the main condition, and only their choice query moved. Jev answers the whole bundle
in one request, and its answers to a control come from new calls; its repeated calls under the main
condition do not return identical answers, and $(1 + H)$ times its shift was $0.95$ to $1.77$ times
its noise floor on AG News and $0.94$ to $0.96$ times on DBpedia-14, so what we report does not
separate its movement under the controls from that variation. The accuracy of the choice query, a
share of items rather than a bundle value, moved by up to $0.063$ (Qwen3.5 under the reversed order
on AG News, from $0.858$ to $0.795$); Table~\ref{tab:controls} lists the rest.
\citet{sun2026typesafe} report that typed decision models follow an option's name when names are
reassigned to other definitions; our named control keeps each name with its own label, so it does
not test that. The certificates of Section~\ref{sec:exp-real} concern the configurations as frozen,
and we do not carry them over to the controls. Each comparison rests on one stream of 200 items and
is descriptive; we did not establish why any of them came out as it did.

\begin{table}[t]
  \centering
  \scriptsize
  \setlength{\tabcolsep}{2pt}
  \begin{tabular}{@{}l>{\raggedright\arraybackslash}p{2.1cm}rrrrrrrrrr@{}}
\toprule
 & & \multicolumn{2}{c}{anytime bound} & \multicolumn{2}{c}{plug-in $L_{H,G}$} & \multicolumn{2}{c}{Choice accuracy} & \multicolumn{2}{c}{shift} & & \\
\cmidrule(lr){3-4}\cmidrule(lr){5-6}\cmidrule(lr){7-8}\cmidrule(lr){9-10}
System & Control & main & control & main & control & main & control & all & Choice & $(1 + H)\,\cdot$ all & noise floor \\
\midrule
\multicolumn{12}{@{}l}{AG News ($H = 2$, $G = 6$)} \\
Jev & option names & 0.000 & 0.000 & 0.0397 & 0.0398 & 0.845 & 0.855 & 0.0060 & 0.0017 & 0.0181 & 0.0158 \\
Eikos & option names & 0.006 & 0.007 & 0.1063 & 0.1063 & 0.823 & 0.840 & 0.0037 & 0.0037 & 0.0111 & 0 \\
Jev & option order & 0.000 & 0.000 & 0.0397 & 0.0394 & 0.845 & 0.823 & 0.0093 & 0.0052 & 0.0280 & 0.0158 \\
Eikos & option order & 0.006 & 0.006 & 0.1063 & 0.1063 & 0.823 & 0.835 & 0.0046 & 0.0046 & 0.0137 & 0 \\
Qwen3.5 & option order & 0.117 & 0.115 & 0.2002 & 0.1969 & 0.858 & 0.795 & 0.0175 & 0.0175 & 0.0526 & 0 \\
Jev & one question per request & 0.000 & 0.000 & 0.0397 & 0.0396 & 0.845 & 0.845 & 0.0050 & 0.0009 & 0.0151 & 0.0158 \\
\midrule
\multicolumn{12}{@{}l}{DBpedia-14 ($H = 1$, $G = 6$)} \\
Jev & option names & 0.000 & 0.000 & 0.0128 & 0.0128 & 0.995 & 0.995 & 0.0025 & 0.0001 & 0.0050 & 0.0053 \\
Eikos & option names & 0.000 & 0.000 & 0.0789 & 0.0790 & 0.985 & 0.985 & 0.0005 & 0.0005 & 0.0009 & 0 \\
Jev & option order & 0.000 & 0.000 & 0.0128 & 0.0128 & 0.995 & 0.995 & 0.0026 & 0.0001 & 0.0051 & 0.0053 \\
Eikos & option order & 0.000 & 0.000 & 0.0789 & 0.0788 & 0.985 & 0.980 & 0.0014 & 0.0014 & 0.0028 & 0 \\
Qwen3.5 & option order & 0.118 & 0.118 & 0.2110 & 0.2116 & 0.990 & 0.950 & 0.0034 & 0.0034 & 0.0068 & 0 \\
Jev & one question per request & 0.000 & 0.000 & 0.0128 & 0.0129 & 0.995 & 0.995 & 0.0025 & 0.0000 & 0.0050 & 0.0053 \\
\bottomrule
\end{tabular}

  \normalsize
  \caption{Preregistered controls on the 200 exploration items of each data set, each against the
    main condition it changes in one factor. ``anytime bound'' is the lower bound on $L_{H,G}$
    computed as in the pilot, at $\alpha = 0.05$ and the selected $H$ and $G$, on a grid of
    $10^{-3}$; ``plug-in $L_{H,G}$'' is the uncorrected empirical value, never a bound; ``Choice
    accuracy'' is the accuracy of the choice query, computed with the items' labels, ties for the
    largest probability counted fractionally. ``shift'' is the item-matched weighted $\ell_1$
    difference of the answers, over all components and over the choice query's alone (at the
    bundle's weights); $(1 + H)$ times the shift over all components bounds how far apart the
    plug-in values of $L_H$ and $L_{H,G}$ under the two conditions can be. The noise floor is $(1 +
    H)$ times the largest $W_{1,w}$ between the empirical laws of two of the system's five
    main-condition calls; it is $0$ for the local systems, whose calls return identical answers. A
    $0$ is exactly zero; $0.0000$ is a value rounded to zero. None of these values is a confirmatory
    certificate. The main conditions' anytime bounds and plug-in values and the noise floors are the
    pilot's; the accuracy of the choice query, under both conditions, and the shifts were computed
    with the comparison.}
  \label{tab:controls}
\end{table}

\clearpage
\section{The Real-Data Study in Full}
\label{app:study}

This appendix lists what the protocol froze for the study of Section~\ref{sec:exp-real} and what the
study recorded, using the protocol's names for its rounds, its decision rule and its parts. The
study ran in two \emph{rounds}: the \emph{pilot}, on the exploration items, and the \emph{main
collection}, which asked the confirmation items and gave the confirmatory certificates.
\emph{Gate~B} is the rule, fixed before the pilot, that decided after it whether each system went on
to the main collection (Appendix~\ref{app:pilot}). \emph{Part~A} and \emph{Part~B} are the two parts
in which the protocol was frozen, before the pilot and before the main collection
(Appendix~\ref{app:preregistration}). Every value in the tables of this appendix is read from a
committed file: the frozen objects, which the registered protocol lists by SHA-256; the pilot's
Gate~B results, which Part~B names by SHA-256; and the records of the main collection, each written
once by our collection code. The one exception is Table~\ref{tab:noise-floor-pairs}, whose distances
we computed afterwards from the recorded responses (Appendix~\ref{app:preregistration}). The tables
are generated from those files (Table~\ref{tab:reproduction}).

\subsection{Data and Split}

Each data set is read from its Hugging Face repository at a fixed revision, every file pinned by its
SHA-256 (Table~\ref{tab:data}). Every row of the train split is an item, treated as an i.i.d.\ draw
from an unknown population; the label spaces are treated as exhaustive, an assumption we do not
check, so no ``none'' enters a bundle. One seeded random permutation of each pool fixes the arrival
order and splits the pool: the exploration side takes its first 200 items, and the confirmation side
the next 10{,}000 that no predictor had answered before the split was drawn. Twelve items had been
answered by then: ten rows of the AG News test split, used to test our code, and two rows queried by
hand. We chose them without looking at their values, which the code does not check; only one of them
lies in a pool, and the permutation placed it on neither side. The committed assignment tables list
item ids and text hashes, never texts or labels.

\begin{table}[t]
  \centering
  \scriptsize
  \setlength{\tabcolsep}{2.5pt}
  \begin{tabular}{@{}llrrlrrr@{}}
\toprule
Data set & Repository at revision & pool & $K$ & item text & seed & exploration & confirmation \\
\midrule
AG News & \texttt{fancyzhx/ag\_news@eb185aa} & 120,000 & 4 & \texttt{\{text\}} & 20261001 & 200 & 10,000 \\
DBpedia-14 & \texttt{fancyzhx/dbpedia\_14@9abd46c} & 560,000 & 14 & \texttt{\{title\}\textbackslash n\{content\}} & 20261001 & 200 & 10,000 \\
\bottomrule
\end{tabular}

  \normalsize
  \caption{The data sets and the exploration/confirmation split. ``item text'' is how a row becomes
    the text the predictors see, unnormalised (\texttt{\textbackslash n} is a line break); the pool
    is the whole train split, in file order, and the seed fixes its permutation.}
  \label{tab:data}
\end{table}

\subsection{Bundles}

The main bundle of each data set asks only unconditional events (Table~\ref{tab:bundles}): the
choice query over the labels, whether the item has each label (an atom) and whether it does not (a
complement), and on DBpedia-14 five unions of labels, each asked as the explicit disjunction of its
member labels and never by a superclass name (Table~\ref{tab:labels}). Every role has the same total
weight, split equally among its components; these weights fix the audited $\Theta_w$. Before the
pilot, we checked that every question template means what its role asks: that a complement negates
and that a union is a disjunction. The option controls of Appendix~\ref{app:controls} keep the
components, the weights and the questions of the atoms, complements and unions, and change only the
choice query: its options named by their labels, or listed in the reverse order, with the list of
labels in its question reversed too and the neutral names given by position
(Table~\ref{tab:labels}).

\begin{table}[t]
  \centering
  \small
  \setlength{\tabcolsep}{4pt}
  \begin{tabular}{@{}lp{0.56\linewidth}rr@{}}
\toprule
Role & Question & Components & Weight each \\
\midrule
\multicolumn{4}{@{}l}{AG News ($K = 4$, $m = 12$)} \\
Choice & What is the topic of this news article? & 4 & $1/12$ \\
Atom & Is the topic of this news article $\langle$\textit{label}$\rangle$? & 4 & $1/12$ \\
Complement & Is the topic of this news article something other than $\langle$\textit{label}$\rangle$? & 4 & $1/12$ \\
\multicolumn{4}{@{}>{\raggedright\arraybackslash}p{0.97\linewidth}}{Every question ends with: Every article has exactly one of these topics: $\langle$\textit{labels}$\rangle$.} \\
\multicolumn{4}{@{}>{\raggedright\arraybackslash}p{0.97\linewidth}}{\footnotesize For example: \texttt{Is the topic of this news article world news? Every article has exactly one of these topics: \textquotesingle{}world news\textquotesingle{}, \textquotesingle{}sports\textquotesingle{}, \textquotesingle{}business\textquotesingle{} and \textquotesingle{}science and technology\textquotesingle{}.}} \\
\midrule
\multicolumn{4}{@{}l}{DBpedia-14 ($K = 14$, $m = 47$)} \\
Choice & What is the subject of this encyclopedia entry? & 14 & $1/56$ \\
Atom & Is the subject of this encyclopedia entry $\langle$\textit{label}$\rangle$? & 14 & $1/56$ \\
Complement & Is the subject of this encyclopedia entry something other than $\langle$\textit{label}$\rangle$? & 14 & $1/56$ \\
Union & Is the subject of this encyclopedia entry $\langle$\textit{labels}$\rangle$? & 5 & $1/20$ \\
\multicolumn{4}{@{}>{\raggedright\arraybackslash}p{0.97\linewidth}}{Every question ends with: Every entry is about exactly one of these: $\langle$\textit{labels}$\rangle$.} \\
\multicolumn{4}{@{}>{\raggedright\arraybackslash}p{0.97\linewidth}}{\footnotesize For example: \texttt{Is the subject of this encyclopedia entry a company? Every entry is about exactly one of these: \textquotesingle{}a company\textquotesingle{}, \textquotesingle{}an educational institution\textquotesingle{}, \textquotesingle{}an artist\textquotesingle{}, \textquotesingle{}an athlete\textquotesingle{}, \textquotesingle{}an office holder\textquotesingle{}, \textquotesingle{}a means of transportation\textquotesingle{}, \textquotesingle{}a building\textquotesingle{}, \textquotesingle{}a natural place\textquotesingle{}, \textquotesingle{}a village\textquotesingle{}, \textquotesingle{}an animal\textquotesingle{}, \textquotesingle{}a plant\textquotesingle{}, \textquotesingle{}an album\textquotesingle{}, \textquotesingle{}a film\textquotesingle{} and \textquotesingle{}a written work\textquotesingle{}.}} \\
\bottomrule
\end{tabular}

  \normalsize
  \caption{The main bundles. $\langle$\textit{label}$\rangle$ is a label's description
    (Table~\ref{tab:labels}); $\langle$\textit{labels}$\rangle$ lists descriptions, each in straight
    single quotes, separated by commas and with ``and'' before the last one, or ``or'' in a union,
    which lists its member labels. The choice query offers every label as an option, named by a
    neutral letter in the main bundle and described by the label's description; how the options are
    shown is the predictor's readout (Appendix~\ref{app:readouts}). The weights sum to $1$ over the
    $m$ components.}
  \label{tab:bundles}
\end{table}

\begin{table}[t]
  \centering
  \small
  \setlength{\tabcolsep}{3pt}
  \begin{tabular}{@{}llcccc@{}}
\toprule
 & & \multicolumn{3}{c}{Option name} & \\
\cmidrule(lr){3-5}
Label & Description & main & reversed & named & Union \\
\midrule
\multicolumn{6}{@{}l}{AG News} \\
World & world news & A & D & World &  \\
Sports & sports & B & C & Sports &  \\
Business & business & C & B & Business &  \\
Sci/Tech & science and technology & D & A & Sci/Tech &  \\
\midrule
\multicolumn{6}{@{}l}{DBpedia-14} \\
Company & a company & A & N & Company & 1 \\
EducationalInstitution & an educational institution & B & M & EducationalInstitution & 1 \\
Artist & an artist & C & L & Artist & 2 \\
Athlete & an athlete & D & K & Athlete & 2 \\
OfficeHolder & an office holder & E & J & OfficeHolder & 2 \\
MeanOfTransportation & a means of transportation & F & I & MeanOfTransportation & -- \\
Building & a building & G & H & Building & 3 \\
NaturalPlace & a natural place & H & G & NaturalPlace & 3 \\
Village & a village & I & F & Village & 3 \\
Animal & an animal & J & E & Animal & 4 \\
Plant & a plant & K & D & Plant & 4 \\
Album & an album & L & C & Album & 5 \\
Film & a film & M & B & Film & 5 \\
WrittenWork & a written work & N & A & WrittenWork & 5 \\
\multicolumn{6}{@{}l}{\footnotesize Union 1: Company or EducationalInstitution} \\
\multicolumn{6}{@{}l}{\footnotesize Union 2: Artist, Athlete or OfficeHolder} \\
\multicolumn{6}{@{}l}{\footnotesize Union 3: Building, NaturalPlace or Village} \\
\multicolumn{6}{@{}l}{\footnotesize Union 4: Animal or Plant} \\
\multicolumn{6}{@{}l}{\footnotesize Union 5: Album, Film or WrittenWork} \\
\bottomrule
\end{tabular}

  \normalsize
  \caption{The labels in the order of the label space, the description that fills
    $\langle$\textit{label}$\rangle$, the name of each label's option in the choice query of the
    main bundle and of the two option controls, and, on DBpedia-14, the union a label belongs to,
    numbered as listed below the labels. AG News has no union.}
  \label{tab:labels}
\end{table}

\subsection{Readouts}
\label{app:readouts}

Each system is one predictor configuration, an audit target of its own: changing any of its
settings, such as the model version, how the answers are read or whether the bundle is sent in one
request, makes another one. The local models' weights are pinned by revision and by the SHA-256 of
every file. The prompts and templates below are those of the frozen configurations, verbatim.

{\raggedright
\begin{description}
\item[Jev] \texttt{jev-1.13.0}, named explicitly in every request to the official TypeSafe API; no temperature is set and nothing is rounded on our side. The whole bundle goes in one request per item: a JSON body with the item text verbatim as \texttt{state}, the model and one question per query, each yes/no question (an atom, a complement or a union) as a \texttt{noul} and the choice query as a \texttt{choice} whose criteria map each option name to its description in the bundle's order, each with the question as its instructions. The probabilities are read as returned; in the pilot's first call under the main condition, every one was within $10^{-9}$ of a multiple of $10^{-2}$. The response's \texttt{model} field is recorded on every attempt that carries one, and an answer from another version stops the collection. The control of how the bundle is sent asks the same questions one per request, each with the item text.
\item[Eikos] \texttt{caiovicentino1/Eikos-4B-MLX-8bit} at revision \texttt{5b1a885} through mlx-lm (mlx 0.32.3, mlx-lm 0.31.3), one forward pass per question, the model's chat template with a generation prompt and with thinking disabled. The system prompt is \texttt{Apply the supplied criterion to the supplied evidence. Choose exactly one listed option. Respond with only its uppercase letter, with no explanation or reasoning.} The user turn is, in Python, \texttt{json.dumps(\{\textquotedbl{}evidence\textquotedbl{}: text, \textquotedbl{}criterion\textquotedbl{}: question, \textquotedbl{}options\textquotedbl{}: [\{\textquotedbl{}letter\textquotedbl{}: L, \textquotedbl{}description\textquotedbl{}: \textquotedbl{}\textless{}name\textgreater{}: \textless{}description\textgreater{}\textquotedbl{}\}, ...]\}, ensure\_ascii=False)}. A yes/no question is asked with the two options \texttt{true: yes} (letter A) and \texttt{false: no} (letter B); the choice query lists its options in the bundle's order under the letters A, B, \dots\ by position. The answer is read from the next-token logits of the question's letters at the last position (float32), normalised by a softmax at $T = 1$ (the release's calibration file sets $T = 1$); a yes/no question's answer is the probability of A.
\item[Qwen3.5] \texttt{Qwen/Qwen3.5-4B} at revision \texttt{851bf6e} through mlx-lm (mlx 0.32.3, mlx-lm 0.31.3), one forward pass per question, the model's chat template with a generation prompt and with thinking disabled and no system prompt. A yes/no question is asked as \texttt{\{state\}\textbackslash n\textbackslash nQuestion: \{question\}\textbackslash nAnswer with Yes or No.} and read from the next-token logits of \texttt{Yes} and \texttt{No}; the choice query is asked as \texttt{\{state\}\textbackslash n\textbackslash nQuestion: \{question\}\textbackslash nOptions:\textbackslash n\{options\}\textbackslash nAnswer with the letter of the correct option.}, one line \texttt{\{name\}. \{description\}} per option, and read from the logits of the option names, each a single token. Here \texttt{\{state\}} is the item text and \texttt{\{question\}} the question. The logits (float32, at the last position) are normalised by a softmax at $T = 1$; a yes/no question's answer is the probability of \texttt{Yes}.
\item[DeBERTa] \texttt{MoritzLaurer/deberta-v3-large-zeroshot-v2.0-c} at revision \texttt{b2730f1} through PyTorch (torch 2.14.1, transformers 5.18.0), in float32 on MPS: one NLI pair per queried event, with the item text as the premise (truncated first if the pair exceeds 512 tokens) and a hypothesis built from the labels' descriptions in the bundle. The hypotheses for an atom, a complement and (on DBpedia-14) a union are, on AG News \texttt{This news article is about \{label\}.}, \texttt{This news article is not about \{label\}.}; on DBpedia-14 \texttt{This encyclopedia entry is about \{label\}.}, \texttt{This encyclopedia entry is not about \{label\}.}, \texttt{This encyclopedia entry is about \{labels\}.}; a union lists its labels as the questions do. The answer for an atom, a complement or a union is the softmax of the pair's own entailment and not-entailment logits; the choice query takes the atom hypothesis of each option, and its answer is the softmax over the options' entailment logits.
\end{description}

\par}

\subsection{The Pilot}
\label{app:pilot}

The pilot queried the 200 exploration items of each data set: each system under its main condition
five times and under each control once. Each such pass over the items is a \emph{collection}. The
five calls under the main condition give the \emph{noise floor}, $(1 + H)$ times the largest
$W_{1,w}$ between the empirical laws of two of them; the first call (repeat index $0$) gives the
pilot's values. An item is \emph{settled} when its outcome, an answer or a failure, is logged. A
\emph{defect} is a failing check of our code, a record that our code rejects because of its own
code, or a crash of our code while an item is asked; an invalid answer of the predictor is a
failure, never a defect. All $52$ collections ($26$ per data set) answered every item, with no
failure and no defect, and no item of a main condition was answered from the cache: the pilot's
committed records list every call of a main condition, and each comparison of
Table~\ref{tab:controls} uses all 200 items of its control. For every candidate $(H, G)$, each
system's anytime lower bound of $\mathcal{U}_{H,G}$ learnt sequentially was computed on its main
condition's stream, with the pilot's own horizon in the step size, $\eta = \sqrt{G/200}$; the pair
with the largest mean of these bounds, on the $\tau$ grid of $10^{-3}$, was taken for the data set,
a tie going to the smaller $H$, then to the smaller $G$ (Table~\ref{tab:pilot-selection}). On AG
News $(2, 6)$ and $(4, 6)$ tied at $0.031$ and $(2, 6)$ was taken; on DBpedia-14 $(1, 6)$, $(2, 6)$
and $(4, 6)$ tied at $0.030$ and $(1, 6)$ was taken. On both data sets the rule took a pair whose
exact mean was not the largest. Gate~B then decided for each system in four steps
(Table~\ref{tab:pilot}). Step~(1) looks for defects, and found none in any system. The later steps
apply in order, and the first that applies decides: (2) a positive anytime bound sends the system on
to the main collection; (3) otherwise, so does a plug-in value of $L_{H,G}$ above both the noise
floor and the threshold of the system's $T_{\max}$; (4) otherwise the system is not collected. Jev
takes $T_{\max} = 10{,}000$; a local system the largest candidate whose collection fits in 48 hours
at its pilot time per item, or $2{,}000$ if none does. The thresholds are, rounded to three
decimals, the smallest synthetic effects whose anytime bound, with $\mathcal{U}_{H,G}$ learnt
sequentially, was positive in at least nine seeds of ten within that many items in the power design
(Table~\ref{tab:power}; $K = 4$, $H = 2$, $G = 6$, so for the real bundles they are only a guide),
and the plug-in value, biased upwards, is read only as an effect size. None of the pilot's values
enters a confirmatory certificate.

Part~A also lists among the noise floor's reported values the $W_{1,w}$ between every pair of
repeats (Table~\ref{tab:noise-floor-pairs}). The pilot recorded only their maximum, which Gate~B
uses; we computed all ten pairs again after the main collection
(Appendix~\ref{app:preregistration}). The local systems answered identically in all five calls;
Jev's distances lay between $0.0048$ and $0.0053$ on AG News and between $0.0024$ and $0.0027$ on
DBpedia-14.

\begin{table}[t]
  \centering
  \small
  \begin{tabular}{@{}lrrrrrrrrc@{}}
\toprule
Data set & $H$ & $G$ & Jev & Eikos & Qwen3.5 & DeBERTa & mean & on the grid & selected \\
\midrule
AG News & 1 & 6 & 0.000 & 0.004 & 0.113 & 0.000 & 0.02925 & 0.029 &  \\
 & 1 & 11 & 0.000 & 0.000 & 0.104 & 0.000 & 0.026 & 0.026 &  \\
 & 2 & 6 & 0.000 & 0.006 & 0.117 & 0.000 & 0.03075 & 0.031 & $\checkmark$ \\
 & 2 & 11 & 0.000 & 0.000 & 0.111 & 0.000 & 0.02775 & 0.028 &  \\
 & 4 & 6 & 0.000 & 0.006 & 0.118 & 0.000 & 0.031 & 0.031 &  \\
 & 4 & 11 & 0.000 & 0.000 & 0.113 & 0.000 & 0.02825 & 0.028 &  \\
\midrule
DBpedia-14 & 1 & 6 & 0.000 & 0.000 & 0.118 & 0.000 & 0.0295 & 0.030 & $\checkmark$ \\
 & 1 & 11 & 0.000 & 0.000 & 0.108 & 0.000 & 0.027 & 0.027 &  \\
 & 2 & 6 & 0.000 & 0.000 & 0.120 & 0.000 & 0.03 & 0.030 &  \\
 & 2 & 11 & 0.000 & 0.000 & 0.111 & 0.000 & 0.02775 & 0.028 &  \\
 & 4 & 6 & 0.000 & 0.000 & 0.120 & 0.000 & 0.03 & 0.030 &  \\
 & 4 & 11 & 0.000 & 0.000 & 0.113 & 0.000 & 0.02825 & 0.028 &  \\
\bottomrule
\end{tabular}

  \normalsize
  \caption{Selection of $H$ and $G$ on the pilot. Each system's anytime lower bound
    (Proposition~\ref{prop:anytime}, $\alpha = 0.05$) of $\mathcal{U}_{H,G}$ learnt sequentially,
    $\eta = \sqrt{G/200}$, on the 200 exploration items of its main condition, for every candidate
    pair, with the mean over the four systems, exact and on the grid of $10^{-3}$ (halves rounded
    up). The selected pair has the largest mean on the grid, a tie going to the smaller $H$, then to
    the smaller $G$. A larger $H$ bounds a different $\Gamma_H$, so a larger mean does not mean a
    better configuration.}
  \label{tab:pilot-selection}
\end{table}

\begin{table}[t]
  \centering
  \scriptsize
  \setlength{\tabcolsep}{2pt}
  \begin{tabular}{@{}llrrrrrrrrrcr@{}}
\toprule
 & & & & \multicolumn{4}{c}{noise floor} & & & & & \\
\cmidrule(lr){5-8}
Data set & System & answered & s/item & $\max W_{1,w}$ & $(1 + H)\,\cdot$ & identical & $\max\lvert\Delta\rvert$ & bound & plug-in & threshold & step & $T_{\max}$ \\
\midrule
\multicolumn{13}{@{}l}{AG News ($H = 2$, $G = 6$)} \\
 & Jev & 200/200 & 0.21 & 0.0053 & 0.0158 & 0.05 & 0.2100 & 0.000 & 0.0397 & 0.013 & (3) & 10,000 \\
 & Eikos & 200/200 & 9.97 & 0 & 0 & 1.00 & 0 & 0.006 & 0.1063 & 0.013 & (2) & 10,000 \\
 & Qwen3.5 & 200/200 & 5.07 & 0 & 0 & 1.00 & 0 & 0.117 & 0.2002 & 0.013 & (2) & 10,000 \\
 & DeBERTa & 200/200 & 0.66 & 0 & 0 & 1.00 & 0 & 0.000 & 0.0819 & 0.013 & (3) & 10,000 \\
\midrule
\multicolumn{13}{@{}l}{DBpedia-14 ($H = 1$, $G = 6$)} \\
 & Jev & 200/200 & 0.20 & 0.0027 & 0.0053 & 0.00 & 0.2200 & 0.000 & 0.0128 & 0.013 & (4) & 10,000 \\
 & Eikos & 200/200 & 39.90 & 0 & 0 & 1.00 & 0 & 0.000 & 0.0789 & 0.029 & (3) & 2,000 \\
 & Qwen3.5 & 200/200 & 34.16 & 0 & 0 & 1.00 & 0 & 0.118 & 0.2110 & 0.016 & (2) & 5,000 \\
 & DeBERTa & 200/200 & 3.62 & 0 & 0 & 1.00 & 0 & 0.000 & 0.0401 & 0.013 & (3) & 10,000 \\
\bottomrule
\end{tabular}

  \normalsize
  \caption{Gate~B on the pilot, at the selected $(H, G)$. ``s/item'' is the mean time of asking an
    item in the first call under the main condition (repeat index $0$), model loading excluded. The
    noise floor is $(1 + H)$ times the largest $W_{1,w}$ between the empirical laws of two of the
    system's five main-condition calls (Theorem~\ref{thm:stable} and Lemma~\ref{lem:grid}: it bounds
    how far the empirical $L_H$ and $L_{H,G}$ moved between the calls); ``identical'' is the share
    of items answered identically in all five, and $\max\lvert\Delta\rvert$ the largest difference
    of one component of one item between two calls. ``bound'' is the anytime bound of
    Table~\ref{tab:pilot-selection}; ``plug-in'' is the uncorrected empirical $L_{H,G}$, an effect
    size and never a bound; the threshold is that of the system's $T_{\max}$. ``step'' is the Gate~B
    step that decided; steps (2) and (3) send a system on to the main collection, and step (4) does
    not. A $0$ is exactly zero.}
  \label{tab:pilot}
\end{table}

\begin{table}[t]
  \centering
  \scriptsize
  \setlength{\tabcolsep}{3pt}
  \begin{tabular}{@{}lrrrrrrrrrrrr@{}}
\toprule
 & \multicolumn{10}{c}{$W_{1,w}$ between repeats $r$ and $s$} & & \\
\cmidrule(lr){2-11}
System & 0--1 & 0--2 & 0--3 & 0--4 & 1--2 & 1--3 & 1--4 & 2--3 & 2--4 & 3--4 & max & $(1 + H)\,\cdot$ max \\
\midrule
\multicolumn{13}{@{}l}{AG News ($H = 2$)} \\
Jev & 0.0048 & 0.0049 & 0.0051 & 0.0048 & 0.0048 & 0.0049 & 0.0053 & 0.0049 & 0.0051 & 0.0052 & 0.0053 & 0.0158 \\
Eikos & 0 & 0 & 0 & 0 & 0 & 0 & 0 & 0 & 0 & 0 & 0 & 0 \\
Qwen3.5 & 0 & 0 & 0 & 0 & 0 & 0 & 0 & 0 & 0 & 0 & 0 & 0 \\
DeBERTa & 0 & 0 & 0 & 0 & 0 & 0 & 0 & 0 & 0 & 0 & 0 & 0 \\
\midrule
\multicolumn{13}{@{}l}{DBpedia-14 ($H = 1$)} \\
Jev & 0.0026 & 0.0025 & 0.0024 & 0.0025 & 0.0027 & 0.0026 & 0.0024 & 0.0025 & 0.0024 & 0.0026 & 0.0027 & 0.0053 \\
Eikos & 0 & 0 & 0 & 0 & 0 & 0 & 0 & 0 & 0 & 0 & 0 & 0 \\
Qwen3.5 & 0 & 0 & 0 & 0 & 0 & 0 & 0 & 0 & 0 & 0 & 0 & 0 \\
DeBERTa & 0 & 0 & 0 & 0 & 0 & 0 & 0 & 0 & 0 & 0 & 0 & 0 \\
\bottomrule
\end{tabular}

  \normalsize
  \caption{The noise floor's distances: $W_{1,w}$ between the empirical laws of every pair of a
    system's five calls under the main condition (repeat indices $0$ to $4$) on the 200 exploration
    items, as Part~A lists them among the reported values. ``max'' and ``$(1 + H)\,\cdot$ max'' are
    the pilot's noise floor of Table~\ref{tab:pilot}, the value Gate~B used. The pilot recorded only
    the maximum; all ten pairs were computed again after the main collection
    (Appendix~\ref{app:preregistration}). A $0$ is exactly zero.}
  \label{tab:noise-floor-pairs}
\end{table}

\subsection{The Main Collection}

Each system that Gate~B sent on was asked the confirmation items of its data set once each, in
arrival order, up to its $T_{\max}$, from the commit tagged when Part~B was frozen
(Table~\ref{tab:confirm-runs}). Every collection settled all of its items, with no failure, no
defect, no item that had to be settled from the records after an interruption of the process and no
item answered from the cache, so every stream runs to its $T_{\max}$. A \emph{safe stop} halts a
collection without settling the item in flight, which is asked again when the collection resumes
unless the stop closed the collection. The two safe stops were Jev's on AG News: each time, the
request for one item failed on every retry, the last attempt unable to connect; as the protocol
prescribes, this missing answer was counted once (a third count would have settled the item as
failed), the item was asked again when the collection resumed, and both items were then answered.
The certificates were computed once, after every collection had completed.

\begin{table}[t]
  \centering
  \scriptsize
  \setlength{\tabcolsep}{3pt}
  \begin{tabular}{@{}llrrrrrrrrl@{}}
\toprule
Data set & System & $T_{\max}$ & settled & stream & failed & defects & interrupted & safe stops & s/item & sessions \\
\midrule
AG News & Jev & 10,000 & 10,000 & 10,000 & 0 & 0 & 0 & 2 & 0.39 & 10-05 to 10-06 \\
 & Eikos & 10,000 & 10,000 & 10,000 & 0 & 0 & 0 & 0 & 8.52 & 10-07 to 10-09 \\
 & Qwen3.5 & 10,000 & 10,000 & 10,000 & 0 & 0 & 0 & 0 & 5.06 & 10-07 to 10-07 \\
 & DeBERTa & 10,000 & 10,000 & 10,000 & 0 & 0 & 0 & 0 & 0.75 & 10-06 to 10-06 \\
\midrule
DBpedia-14 & Eikos & 2,000 & 2,000 & 2,000 & 0 & 0 & 0 & 0 & 40.45 & 10-10 to 10-11 \\
 & Qwen3.5 & 5,000 & 5,000 & 5,000 & 0 & 0 & 0 & 0 & 32.07 & 10-09 to 10-10 \\
 & DeBERTa & 10,000 & 10,000 & 10,000 & 0 & 0 & 0 & 0 & 3.60 & 10-06 to 10-07 \\
 & Jev & \multicolumn{9}{l}{no main collection (Gate B step (4))} \\
\bottomrule
\end{tabular}

  \normalsize
  \caption{The main collection. ``settled'' counts the items whose outcome was logged and ``stream''
    those before the first failure or defect, which the certificates use; ``interrupted'' counts
    items settled from the records after the process had stopped; a safe stop (here, an item whose
    answer went missing) leaves the item unsettled, to be asked again when the collection resumes.
    ``s/item'' is the mean time of asking an item; ``sessions'' gives the month and day (2026, UTC)
    on which the collection's first session started and its last one ended. Jev on DBpedia-14
    reached step (4) of Gate~B (Table~\ref{tab:pilot}) and was not collected.}
  \label{tab:confirm-runs}
\end{table}

\clearpage
\section{Preregistration and Reproduction}
\label{app:preregistration}

This appendix describes the two registrations of the protocol, whether anything departed from them,
what was not preregistered, and how the figures and tables are reproduced.

\subsection{The Two Registrations}

The protocol (a file \texttt{PROTOCOL.md}) was frozen in two parts, each at a tagged commit of the
project's repository, and each part was registered on OSF as a registration of its own, not tied to
an OSF project, with the protocol file and the files that part names attached
(Table~\ref{tab:rounds}). Part~A, frozen before the pilot, fixed the data and the split, the bundles
and their wording, the predictor configurations, the local weight files, a fingerprint of the
requests each family builds, the candidate settings, the selection criteria and the budget. Part~B,
frozen after Gate~B and before any confirmation response was seen, fixed the settings of every
confirmatory certificate, read by code from the pilot's Gate~B results, and the confirmatory family
with its $\alpha$. Each frozen object is a canonical JSON file, and the protocol lists the SHA-256
of each, so the SHA-256 of the protocol file covers them all. The code that ran each round is pinned
by the commit of its tag: our code refused to collect or to compute unless it ran from that commit,
unmodified. OSF dates the two registrations 2026-10-01 18:55 and 2026-10-05 07:48 (UTC), before the
rounds started, and our code recorded each registration's OSF address in its round's record when the
round started, before asking any item. Both registrations were embargoed until this paper was first
posted.

\begin{table}[t]
  \centering
  \small
  \setlength{\tabcolsep}{4pt}
  \begin{tabular}{@{}lllrr@{}}
\toprule
Part frozen & Tag at commit & Round & collections & settled \\
\midrule
A (pre-pilot) & \texttt{v0.12.0 at 4921775} & pilot & 52 & 10,400 \\
\multicolumn{5}{@{}>{\raggedright\arraybackslash}p{0.98\linewidth}@{}}{\footnotesize OSF registration \url{https://osf.io/dj3ut}} \\
\multicolumn{5}{@{}>{\raggedright\arraybackslash}p{0.98\linewidth}@{}}{\footnotesize SHA-256 of \texttt{PROTOCOL.md} at the tag: \texttt{90c316be6ab55501e6d8af79da9f7102b59e689726b8b5919563b203e568b3ab}} \\
\multicolumn{5}{@{}>{\raggedright\arraybackslash}p{0.98\linewidth}@{}}{\footnotesize Round opened 2026-10-01 19:07; records 2026-10-01 19:07 to 2026-10-04 23:13; Gate B computed 2026-10-05 00:24 (UTC)} \\
\midrule
B (pre-confirmation) & \texttt{v0.16.0 at cfd9a27} & main collection & 7 & 57,000 \\
\multicolumn{5}{@{}>{\raggedright\arraybackslash}p{0.98\linewidth}@{}}{\footnotesize OSF registration \url{https://osf.io/d5jr6}} \\
\multicolumn{5}{@{}>{\raggedright\arraybackslash}p{0.98\linewidth}@{}}{\footnotesize SHA-256 of \texttt{PROTOCOL.md} at the tag: \texttt{476b77c177117dff2e6904535ee5e11c2648d48d493165e7b892d7d2c98fd0b0}} \\
\multicolumn{5}{@{}>{\raggedright\arraybackslash}p{0.98\linewidth}@{}}{\footnotesize Round opened 2026-10-05 07:54; records 2026-10-05 07:54 to 2026-10-11 19:35; certificates computed 2026-10-11 19:40 (UTC)} \\
\bottomrule
\end{tabular}

  \normalsize
  \caption{The two registered rounds. Each part of the protocol was frozen at the commit its tag
    names and registered on OSF; the round ran from that commit. ``collections'' counts the
    collections the round ran and ``settled'' the items they settled, an item counted once per
    collection (the pilot asked each of its 400 exploration items in the 26 collections of its data
    set). Times are to the minute.}
  \label{tab:rounds}
\end{table}

\subsection{Deviations, Revisions and Analyses That Were Not Preregistered}

We recorded no deviation from the protocol and no revision of it: every run followed the frozen
procedures (Tables~\ref{tab:pilot} and~\ref{tab:confirm-runs}), and the protocol file is byte for
byte the version registered last, whose SHA-256 is that of Part~B's tag in Table~\ref{tab:rounds}.
One reported value was computed late. Part~A lists among the noise floor's reported values the
$W_{1,w}$ between every pair of repeats; the pilot recorded only their maximum, the value Gate~B
uses. We computed all ten pairs again after the main collection (Table~\ref{tab:noise-floor-pairs}),
from the logs Gate~B read, checked by SHA-256, and the recorded responses they name, with the
noise-floor code the pilot ran, unchanged since the pilot's tag, and the current version of the code
that reads the recorded responses. This reproduced every recorded maximum, share of identical items
and largest difference exactly. The recomputed maximum was only compared with the recorded one, and
the other pairs entered no decision. Gate~B sent Jev on DBpedia-14 to step (4): its plug-in value,
$0.0128$, fell short of the threshold of its $T_{\max}$, $0.013$, so it had no main collection, and
its pilot bound of $0$ is reported as it is (Figure~\ref{fig:real-data}); we kept the decision
although the value fell only just short of the threshold.

The protocol covers the real-data study only, although it takes the thresholds of Gate~B from the
synthetic power design; the synthetic experiments of Section~\ref{sec:exp-synthetic} and
Appendix~\ref{app:synthetic} were not registered. Three analyses of the real data were not
preregistered either, and none of them enters the confirmatory family or changes its $\alpha$. The
conditions of the second and the third were committed to the repository before any of their values
was computed. Those of the labelled estimates in the first were fixed before the estimates were
computed but committed together with them, so only our own record supports that order; for the
empirical $L_{\mathrm{joint}}$ beside them we recorded no order.
\begin{enumerate}
  \item The labelled descriptive metrics of Appendix~\ref{app:descriptive}: the estimates of
    $\Theta_w$, the Brier score and their estimation conditions, and the uncorrected empirical
    $L_{\mathrm{joint}}$ shown beside them for reference. The uncorrected empirical $L_0$ and
    $L_{\mathrm{marg}}$ there are the negative controls the protocol names.
  \item The split arm of $\mathcal{U}_0$ in Appendix~\ref{app:u0-split}, added after
    Figure~\ref{fig:real-data} had been seen.
  \item The quantities by which Appendix~\ref{app:controls} compares the controls with the main
    conditions; collecting the controls was preregistered.
\end{enumerate}
Beyond these, Section~\ref{sec:exp-real} quotes how many distinct values each system's outputs took
per coordinate, a descriptive count of the confirmatory streams, chosen after the data had been seen
and not preregistered either. In Appendix~\ref{app:controls} we set aside a reading we had fixed in
advance, that a small change of the plug-in value would mean answers robust to the factor, because
the inequality behind it runs the other way.

\subsection{Reproducing the Figures and Tables}

Every figure and every table of results or of the protocol is written by code into the manuscript's
\texttt{figures} directory, never by hand (Table~\ref{tab:reproduction}); the only table written by
hand is the comparison with related work (Table~\ref{tab:related}). Each module runs as
\begin{center}
  \texttt{uv run python -m coherence\_calibration.experiments.}\textit{module} \textit{step}:
\end{center}
\texttt{render} rebuilds its figures and tables from committed files alone, \texttt{compute}, where
the module has one, recomputes the results it reads first, and \texttt{all} does both. The synthetic
results are recomputed from the code, with fixed seeds where anything is drawn; \texttt{coverage}
takes about ten minutes. The results of the real-data study were computed from the response ledger,
which keeps every request and response verbatim, including the items' texts, and is not part of the
committed results. \texttt{uv run python -m coherence\_calibration.protocol check} recomputes every
hash the protocol lists, reads every frozen object back and checks that the code still builds
exactly those objects. By default it also draws both splits again from the downloaded data sets and
checks the pilot round that Part~B names against the full git history, and it fails when either is
missing; \texttt{--no-replay} and \texttt{--no-history} leave these checks out. \texttt{uv run
python -m coherence\_calibration.manuscript build} builds this manuscript and records the SHA-256 of
the PDF and of every source. The code runs on Python~3.12 with the versions of its dependencies
locked; the local predictors need an Apple Silicon Mac (MLX for Eikos and Qwen3.5, PyTorch on MPS
for DeBERTa), and Jev a key to the TypeSafe API.

\begin{table}[t]
  \centering
  \scriptsize
  \setlength{\tabcolsep}{2.4pt}
  \begin{tabular}{@{}l>{\raggedright\arraybackslash}p{0.28\linewidth}>{\raggedright\arraybackslash}p{0.3\linewidth}>{\raggedright\arraybackslash}p{0.22\linewidth}@{}}
\toprule
Module & Figures and tables & \texttt{render} reads & computed \\
\midrule
\texttt{fig1} & Figure~\ref{fig:four-point}, Table~\ref{tab:four-point} & \path{results/fig1/} & synthetic, from the code (\texttt{compute}) \\
\texttt{fig2} & Figure~\ref{fig:two-point}, Table~\ref{tab:perturbations} & \path{results/fig2/} & synthetic, from the code (\texttt{compute}) \\
\texttt{fig3} & Figure~\ref{fig:decomposition}, Table~\ref{tab:decomposition} & \path{results/fig3/} & synthetic, from the code (\texttt{compute}) \\
\texttt{fig4} & Figure~\ref{fig:real-data}, Table~\ref{tab:real-data} & \path{results/confirm/v0.16.0/} & \texttt{confirm certify}, once, from the response ledger \\
\texttt{power} & Figure~\ref{fig:power}, Table~\ref{tab:power} & \path{results/power/} & synthetic, from the code (\texttt{compute}) \\
\texttt{coverage} & Table~\ref{tab:coverage} & \path{results/coverage/} & synthetic, from the code (\texttt{compute}) \\
\texttt{blind\_spot} & Figure~\ref{fig:blind-spot}, Table~\ref{tab:blind-spot} & \path{results/blind_spot/} & synthetic, from the code (\texttt{compute}) \\
\texttt{descriptive} & Table~\ref{tab:descriptive} & \path{results/descriptive/v0.16.0/}, \path{results/confirm/v0.16.0/} & from the response ledger and the Items' labels (\texttt{compute}) \\
\texttt{u0\_split} & Table~\ref{tab:u0-split} & \path{results/u0_split/v0.16.0/}, \path{results/confirm/v0.16.0/} & from the response ledger (\texttt{compute}) \\
\texttt{controls} & Table~\ref{tab:controls} & \path{results/controls/v0.12.0/} & from the response ledger and the Items' labels (\texttt{compute}) \\
\texttt{noise\_floor\_pairs} & Table~\ref{tab:noise-floor-pairs} & \path{results/noise_floor/v0.12.0/} & from the response ledger (\texttt{compute}) \\
\texttt{protocol\_tables} & Table~\ref{tab:data}, Table~\ref{tab:bundles}, Table~\ref{tab:labels}, Appendix~\ref{app:readouts}, Table~\ref{tab:pilot-selection}, Table~\ref{tab:pilot}, Table~\ref{tab:confirm-runs}, Table~\ref{tab:rounds} & \path{PROTOCOL.md}, \path{results/protocol/}, \path{results/pilot/v0.12.0/}, \path{results/confirm/v0.16.0/} & frozen with the protocol; \texttt{pilot gate} and \texttt{confirm certify}, once each \\
\texttt{reproduction} & Table~\ref{tab:reproduction} & -- & this list, in the code \\
\bottomrule
\end{tabular}

  \normalsize
  \caption{How the generated figures and tables are produced. The modules belong to the package
    \texttt{coherence\_calibration.experiments}. A module's \texttt{render} step reads the committed
    files listed, and ``computed'' says how those were computed: by the module's own
    \texttt{compute} step where it says so, otherwise once, by the step named, which ran only from
    its round's tag. Files under \texttt{results/} are committed; the response ledger is not.}
  \label{tab:reproduction}
\end{table}

\vskip 0.2in
\bibliography{references}

\end{document}